\documentclass{article}

\usepackage[utf8]{inputenc} 
\usepackage[T1]{fontenc}    
\usepackage{hyperref}       
\usepackage{url}            
\usepackage{booktabs}       
\usepackage{amsfonts}       
\usepackage{nicefrac}       
\usepackage{xcolor}         
\usepackage{multirow}
\usepackage{graphicx}
\usepackage{natbib} 
\usepackage{soul}
\usepackage{wrapfig}
\usepackage{subcaption}
\usepackage{enumitem}
\usepackage{microtype}
\usepackage{xspace}

\usepackage{hyperref}

\usepackage[accepted]{icml2024}

\usepackage{amsmath}
\usepackage{amssymb}
\usepackage{mathtools}
\usepackage{amsthm}
\usepackage{pifont}

\usepackage[capitalize,noabbrev]{cleveref}

\usepackage{background}

\backgroundsetup{angle=0, 
scale=1,
color=black,
firstpage=true,
position=current page.north,
hshift=0pt,
vshift=-20pt,
contents={\ifnum\value{page}=1 PREPRINT: Accepted at the 41st Conference on Machine Learning (ICML 2024) \else \fi}
}

\theoremstyle{plain}

\theoremstyle{definition}

\theoremstyle{remark}

\DeclareMathOperator*{\argmin}{arg\,min}
\usepackage[textsize=tiny]{todonotes}
\usepackage{float}

\icmltitlerunning{Recurrent Early Exits for Federated Learning with Heterogeneous Clients}
\newcommand{\method}{ReeFL\xspace}
\newcommand{\module}{Ree\xspace}
\newcommand{\cmark}{\ding{51}}%
\newcommand{\xmark}{\ding{55}}%
\begin{document}

\twocolumn[
\icmltitle{Recurrent Early Exits for Federated Learning with Heterogeneous Clients}




\begin{icmlauthorlist}
\icmlauthor{Royson Lee}{sam,cam}
\icmlauthor{Javier Fernandez-Marques}{flow}
\icmlauthor{Shell Xu Hu}{sam}
\icmlauthor{Da Li}{sam}
\icmlauthor{Stefanos Laskaridis}{bra}
\icmlauthor{\L{}ukasz Dudziak}{sam}
\icmlauthor{Timothy Hospedales}{sam,edi}
\icmlauthor{Ferenc Huszár}{cam}
\icmlauthor{Nicholas D. Lane}{cam,flow}
\end{icmlauthorlist}

\icmlaffiliation{sam}{Samsung AI Center, Cambridge, UK}
\icmlaffiliation{flow}{Flower Labs, Cambridge, UK}
\icmlaffiliation{bra}{Brave Software, London, UK}
\icmlaffiliation{cam}{University of Cambridge, Cambridge, UK}
\icmlaffiliation{edi}{University of Edinburgh, Edinburgh, UK}

\icmlcorrespondingauthor{Royson Lee}{dsrl2@cam.ac.uk}

\icmlkeywords{Machine Learning, ICML}

\vskip 0.3in
]



\printAffiliationsAndNotice{}  

\begin{abstract}
Federated learning (FL) has enabled distributed learning of a model across multiple clients in a privacy-preserving manner. One of the main challenges of FL is to accommodate clients with varying hardware capacities; clients have differing compute and memory requirements. To tackle this challenge, recent state-of-the-art approaches leverage the use of early exits. Nonetheless, these approaches fall short of mitigating the challenges of joint learning multiple exit classifiers, often relying on hand-picked heuristic solutions for knowledge distillation among classifiers and/or utilizing additional layers for weaker classifiers. In this work, instead of utilizing multiple classifiers, we propose a recurrent early exit approach named \method{} that fuses features from different sub-models into a single shared classifier. Specifically, we use a transformer-based early-exit module shared among sub-models to \textit{i)}~better exploit multi-layer feature representations for task-specific prediction and \textit{ii)}~modulate the feature representation of the backbone model for subsequent predictions. We additionally present a per-client self-distillation approach where the best sub-model is automatically selected as the teacher of the other sub-models at each client. Our experiments on standard image and speech classification benchmarks across various emerging federated fine-tuning baselines demonstrate \method's effectiveness over previous works.

\end{abstract}

\section{Introduction}

Federated Learning (FL) has become an indispensable tool to train machine learning models collaboratively without exchanging raw data from clients' edge devices. In many practical scenarios, especially in the cross-device setting, these devices often differ in computational resources and may not have sufficient compute and/or memory resources to participate in federated training. 
This may yield convergence and fairness issues, especially if the least capable devices are consistently excluded from training.
Hence, a key challenge of FL is to divide the global model into heterogeneous sub-models to fit into a wide range of diverse devices while maintaining high performance of the global model. To this end, various existing works split the global model by pruning its channels, also known as \emph{width-based scaling}~\cite{diao2021heterofl,horvath2021fjord,mei2022resource,hong2022efficientsplitmix}, and/or utilizing early exits, also known as \emph{depth-based scaling}~\cite{liu2022inclusivefl,kim2023depthfl,ilhan2023scalefl,kang2023nefl}.

Recently, approaches that use depth-based scaling, some of which, in addition to width-based scaling, showed significantly better performance compared to approaches using width-based scaling solely. These depth-based works split the global model based on depth and deploy additional classifiers, allowing each sub-model to exit early. Having individual classifiers in a network, however, has been previously observed to degrade performance due to the competing optimization criteria across sub-models, leading to the accumulation of conflicting gradients from these classifiers~\cite{huang2018multi,song2018collaborative,li2019improved,laskaridis2020hapi}.
Another limitation of depth-based scaling is that shallower sub-models are trained more often than their deeper counterparts as these deep sub-models are not trained with data from clients with lower resource budgets~\cite{kim2023depthfl}. In addition, training with more data does not necessarily translate to performance gains as these shallower sub-models might not have sufficient parameters to learn good representations~\cite{liu2022inclusivefl}. 

To counteract these limitations, existing FL works use knowledge distillation among either classifiers or layers in the backbone model and/or add additional layers to improve the representation capacity of shallower sub-models at a cost of additional resources. These approaches often require manual selection of which layers to distill from/to, \textit{e.g.} the largest sub-model acts as a teacher for the other sub-models. However, the optimal choice of which layers to distill differs greatly for different FL scenarios; the best-performing sub-model is dependent on the client's dataset. For instance, the largest sub-model might not be the best performing as it is only trained with a subset of data and the smallest sub-model might be insufficiently large to learn good representations of the dataset.

This paper proposes a different depth-based scaling early exit approach to counteract existing limitations better. Unlike previous works, we use a lightweight transformer-based recurrent early exit module \method{} which is shared across all sub-models. \method{} is trained to \textit{1)} exploit and fuse features from multiple sub-models for task-based prediction, allowing us to use a single shared classifier, and \textit{2)} modulate the features for deeper sub-models to yield better representations. Learning to perform task-based prediction using a shared classifier on aggregated features of multiple sub-models allows deeper sub-models to leverage earlier sub-models' features. Additionally, as the classifier is trained on the full dataset, it overcomes previous limitations where deeper classifiers are trained on partial data. To pick the right teacher sub-model for knowledge distillation, we propose using the estimated best-performing sub-model per client to distil knowledge to the other sub-models. This adaptive best exit selection can also potentially induce computation savings during inference as each client picks its best-performing exit rather than its deepest exit. Our contributions are summarized as follows:

\begin{itemize}[leftmargin=*,noitemsep,topsep=0pt]

  \item {We present a novel approach to tackle heterogeneous clients in FL where representations of different sub-models are implicitly leveraged for both early exiting and feature learning for deeper sub-models. }

  \item {We propose a dynamic way to select the best-performing sub-model as the teacher model for knowledge distillation for each client.}

  \item {Through our experiments, we show that our approach, \method{}, is scalable by accommodating a diverse range of client resources while consistently outperforming state-of-the-art baselines in both full federated fine-tuning~\cite{qu2022rethinking,nguyen2022begin,chen2023importance} and emerging federated parameter-efficient fine-tuning (PEFT)~\cite{sun2022conquering, zhang2023fedpetuning,zhao2023breaking} scenarios on standard image and speech benchmarks. Lastly, our comprehensive ablation studies help to elucidate the contributions of knowledge distillation and different federated aggregation strategies.}

\end{itemize}

\section{Related Work}


\noindent\textbf{Federated Learning.} 
FL research spans a wide range of problems and challenges such as privacy, fairness, communication, personalization, and many more. More details can be found in surveys~\cite{kairouz2021advances,zhang2021survey,wen2023survey}. In this paper, we focus on the learning of a global model, first proposed in FedSGD~\citep{fedsgd} and made popular when FedAvg~\citep{fedavg} was introduced. Subsequent works aim to optimize the global model performance by better handling the data heterogeneity among clients. For instance, existing works finetune the global model with IID data~\cite{emd}, use regularizers to minimize the Euclidean distance~\cite{li2020federated} or maximize the agreement~\cite{moon} between the global and local models, shift the local models to alleviate its divergence with the global model~\cite{karimireddy2020scaffold}, or perform exact minimization such that local models converge to a stationary point of the global loss~\cite{feddyn}.


\noindent\textbf{Transformers.} 
Besides algorithmic changes, researchers also found that self-attention-based architectures such as Transformers~\citep{vaswani2017attention} are better than convolution-based models at handling data heterogeneity~\citep{qu2022rethinking}. Additionally, starting from pre-trained models as opposed to random model initialization plays a key role in stabilizing federated training and improving performance~\cite{nguyen2022begin,chen2023importance}. More recently, recent FL works utilize Parameter-Efficient Fine-Tuning (PEFT) methods to substantially reduce the communication cost with little or no degradation in performance~\cite{sun2022conquering,zhang2023fedpetuning,zhao2023breaking}. We thus extend the conventional FL experiments to incorporate the PEFT training in our evaluation.

\noindent\textbf{System Heterogeneity in FL.}
Towards handling client heterogeneity, where participants have different hardware resources, 
researchers have leveraged various techniques such as \textit{quantization}~\cite{yoon2022bitwidth}, \textit{pruning}~\citep{caldas2018expanding,jiang2022model}, low-rank decomposition~\citep{yao2021fedhm}, \textit{neural architecture search}~\cite{fedoras2022}, \textit{client selection}~\cite{oort2021}, \textit{asynchronous aggregation}~\cite{huba2022papaya}, or simply varying the \textit{number of local epochs}~\cite{nguyen2022begin,wang2020tackling,royson2023fedl2p}. Orthogonal to these approaches, but more closely to our setting, many works proposed to \textit{partition} the global model by \textit{width}~\cite{diao2021heterofl,horvath2021fjord,hong2022efficientsplitmix,mei2022resource} or \textit{depth}~\cite{kim2023depthfl,liu2022inclusivefl} or both width and depth~\cite{kang2023nefl,ilhan2023scalefl}. 
These approaches underperform due to the joint training of several classifiers, which are often competing with one another, along with a partial view over the dataset and selection of a sub-optimal teacher sub-model for knowledge distillation~\cite{kd2015hinton,zhang2019your,phuong2019distillation}. 
In contrast, our method, \method, learns to leverage representations across sub-models for task-based prediction on a single shared classifier. The use of a shared classifier also allows us to overcome the limitation of previous depth-based scaling works where deeper classifiers are trained with insufficient data. Additionally, we dynamically define the teacher-student combination, based on their per-client performance on the downstream task. 

\section{Proposed Method}

\begin{figure*}[ht]
     \centering 
     \includegraphics[width=0.95\linewidth]{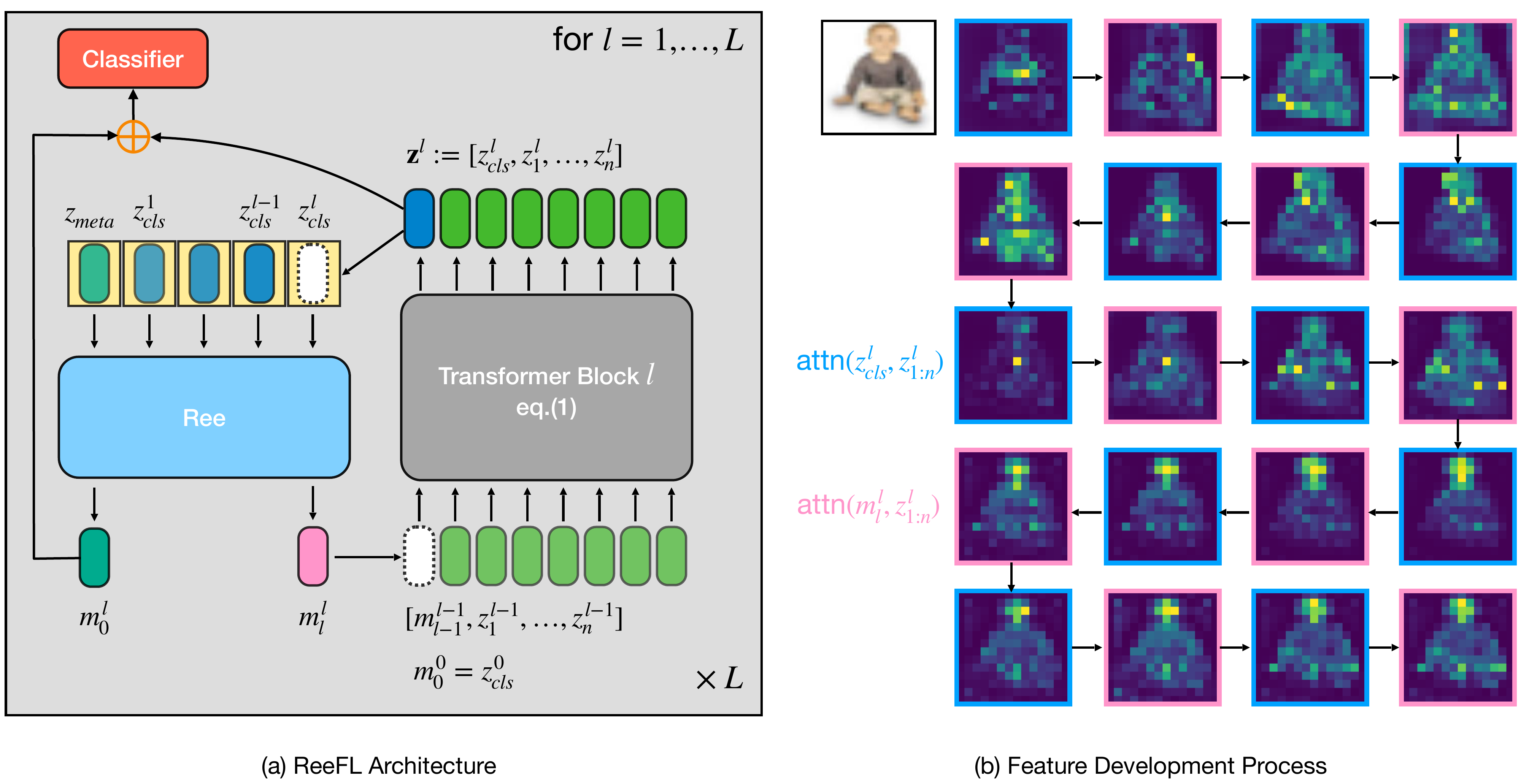}
    \caption{ \textbf{Overview of ReeFL.}
    (a) Early exiting of block $l$: \module{} takes as input the meta class token $z_{\text{meta}}$, the history of class tokens, [$z_{\text{cls}}^1 \ldots z_{\text{cls}}^{l-1}]$, and the most recent class token, $z_{\text{cls}}^l$, and produces two tokens: \textit{1)} the modulated meta-class token $m_0^l$ which participates in early-exit classification and \textit{2)} the modulated latest class token $m_l^l$ which is used to replace $z_{\text{cls}}^l$ as a part of input to block $l+1$. Assuming the case where there is an early exit after every block, the forward pass involves running the shown architecture $L$ times with shared \module{} module. We assume $m_0^0 \equiv z_{\text{cls}}^0$ to be the starting point.
    (b) We visualize \module{}'s feature modulation during the forward pass of a CIFAR-100 image by showing a sequence of attention maps. Starting from block $l=1$, we show the attention map between $z_{\text{cls}}^l$ vs. $z_{1:n}^l$ (in blue) and the attention map between $m_l^l$ and $z_{1:n}^l$ (in pink) alternatively. In this particular example of an image of a baby, the distinctive feature is the face, as learnt by later layers. In the earlier layers, particularly the 2nd, 3rd, and 4th layers, the modulated class token shown in pink aids the backbone model to focus more on the distinguishing parts of the image as compared to the use of the unmodulated class token shown in blue. The figure hence offers some interpretability as to how \module{}’s feature modulation affects the self-attention module in the backbone model especially in the earlier layers.
    }
    \label{fig:overview}
\end{figure*}

\subsection{Preliminaries}


\textbf{Transformers with Early Exits.}~We primarily focus on the classical Transformers~\cite{vaswani2017attention, dosovitskiy2020image} due to their surge in popularity and also growing evidence of their capacity to better handle heterogeneous data in FL~\cite{qu2022rethinking}. A Transformer model typically consists of:
\begin{itemize}[leftmargin=*,noitemsep,topsep=0pt]
    \item A tokenizer that maps each element of the input sequence to a $d$-dimensional vector (aka token).
    \item A learnable positional embedding for each position $i$ is added to the outcome of tokenization, which yields the $i$-th token denoted by $z_i^0$.
    \item A learnable class token: $z_{\text{cls}}^0 \in \mathbb{R}^d$, where the superscript $0$ indicates it is located at the entrance.
    \item A stack of architecturally identical blocks that apply multi-head self-attention (MSA), layer normalization (LN) and feed-forward multilayer perceptron (MLP) transformations on the initial sequence of tokens $\mathbf{z}^0$. 
\end{itemize}
The forward pass of each Transformer block, for $l=1\cdots L$, takes the form:
{\small
\begin{align}
        &\mathbf{z}^l = \bar{\mathbf{z}} + \text{MLP}^l(\text{LN}_2^l(\bar{\mathbf{z}})) \notag \\
        &\bar{\mathbf{z}} = \mathbf{z}^{l-1} + \text{MSA}^l(\text{LN}_1^l(\mathbf{z}^{l-1}))
        \label{eq:transformer}
\end{align}
}%
where $\mathbf{z}^l := [z_{\text{cls}}^l, z_1^l, \ldots, z_n^l]$ is the output from block $l$, yielding the $l$-th transformation of $\mathbf{z}^0$. Since none of the transformations change the shape of the intermediate representation $\mathbf{z}^l$, it can be fed into a classifier for early exiting.

\textbf{Federated Learning with Heterogeneous Clients.}~We consider a centralized setting of FL where the central server holds a global transformer model $\theta_g$ with sub-models $\theta_g[:l]$, for $l=1\cdots L$, representing $L$ early exits.
Each client, $i$, has a device with a maximum resource budget of $r_i$
and $N_i$ private training examples $\{(x_j, y_j)\}_{j=1}^{N_i}$ with $x_j$ the input sample and $y_j$ the target label. For each round, $c$ participating clients are randomly sampled from a pool of $C$ clients and the corresponding sub-models $\theta_i := \theta_g[:r_i]$ are selected to fit into each client's budget. These sub-models are then sent to the respective clients for training (i.e., one or few sub-model updating steps) and the updated sub-models are sent back to the server for central aggregation. Existing aggregation strategies such as FedAvg~\cite{fedavg}, FedAdam~\cite{fedadam}, FedDyn~\cite{feddyn} can be naturally adapted for the early-exit setting.

\subsection{Recurrent Early Exits}\label{sec:reefl}

Given the heterogeneous setting, there are two main challenges for learning early-exiting sub-models: i) shallow sub-models face a dilemma of choosing between modulating subsequent features for later predictions or exploiting current features for prediction at early exits; ii) deeper sub-models are likely overfitted because they are only affordable on high-capacity clients. Prior work attempts to mitigate these issues via knowledge distillation, typically hand-picking the largest sub-model as the teacher~\cite{ilhan2023scalefl} or utilizing deep mutual learning~\cite{zhang2018deep} where each sub-model learned from one another~\cite{kim2023depthfl} to counteract these limitations.

\textbf{Architecture of \method{}.}~We propose a single Transformer-based block module, named \emph{recurrent early exits (\module{})} with parameters $\phi$, to facilitate fine-tuning of early-exiting sub-models based on a pre-trained backbone transformer. As suggested by its name, \module{} is applied recurrently at each early exit to simultaneously achieve feature exploration and exploitation of the class token.

Specifically, as illustrated in Fig~\ref{fig:overview}, if the client is required to early exit at block $l$, \module{} appends the class token $z_{\text{cls}}^l$ to the queue of the class tokens $\mathbf{q}_{\text{cls}}[:l] := [z_{\text{meta}}, z_{\text{cls}}^1, \ldots, z_{\text{cls}}^l]$, and then transform it to obtain the ``exploitation'' and ``exploration'' versions of class token $z_{\text{cls}}^l$: 
\begin{equation}
    \mathbf{m}^l = \text{\module{}}_{\phi}(\mathbf{q}_{\text{cls}}[:l] + \mathbf{p}[:l]).
\end{equation}
The outputs $\mathbf{m}^l := [m_0^l, \ldots, m_l^l]$ are considered as modulated class tokens, where $z_{\text{meta}}$ and $\mathbf{p}$ are learnable parameters analogous to the class token and the positional embedding of the backbone transformer. 

There are numerous ways to parse the output $\mathbf{m}^l$. Empirically, we find that it is important to make the most of the foundation model and use the modulated meta-class token, $m_0^l$, as an additive modification to enhance the discriminative power of $z_{\text{cls}}^l$. Thus, we build the shared classifier (with parameters $\psi$) among all early-exit blocks as follows:
\begin{equation}
    \hat{y}^l = \text{Classifier}_{\psi}(m_0^l + z_{\text{cls}}^l).
\end{equation}

To improve the feature representation of deeper layers for subsequent prediction, \module{} modulates the features of the backbone transformer by simply replacing
\begin{equation}
\label{eq:mod_cls_token}
    z_{\text{cls}}^l \leftarrow m_l^l.
\end{equation}
As such, the input to the next transformer block of the backbone is slightly changed: $[z_{\text{cls}}^l, z_1^l, \ldots, z_n^l] \rightarrow [m_l^l, z_1^l, \ldots, z_n^l]$. We show in Fig~\ref{fig:overview}(b) an example of the modulated feature development process, which alternates between the original backbone transformer blocks and \module{}. 

\textbf{Federated Training of \method{}.}~In each federated round, for each client, we train $\theta_i$ which consists of both $\phi$ and $\psi$ on the local train dataset using SGD. Note that in the case where the backbone model is frozen, then the learnable parameters of $\theta_i$ will be exactly $\phi$ and $\psi$. Following previous works~\cite{kim2023depthfl,ilhan2023scalefl}, each client, $i$, aims to minimize the following loss:
\begin{equation}
    \mathcal{L}_i = \sum_{e=1}^{E_{r_i}} \mathcal{L}_e^{ce} + \eta \mathcal{L}^{kl}
\end{equation}
where $E_{r_i}$ is the maximum exit within the client's budget, $\mathcal{L}_e^{ce}=\frac{1}{N_i}\sum^{N_i}_{j=1} \mathcal{L}^{ce}(\hat{y}^e_j,y_j)$ is the cross-entropy (CE) loss of the $e$-th exit, $\mathcal{L}^{kl}$ is the Kullback–Leibler (KL) loss denoting knowledge transfer between sub-models with its corresponding hyperparameter $\eta$. 
At the end of each round, participated clients send their locally trained parameters back to the server where these parameters will be aggregated using FedAvg~\cite{fedavg}. As the clients have different resource budgets, we weight the overlapping parameters accordingly to the number of data samples used to train each parameter. 

\textbf{\method{}'s Knowledge Distillation.}~In many prior works, the teacher and student sub-models in $\mathcal{L}^{kl}$ are manually defined, \textit{e.g.} from a bigger model to a smaller model~\cite{horvath2021fjord,ilhan2023scalefl} or the use of deep mutual learning~\cite{zhang2018deep} where all sub-models learn from one another~\cite{kim2023depthfl}. A major limitation of these approaches is that the teacher sub-model selected might be under-performing, especially in depth-scaling methods where the deepest sub-model is trained with partial data and the shallowest sub-model might not be sufficiently parameterized to learn good representations. The best-performing sub-model is hence dependent on the client, the sub-model, and the dataset on which it is optimized. Hence, instead of manually picking a teacher sub-model, \method{} uses the best training loss per-client to select the teacher sub-model and compute the KL loss as follows:
\begin{equation}    
    \mathcal{L}^{kl} = \frac{1}{N_i} \sum^{E_{r_i}}_{e \neq \tilde{e}} \sum^{N_i}_{j=1} (
    \sigma(\frac{\hat{y}_{j}^{\tilde{e}}}{\tau}) 
    \log \frac{\sigma(\frac{\hat{y}^{\tilde{e}}_j}{\tau})}                  {\sigma(\frac{\hat{y}_{j}^e}{\tau})}
    ) \tau^{2}
\end{equation}
where $\tilde{e} = \argmin \bar{\mathcal{L}^{ce}} = \argmin [\mathcal{L}_1^{ce}, \ldots, \mathcal{L}_{E_{r_i}}^{ce}]$ is the exit with the lowest training CE loss, $\sigma$ is the softmax function and $\tau$ is the temperature.
In practice, as the training loss per mini-batch of data is noisy, we take the running estimate of the training loss per client: $\bar{\mathcal{L}^{ce}} = (1 - \zeta) * \bar{\mathcal{L}^{ce}} + \zeta * \bar{\mathcal{L}^{ce}_{new}}$ where $\zeta$ is a hyperparameter and $\bar{\mathcal{L}^{ce}_{new}}$ is the training CE loss of all exits within budget computed on a new mini-batch of data.

\section{Evaluation} \label{sec:exp}

\subsection{Experimental Setup}\label{sec:setup}

\subsubsection{Datasets}

We conduct experiments on classification tasks using standard FL vision \& speech benchmarks of differing degree of data heterogeneity in both feature and label distributions\footnote{Code is available at https://github.com/royson/reefl.}. 

\textbf{CIFAR-100~\citep{cifar}.} We use the default partitions for train and test. Following prior works~\citep{karimireddy2020scaffold,wang2020federated}, we set the number of clients to $100$ and partition the data using the latent Dirichlet allocation (LDA) method: $y \sim Dir(\alpha)$ for each client. Hence, the lower the $\alpha$, the greater the degree of data heterogeneity in label distributions.

\textbf{FEMNIST~\citep{leaf}.} We use the LEAF benchmark~\citep{leaf}'s natural partition, each client corresponds to its own handwriting, hence non-IID in both feature and label distributions. We use a total of $381$ clients.

\textbf{SpeechCommandV2~\citep{speechcommands}.} We adopt the setup from~\citet{royson2023fedl2p}, sample $250$ speakers from the training set, and split each speaker's data into $80\%$/$20\%$ train/test sets. However, instead of adopting the simpler 12-classes version, we use the full version comprising of 35 classes. Each speaker corresponds to its own voice, resulting in a challenging setup with both non-IID features and labels. 

\begin{table*}[t]
\caption{Mean and standard deviation (SD) of the mean performance of all \textbf{4 exits} across 3 runs and the mean communication cost per round for each approach.}

\label{tab:mainres_e4}
\begin{scriptsize}
\resizebox{1.0\textwidth}{!}{\begin{minipage}{\textwidth}
\begin{center}
\begin{tabular}{|l|l|ccc|c|c|c|}
\hline
\multicolumn{1}{|c|}{\multirow{2}{*}{Finetuning}} & \multicolumn{1}{c|}{\multirow{2}{*}{Approach}} & \multicolumn{3}{c|}{CIFAR-100} & \multirow{2}{*}{FEMNIST} & \multirow{2}{*}{SpeechCmds} & \multirow{2}{*}{\begin{tabular}[c]{@{}c@{}}Comm.\\ Cost (MB)\end{tabular}} \\ \cline{3-5}
\multicolumn{1}{|c|}{} & \multicolumn{1}{c|}{} & \multicolumn{1}{c|}{$\alpha$=1000} & \multicolumn{1}{c|}{$\alpha$=1.0} & $\alpha$=0.1 &  &  &  \\ \hline
\multirow{5}{*}{Full} & ExclusiveFL & \multicolumn{1}{c|}{67.29±0.06} & \multicolumn{1}{c|}{66.66±0.18} & 60.91±0.11 & 84.07±0.09 & 73.69±0.05 & 53.31 \\
 & InclusiveFL & \multicolumn{1}{c|}{61.04±0.03} & \multicolumn{1}{c|}{60.92±0.26} & 54.97±0.41 & 84.3±0.05 & 77.87±1.0 & 52.33 \\
 & ScaleFL & \multicolumn{1}{c|}{57.84±0.1} & \multicolumn{1}{c|}{56.83±0.06} & 48.35±0.05 & 82.56±0.09 & 71.96±0.09 & 34.91 \\
 & DepthFL & \multicolumn{1}{c|}{57.52±0.4} & \multicolumn{1}{c|}{55.25±0.03} & 45.79±0.27 & 81.24±0.37 & 78.44±0.38 & 53.31 \\
 & ReeFL (ours) & \multicolumn{1}{c|}{\textbf{76.42±0.12}} & \multicolumn{1}{c|}{\textbf{75.69±0.17}} & \textbf{72.58±0.33} & \textbf{86.13±0.08} & \textbf{84.47±0.26} & 53.98 \\ \hline
\multirow{5}{*}{Frozen} & ExclusiveFL & \multicolumn{1}{c|}{48.02±0.03} & \multicolumn{1}{c|}{47.04±0.08} & 41.2±0.08 & 48.09±0.03 & 17.26±0.19 & 1.12 \\
 & InclusiveFL & \multicolumn{1}{c|}{53.99±0.03} & \multicolumn{1}{c|}{53.23±0.02} & 49.08±0.01 & 63.0±0.02 & 26.53±0.03 & 0.15 \\
 & ScaleFL & \multicolumn{1}{c|}{28.5±0.04} & \multicolumn{1}{c|}{27.9±0.01} & 25.18±0.04 & 28.51±0.03 & 10.8±0.1 & 0.89 \\
 & DepthFL & \multicolumn{1}{c|}{51.27±0.01} & \multicolumn{1}{c|}{49.07±0.05} & 28.63±0.54 & 45.49±0.84 & 24.37±0.09 & 1.12 \\
 & ReeFL (ours) & \multicolumn{1}{c|}{\textbf{67.52±0.1}} & \multicolumn{1}{c|}{\textbf{66.48±0.03}} & \textbf{61.36±0.12} & \textbf{82.33±0.04} & \textbf{66.09±0.08} & 1.79 \\ \hline
\multirow{5}{*}{LoRA} & ExclusiveFL & \multicolumn{1}{c|}{67.46±0.03} & \multicolumn{1}{c|}{66.4±0.03} & 58.78±0.06 & 82.98±0.09 & 69.93±0.1 & 2.53 \\
 & InclusiveFL & \multicolumn{1}{c|}{69.37±0.02} & \multicolumn{1}{c|}{69.03±0.07} & 63.55±0.13 & 83.76±0.09 & 76.79±0.1 & 1.56 \\
 & ScaleFL & \multicolumn{1}{c|}{45.68±0.04} & \multicolumn{1}{c|}{44.6±0.07} & 37.02±0.02 & 70.69±0.02 & 43.04±0.2 & 2.00 \\
 & DepthFL & \multicolumn{1}{c|}{71.56±0.29} & \multicolumn{1}{c|}{70.18±0.27} & 64.33±0.05 & 82.35±0.21 & 77.15±0.13 & 2.53 \\
 & ReeFL (ours) & \multicolumn{1}{c|}{\textbf{73.9±0.03}} & \multicolumn{1}{c|}{\textbf{73.37±0.07}} & \textbf{69.29±0.1} & \textbf{84.69±0.03} & \textbf{79.91±0.08} & 3.20 \\ \hline
\multirow{5}{*}{PA} & ExclusiveFL & \multicolumn{1}{c|}{67.42±0.13} & \multicolumn{1}{c|}{66.48±0.19} & 59.0±0.04 & 82.35±0.16 & 69.82±0.02 & 2.55 \\
 & InclusiveFL & \multicolumn{1}{c|}{66.81±0.04} & \multicolumn{1}{c|}{66.57±0.11} & 61.2±0.31 & 83.12±0.0 & 76.47±0.08 & 1.58 \\
 & ScaleFL & \multicolumn{1}{c|}{46.48±0.01} & \multicolumn{1}{c|}{45.16±0.06} & 38.02±0.08 & 72.83±0.04 & 43.53±0.24 & 2.01 \\
 & DepthFL & \multicolumn{1}{c|}{72.01±0.05} & \multicolumn{1}{c|}{70.79±0.08} & 64.92±0.23 & 82.54±0.17 & 74.46±0.24 & 2.55 \\
 & ReeFL (ours) & \multicolumn{1}{c|}{\textbf{72.33±0.08}} & \multicolumn{1}{c|}{\textbf{71.44±0.05}} & \textbf{66.92±0.04} & \textbf{84.2±0.04} & \textbf{78.51±0.34} & 3.22 \\ \hline
\multirow{5}{*}{SA} & ExclusiveFL & \multicolumn{1}{c|}{68.2±0.02} & \multicolumn{1}{c|}{67.39±0.07} & 59.71±0.09 & 82.07±0.04 & 68.98±0.2 & 2.55 \\
 & InclusiveFL & \multicolumn{1}{c|}{67.5±0.18} & \multicolumn{1}{c|}{67.01±0.14} & 61.88±0.27 & 82.57±0.11 & 75.01±0.15 & 1.58 \\
 & ScaleFL & \multicolumn{1}{c|}{41.88±0.01} & \multicolumn{1}{c|}{40.48±0.01} & 33.6±0.15 & 58.51±0.03 & 26.62±0.03 & 2.01 \\
 & DepthFL & \multicolumn{1}{c|}{\textbf{72.63±0.18}} & \multicolumn{1}{c|}{71.3±0.02} & 64.73±0.04 & 81.92±0.17 & 74.7±0.23 & 2.55 \\
 & ReeFL (ours) & \multicolumn{1}{c|}{72.15±0.07} & \multicolumn{1}{c|}{\textbf{71.48±0.1}} & \textbf{66.52±0.25} & \textbf{84.07±0.09} & \textbf{78.39±0.25} & 3.22 \\ \hline
\multirow{5}{*}{SSF} & ExclusiveFL & \multicolumn{1}{c|}{66.06±0.02} & \multicolumn{1}{c|}{65.29±0.01} & 57.94±0.04 & 79.27±0.07 & 57.03±0.46 & 1.39 \\
 & InclusiveFL & \multicolumn{1}{c|}{67.6±0.04} & \multicolumn{1}{c|}{67.35±0.02} & 62.23±0.1 & 82.11±0.02 & 71.43±0.04 & 0.40 \\
 & ScaleFL & \multicolumn{1}{c|}{40.12±0.09} & \multicolumn{1}{c|}{39.3±0.04} & 33.4±0.06 & 52.68±0.06 & 27.01±0.14 & 1.10 \\
 & DepthFL & \multicolumn{1}{c|}{45.61±0.03} & \multicolumn{1}{c|}{42.87±0.11} & 28.38±0.42 & 74.66±0.22 & 66.22±0.12 & 1.39 \\
 & ReeFL (ours) & \multicolumn{1}{c|}{\textbf{70.12±0.07}} & \multicolumn{1}{c|}{\textbf{69.54±0.02}} & \textbf{64.77±0.11} & \textbf{83.42±0.04} & \textbf{73.6±0.04} & 2.04 \\ \hline
\end{tabular}
\end{center}
\end{minipage}
}
\end{scriptsize}
\vspace{-0.5em}
\end{table*}

\begin{table*}[t]
\caption{Mean and standard deviation (SD) of the mean performance of all \textbf{12 exits} across 3 runs and the mean communication cost per round for each approach.}

\label{tab:mainres_e12}
\begin{scriptsize}
\resizebox{1.0\textwidth}{!}{\begin{minipage}{\textwidth}
\begin{center}
\begin{tabular}{|l|l|ccc|c|c|c|}
\hline
\multicolumn{1}{|c|}{\multirow{2}{*}{Finetuning}} & \multicolumn{1}{c|}{\multirow{2}{*}{Approach}} & \multicolumn{3}{c|}{CIFAR-100} & \multirow{2}{*}{FEMNIST} & \multirow{2}{*}{SpeechCmds} & \multirow{2}{*}{\begin{tabular}[c]{@{}c@{}}Comm.\\ Cost (MB)\end{tabular}} \\ \cline{3-5}
\multicolumn{1}{|c|}{} & \multicolumn{1}{c|}{} & \multicolumn{1}{c|}{$\alpha$=1000} & \multicolumn{1}{c|}{$\alpha$=1.0} & $\alpha$=0.1 &  &  &  \\ \hline
\multirow{5}{*}{Full} & ExclusiveFL & \multicolumn{1}{c|}{39.47±0.48} & \multicolumn{1}{c|}{38.26±0.21} & 31.11±0.08 & 77.07±0.06 & 64.71±0.18 & 46.39 \\
 & InclusiveFL & \multicolumn{1}{c|}{42.35±2.18} & \multicolumn{1}{c|}{44.05±0.13} & 33.42±0.43 & 80.33±0.01 & 70.93±0.25 & 45.57 \\
 & ScaleFL & \multicolumn{1}{c|}{35.63±0.09} & \multicolumn{1}{c|}{34.29±1.33} & 23.41±0.17 & 74.68±0.05 & 61.57±0.25 & 26.40 \\
 & DepthFL & \multicolumn{1}{c|}{46.66±0.05} & \multicolumn{1}{c|}{43.6±0.52} & 33.69±0.4 & 78.34±0.23 & 71.23±0.34 & 46.39 \\
 & ReeFL (ours) & \multicolumn{1}{c|}{\textbf{66.83±0.24}} & \multicolumn{1}{c|}{\textbf{65.89±0.19}} & \textbf{57.55±0.78} & \textbf{82.98±0.03} & \textbf{82.34±0.27} & 47.21 \\ \hline
\multirow{5}{*}{Frozen} & ExclusiveFL & \multicolumn{1}{c|}{39.19±0.01} & \multicolumn{1}{c|}{37.6±0.02} & 28.95±0.02 & 42.1±0.12 & 14.75±0.06 & 0.97 \\
 & InclusiveFL & \multicolumn{1}{c|}{43.53±0.03} & \multicolumn{1}{c|}{41.6±0.04} & 32.28±0.05 & 53.35±0.03 & 20.22±0.05 & 0.15 \\
 & ScaleFL & \multicolumn{1}{c|}{16.66±0.02} & \multicolumn{1}{c|}{15.4±0.01} & 11.77±0.01 & 20.98±0.05 & 7.32±0.1 & 0.70 \\
 & DepthFL & \multicolumn{1}{c|}{45.22±0.06} & \multicolumn{1}{c|}{42.93±0.18} & 25.2±1.89 & 41.29±0.1 & 21.3±0.02 & 0.97 \\
 & ReeFL (ours) & \multicolumn{1}{c|}{\textbf{58.74±0.05}} & \multicolumn{1}{c|}{\textbf{58.1±0.09}} & \textbf{51.33±0.36} & \textbf{75.36±0.02} & \textbf{59.5±0.5} & 1.79 \\ \hline
\multirow{5}{*}{LoRA} & ExclusiveFL & \multicolumn{1}{c|}{48.04±0.05} & \multicolumn{1}{c|}{46.22±0.13} & 36.34±0.18 & 74.9±0.11 & 59.1±0.08 & 2.19 \\
 & InclusiveFL & \multicolumn{1}{c|}{55.9±0.08} & \multicolumn{1}{c|}{54.54±0.15} & 41.53±0.15 & 78.21±0.15 & 66.18±0.58 & 1.37 \\
 & ScaleFL & \multicolumn{1}{c|}{35.14±0.09} & \multicolumn{1}{c|}{33.05±0.02} & 22.62±0.1 & 59.57±0.05 & 44.23±0.14 & 1.58 \\
 & DepthFL & \multicolumn{1}{c|}{62.74±0.09} & \multicolumn{1}{c|}{61.3±0.11} & 51.9±0.15 & 78.97±0.15 & 70.66±0.07 & 2.19 \\
 & ReeFL (ours) & \multicolumn{1}{c|}{\textbf{65.7±0.09}} & \multicolumn{1}{c|}{\textbf{65.01±0.03}} & \textbf{58.85±0.05} & \textbf{81.06±0.06} & \textbf{75.96±0.18} & 3.01 \\ \hline
\multirow{5}{*}{PA} & ExclusiveFL & \multicolumn{1}{c|}{45.63±0.14} & \multicolumn{1}{c|}{43.57±0.27} & 32.94±2.19 & 73.1±0.02 & 59.06±0.12 & 2.21 \\
 & InclusiveFL & \multicolumn{1}{c|}{52.27±0.12} & \multicolumn{1}{c|}{50.97±0.06} & 39.41±0.03 & 73.04±0.03 & 63.05±0.23 & 1.39 \\
 & ScaleFL & \multicolumn{1}{c|}{33.38±0.1} & \multicolumn{1}{c|}{31.48±0.04} & 21.72±0.22 & 62.0±0.05 & 44.08±0.07 & 1.60 \\
 & DepthFL & \multicolumn{1}{c|}{61.23±0.23} & \multicolumn{1}{c|}{59.8±0.07} & 6.88±5.88 & 4.95±0.2 & 65.69±0.41 & 2.21 \\
 & ReeFL (ours) & \multicolumn{1}{c|}{\textbf{62.35±0.07}} & \multicolumn{1}{c|}{\textbf{61.47±0.12}} & \textbf{55.67±0.14} & \textbf{77.62±0.04} & \textbf{72.37±0.02} & 3.03 \\ \hline
\multirow{5}{*}{SA} & ExclusiveFL & \multicolumn{1}{c|}{47.69±0.04} & \multicolumn{1}{c|}{46.18±0.19} & 31.54±0.53 & 73.06±0.05 & 58.93±0.13 & 2.21 \\
 & InclusiveFL & \multicolumn{1}{c|}{53.37±0.11} & \multicolumn{1}{c|}{52.02±0.06} & 40.06±0.1 & 72.54±0.07 & 61.8±0.19 & 1.39 \\
 & ScaleFL & \multicolumn{1}{c|}{31.33±0.07} & \multicolumn{1}{c|}{28.97±0.05} & 19.27±0.02 & 53.08±0.01 & 31.65±0.11 & 1.60 \\
 & DepthFL & \multicolumn{1}{c|}{62.66±0.09} & \multicolumn{1}{c|}{60.79±0.11} & 1.02±0.01 & 4.82±0.27 & 65.68±0.59 & 2.21 \\
 & ReeFL (ours) & \multicolumn{1}{c|}{\textbf{63.28±0.13}} & \multicolumn{1}{c|}{\textbf{61.94±0.2}} & \textbf{54.99±0.1} & \textbf{77.43±0.04} & \textbf{71.52±0.13} & 3.03 \\ \hline
\multirow{5}{*}{SSF} & ExclusiveFL & \multicolumn{1}{c|}{49.09±0.02} & \multicolumn{1}{c|}{47.69±0.08} & 18.66±1.7 & 69.43±0.03 & 54.94±0.22 & 1.21 \\
 & InclusiveFL & \multicolumn{1}{c|}{54.43±0.04} & \multicolumn{1}{c|}{52.92±0.07} & 40.63±0.01 & 74.48±0.02 & 59.6±0.09 & 0.37 \\
 & ScaleFL & \multicolumn{1}{c|}{30.01±0.01} & \multicolumn{1}{c|}{27.99±0.05} & 19.24±0.03 & 46.63±0.02 & 25.17±0.12 & 0.87 \\
 & DepthFL & \multicolumn{1}{c|}{47.31±0.2} & \multicolumn{1}{c|}{44.01±0.14} & 1.82±0.71 & 5.01±0.19 & 3.72±0.44 & 1.21 \\
 & ReeFL (ours) & \multicolumn{1}{c|}{\textbf{61.85±0.02}} & \multicolumn{1}{c|}{\textbf{61.25±0.17}} & \textbf{54.42±0.13} & \textbf{78.05±0.03} & \textbf{67.44±0.24} & 2.01 \\ \hline
\end{tabular}
\end{center}
\end{minipage}
}
\end{scriptsize}
\vspace{-0.7em}
\end{table*}

\subsubsection{Model \& Client Heterogeneity}

Recent works showed that starting with a pre-trained model as opposed to a randomly initialized model leads to better stability and performance~\cite{nguyen2022begin,chen2023importance}. Hence, we start with a pre-trained model, using the smaller variant of DeiT~\citep{Touvron21deit}, DeiT-S\footnote{Experiments with other pre-trained models can be found in Appendix Section.~\ref{app:sec:more_models}}, which is pre-trained on ImageNet~\cite{russakovsky2015imagenet}. Besides training the entire model (Full), we also include freezing the backbone (Frozen) and training with a wide range of popular PEFT methods, namely Serial Adapter (SA)~\cite{houlsby2019parameter}, Parallel Adapter (PA)~\cite{he2021towards}, LoRA~\cite{hu2021lora}, and SSF~\cite{lian2022ssf}. We use LN followed by a linear layer for all classifiers. 

As previous works~\cite{diao2021heterofl,horvath2021fjord,kang2023nefl,liu2022inclusivefl,ilhan2023scalefl} typically divide the model into 3-5 submodels, we evaluate on $4$ submodels, an exit every $3$ DeiT-S blocks apart from ScaleFL which selects where to place the exits via a grid search. Additionally, we also adopt a more challenging scenario where we evaluate on $12$ submodels, an exit every block, in order to accommodate a wider range of end-devices. 
The same number of clients is allocated to each submodel, e.g. for $4$ exits, $25$ clients out of a total of $100$ are given a max budget corresponding to each exit.

\begin{table*}[!t]
\caption{Ablation study on proposed aggregation strategies and knowledge distillation of depth-based scaling methods on $4$ exits. Results on $12$ exits can be found in the Appendix.}

\label{tab:ablation1}
\begin{scriptsize}
\resizebox{1.0\textwidth}{!}{\begin{minipage}{\textwidth}
\begin{center}
\begin{tabular}{llclccccc}
\hline
\multicolumn{1}{|c|}{\multirow{2}{*}{Finetuning}} & \multicolumn{1}{c|}{\multirow{2}{*}{Approach}} & \multicolumn{1}{c|}{\multirow{2}{*}{Distillation}} & \multicolumn{1}{c|}{\multirow{2}{*}{Aggregation}} & \multicolumn{3}{c|}{CIFAR-100} & \multicolumn{1}{c|}{\multirow{2}{*}{FEMNIST}} & \multicolumn{1}{c|}{\multirow{2}{*}{SpeechCmds}} \\ \cline{5-7}
\multicolumn{1}{|c|}{} & \multicolumn{1}{c|}{} & \multicolumn{1}{c|}{} & \multicolumn{1}{c|}{} & \multicolumn{1}{c|}{$\alpha$=1000} & \multicolumn{1}{c|}{$\alpha$=1.0} & \multicolumn{1}{c|}{$\alpha$=0.1} & \multicolumn{1}{c|}{} & \multicolumn{1}{c|}{} \\ \hline
\multicolumn{1}{|l|}{\multirow{8}{*}{Frozen}} & InclusiveFL & - & \multicolumn{1}{l|}{FedAvg} & \multicolumn{1}{c|}{47.58±0.04} & \multicolumn{1}{c|}{46.47±0.1} & \multicolumn{1}{c|}{41.79±0.07} & \multicolumn{1}{c|}{49.5±0.03} & \multicolumn{1}{c|}{16.95±0.21} \\
\multicolumn{1}{|l|}{} & InclusiveFL & - & \multicolumn{1}{l|}{FedAdam} & \multicolumn{1}{c|}{53.99±0.03} & \multicolumn{1}{c|}{53.23±0.02} & \multicolumn{1}{c|}{49.08±0.01} & \multicolumn{1}{c|}{63.0±0.02} & \multicolumn{1}{c|}{26.53±0.03} \\
\multicolumn{1}{|l|}{} & DepthFL & \xmark & \multicolumn{1}{l|}{FedAvg} & \multicolumn{1}{c|}{48.88±0.04} & \multicolumn{1}{c|}{48.15±0.07} & \multicolumn{1}{c|}{45.01±0.27} & \multicolumn{1}{c|}{49.22±0.03} & \multicolumn{1}{c|}{17.16±0.34} \\
\multicolumn{1}{|l|}{} & DepthFL & \xmark & \multicolumn{1}{l|}{FedDyn} & \multicolumn{1}{c|}{51.42±0.04} & \multicolumn{1}{c|}{49.31±0.25} & \multicolumn{1}{c|}{38.27±4.84} & \multicolumn{1}{c|}{53.26±0.08} & \multicolumn{1}{c|}{24.57±0.45} \\
\multicolumn{1}{|l|}{} & DepthFL & \cmark & \multicolumn{1}{l|}{FedAvg} & \multicolumn{1}{c|}{49.38±0.03} & \multicolumn{1}{c|}{48.57±0.01} & \multicolumn{1}{c|}{44.88±0.07} & \multicolumn{1}{c|}{49.19±0.07} & \multicolumn{1}{c|}{18.45±0.11} \\
\multicolumn{1}{|l|}{} & DepthFL & \cmark & \multicolumn{1}{l|}{FedDyn} & \multicolumn{1}{c|}{51.27±0.01} & \multicolumn{1}{c|}{49.07±0.05} & \multicolumn{1}{c|}{28.63±0.54} & \multicolumn{1}{c|}{45.49±0.84} & \multicolumn{1}{c|}{24.37±0.09} \\
\multicolumn{1}{|l|}{} & ReeFL & \xmark & \multicolumn{1}{l|}{FedAvg} & \multicolumn{1}{c|}{67.04±0.09} & \multicolumn{1}{c|}{66.32±0.11} & \multicolumn{1}{c|}{61.14±0.18} & \multicolumn{1}{c|}{81.86±0.05} & \multicolumn{1}{c|}{65.9±0.03} \\
\multicolumn{1}{|l|}{} & ReeFL & \cmark & \multicolumn{1}{l|}{FedAvg} & \multicolumn{1}{c|}{\textbf{67.52±0.1}} & \multicolumn{1}{c|}{\textbf{66.48±0.03}} & \multicolumn{1}{c|}{\textbf{61.36±0.12}} & \multicolumn{1}{c|}{\textbf{82.33±0.04}} & \multicolumn{1}{c|}{\textbf{66.09±0.08}} \\ \hline
\multicolumn{1}{|l|}{\multirow{10}{*}{LoRA}} & InclusiveFL & \xmark & \multicolumn{1}{l|}{FedAvg} & \multicolumn{1}{c|}{65.78±0.01} & \multicolumn{1}{c|}{65.34±0.04} & \multicolumn{1}{c|}{58.8±0.04} & \multicolumn{1}{c|}{81.19±0.03} & \multicolumn{1}{c|}{67.24±0.06} \\
\multicolumn{1}{|l|}{} & InclusiveFL & \xmark & \multicolumn{1}{l|}{FedAdam} & \multicolumn{1}{c|}{68.04±0.12} & \multicolumn{1}{c|}{67.97±0.0} & \multicolumn{1}{c|}{61.89±0.29} & \multicolumn{1}{c|}{84.12±0.19} & \multicolumn{1}{c|}{77.81±0.1} \\
\multicolumn{1}{|l|}{} & InclusiveFL & \cmark & \multicolumn{1}{l|}{FedAvg} & \multicolumn{1}{c|}{68.18±0.04} & \multicolumn{1}{c|}{67.99±0.09} & \multicolumn{1}{c|}{61.05±0.2} & \multicolumn{1}{c|}{82.26±0.05} & \multicolumn{1}{c|}{69.0±0.15} \\
\multicolumn{1}{|l|}{} & InclusiveFL & \cmark & \multicolumn{1}{l|}{FedAdam} & \multicolumn{1}{c|}{69.37±0.02} & \multicolumn{1}{c|}{69.03±0.07} & \multicolumn{1}{c|}{63.55±0.13} & \multicolumn{1}{c|}{83.76±0.09} & \multicolumn{1}{c|}{76.79±0.1} \\
\multicolumn{1}{|l|}{} & DepthFL & \xmark & \multicolumn{1}{l|}{FedAvg} & \multicolumn{1}{c|}{71.22±0.02} & \multicolumn{1}{c|}{70.8±0.11} & \multicolumn{1}{c|}{66.35±0.14} & \multicolumn{1}{c|}{83.76±0.02} & \multicolumn{1}{c|}{74.78±0.14} \\
\multicolumn{1}{|l|}{} & DepthFL & \xmark & \multicolumn{1}{l|}{FedDyn} & \multicolumn{1}{c|}{72.89±0.1} & \multicolumn{1}{c|}{71.83±0.08} & \multicolumn{1}{c|}{66.53±0.26} & \multicolumn{1}{c|}{82.2±0.02} & \multicolumn{1}{c|}{77.49±0.29} \\
\multicolumn{1}{|l|}{} & DepthFL & \cmark & \multicolumn{1}{l|}{FedAvg} & \multicolumn{1}{c|}{69.84±0.1} & \multicolumn{1}{c|}{69.28±0.02} & \multicolumn{1}{c|}{63.31±0.07} & \multicolumn{1}{c|}{83.67±0.06} & \multicolumn{1}{c|}{74.88±0.02} \\
\multicolumn{1}{|l|}{} & DepthFL & \cmark & \multicolumn{1}{l|}{FedDyn} & \multicolumn{1}{c|}{71.56±0.29} & \multicolumn{1}{c|}{70.18±0.27} & \multicolumn{1}{c|}{64.33±0.05} & \multicolumn{1}{c|}{82.35±0.21} & \multicolumn{1}{c|}{77.15±0.13} \\
\multicolumn{1}{|l|}{} & ReeFL & \xmark & \multicolumn{1}{l|}{FedAvg} & \multicolumn{1}{c|}{73.47±0.01} & \multicolumn{1}{c|}{72.76±0.07} & \multicolumn{1}{c|}{68.61±0.2} & \multicolumn{1}{c|}{84.53±0.09} & \multicolumn{1}{c|}{79.21±0.21} \\
\multicolumn{1}{|l|}{} & ReeFL & \cmark & \multicolumn{1}{l|}{FedAvg} & \multicolumn{1}{c|}{\textbf{73.9±0.03}} & \multicolumn{1}{c|}{\textbf{73.37±0.07}} & \multicolumn{1}{c|}{\textbf{69.29±0.1}} & \multicolumn{1}{c|}{\textbf{84.69±0.03}} & \multicolumn{1}{c|}{\textbf{79.91±0.08}} \\ \hline
 &  & \multicolumn{1}{l}{} &  & \multicolumn{1}{l}{} & \multicolumn{1}{l}{} & \multicolumn{1}{l}{} & \multicolumn{1}{l}{} & \multicolumn{1}{l}{} \\
 &  & \multicolumn{1}{l}{} &  & \multicolumn{1}{l}{} & \multicolumn{1}{l}{} & \multicolumn{1}{l}{} & \multicolumn{1}{l}{} & \multicolumn{1}{l}{}
\end{tabular}
\end{center}
\end{minipage}
}
\end{scriptsize}
\vspace{-2.0em}
\end{table*}
\begin{table}[!t]
\vspace{-0.6em}
\caption{Mean Accuracy with and without ReeFL's feature modulation on $4$ exits. Results on $12$ exits can be found in the Appendix.}

\label{tab:ablation2}
\begin{scriptsize}
\resizebox{0.76\columnwidth}{!}{\begin{minipage}{\columnwidth}
\begin{center}
\begin{tabular}{|l|c|ccc|c|c|}
\hline
\multicolumn{1}{|c|}{\multirow{2}{*}{Finetuning}} & \multirow{2}{*}{Modulation} & \multicolumn{3}{c|}{CIFAR-100} & \multirow{2}{*}{FEMNIST} & \multirow{2}{*}{SpeechCmds} \\ \cline{3-5}
\multicolumn{1}{|c|}{} &  & \multicolumn{1}{c|}{$\alpha$=1000} & \multicolumn{1}{c|}{$\alpha$=1.0} & $\alpha$=0.1 &  &  \\ \hline
\multirow{2}{*}{Frozen} & \xmark & \multicolumn{1}{c|}{53.89±0.0} & \multicolumn{1}{c|}{53.24±0.02} & 44.85±0.49 & 61.01±0.02 & 20.14±0.21 \\
 & \cmark & \multicolumn{1}{c|}{\textbf{67.52±0.1}} & \multicolumn{1}{c|}{\textbf{66.48±0.03}} & \textbf{61.36±0.12} & \textbf{82.33±0.04} & \textbf{66.09±0.08} \\ \hline
\multirow{2}{*}{LoRA} & \xmark & \multicolumn{1}{c|}{73.69±0.07} & \multicolumn{1}{c|}{72.98±0.14} & 69.14±0.12 & 84.51±0.07 & 79.72±0.02 \\
 & \cmark & \multicolumn{1}{c|}{\textbf{73.9±0.03}} & \multicolumn{1}{c|}{\textbf{73.37±0.07}} & \textbf{69.29±0.1} & \textbf{84.69±0.03} & \textbf{79.91±0.08} \\ \hline
\multirow{2}{*}{PA} & \xmark & \multicolumn{1}{c|}{71.53±0.1} & \multicolumn{1}{c|}{70.83±0.11} & 66.1±0.05 & 84.03±0.03 & 77.42±0.04 \\
 & \cmark & \multicolumn{1}{c|}{\textbf{72.33±0.08}} & \multicolumn{1}{c|}{\textbf{71.44±0.05}} & \textbf{66.92±0.04} & \textbf{84.2±0.04} & \textbf{78.51±0.34} \\ \hline
\multirow{2}{*}{SA} & \xmark & \multicolumn{1}{c|}{71.6±0.01} & \multicolumn{1}{c|}{71.04±0.02} & 66.44±0.02 & 83.73±0.01 & 75.96±0.08 \\
 & \cmark & \multicolumn{1}{c|}{\textbf{72.15±0.07}} & \multicolumn{1}{c|}{\textbf{71.48±0.1}} & \textbf{66.52±0.25} & \textbf{84.07±0.09} & \textbf{78.39±0.25} \\ \hline
\multirow{2}{*}{SSF} & \xmark & \multicolumn{1}{c|}{67.32±0.02} & \multicolumn{1}{c|}{66.88±0.02} & 62.49±0.08 & 81.94±0.03 & 71.01±0.18 \\
 & \cmark & \multicolumn{1}{c|}{\textbf{70.12±0.07}} & \multicolumn{1}{c|}{\textbf{69.54±0.02}} & \textbf{64.77±0.11} & \textbf{83.42±0.04} & \textbf{73.6±0.04} \\ \hline
\end{tabular}
\end{center}
\end{minipage}
}
\end{scriptsize}
\vspace{-1.5em}
\end{table}
\begin{table*}[!t]
\caption{Injecting \module{} only at exit layers as opposed to all layers leads to similar or worse performance.}

\label{tab:ree_exit_only}
\begin{scriptsize}
\resizebox{1.0\textwidth}{!}{\begin{minipage}{\textwidth}
\begin{center}
\begin{tabular}{|l|lll|l|l|}
\hline
\multicolumn{1}{|c|}{\multirow{2}{*}{Finetuning}} & \multicolumn{3}{c|}{CIFAR-100} & \multicolumn{1}{c|}{\multirow{2}{*}{FEMNIST}} & \multicolumn{1}{c|}{\multirow{2}{*}{SpeechCmds}} \\ \cline{2-4}
\multicolumn{1}{|c|}{} & \multicolumn{1}{c|}{$\alpha$=1000} & \multicolumn{1}{c|}{$\alpha$=1.0} & \multicolumn{1}{c|}{$\alpha$=0.1} & \multicolumn{1}{c|}{} & \multicolumn{1}{c|}{} \\ \hline
Frozen & \multicolumn{1}{l|}{60.63±0.02 (-6.89)} & \multicolumn{1}{l|}{59.82±0.06 (-6.66)} & 54.58±0.02 (-6.78) & 68.93±0.03 (-13.4) & 45.42±0.07 (-20.67) \\ \hline
LoRA & \multicolumn{1}{l|}{73.86±0.04 (-0.04)} & \multicolumn{1}{l|}{73.21±0.0 (-0.16)} & 69.30±0.09 (0.01) & 84.54±0.03 (-0.15) & 79.98±0.02 (0.07) \\ \hline
PA & \multicolumn{1}{l|}{71.89±0.13 (-0.44)} & \multicolumn{1}{l|}{71.2±0.01 (-0.24)} & 66.17±0.06 (-0.75) & 84.19±0.01 (-0.01) & 77.93±0.01 (-0.58) \\ \hline
SA & \multicolumn{1}{l|}{71.66±0.02 (-0.49)} & \multicolumn{1}{l|}{71.22±0.07 (-0.26)} & 66.4±0.13 (-0.12) & 83.82±0.03 (-0.25) & 76.79±0.05 (-1.6) \\ \hline
SSF & \multicolumn{1}{l|}{68.75±0.1 (-1.37)} & \multicolumn{1}{l|}{68.09±0.03 (-1.45)} & 63.06±0.04 (-1.71) & 82.89±0.01 (-0.53) & 73.05±0.16 (-0.55) \\ \hline
\end{tabular}
\end{center}
\end{minipage}
}
\end{scriptsize}
\end{table*}

\subsubsection{Baselines}\label{sec:baseline_summary}

We compare with recent depth-based FL approaches: DepthFL~\cite{kim2023depthfl} and InclusiveFL~\cite{liu2022inclusivefl}, a recent width \& depth-based approach ScaleFL~\cite{ilhan2023scalefl}, as well as a popular naive baseline, ExclusiveFL, where clients with insufficient budget to train the full model are excluded during training. For fair comparisons, we use the same classifier architecture in all baselines. Details of each baseline can be found in Appendix Section.~\ref{app:sec:baseline_deets}. 

\begin{figure*}[h]
\centering


\begin{subfigure}{0.32\columnwidth}
    \includegraphics[trim=0 0 0 0, clip, width=0.97\columnwidth]{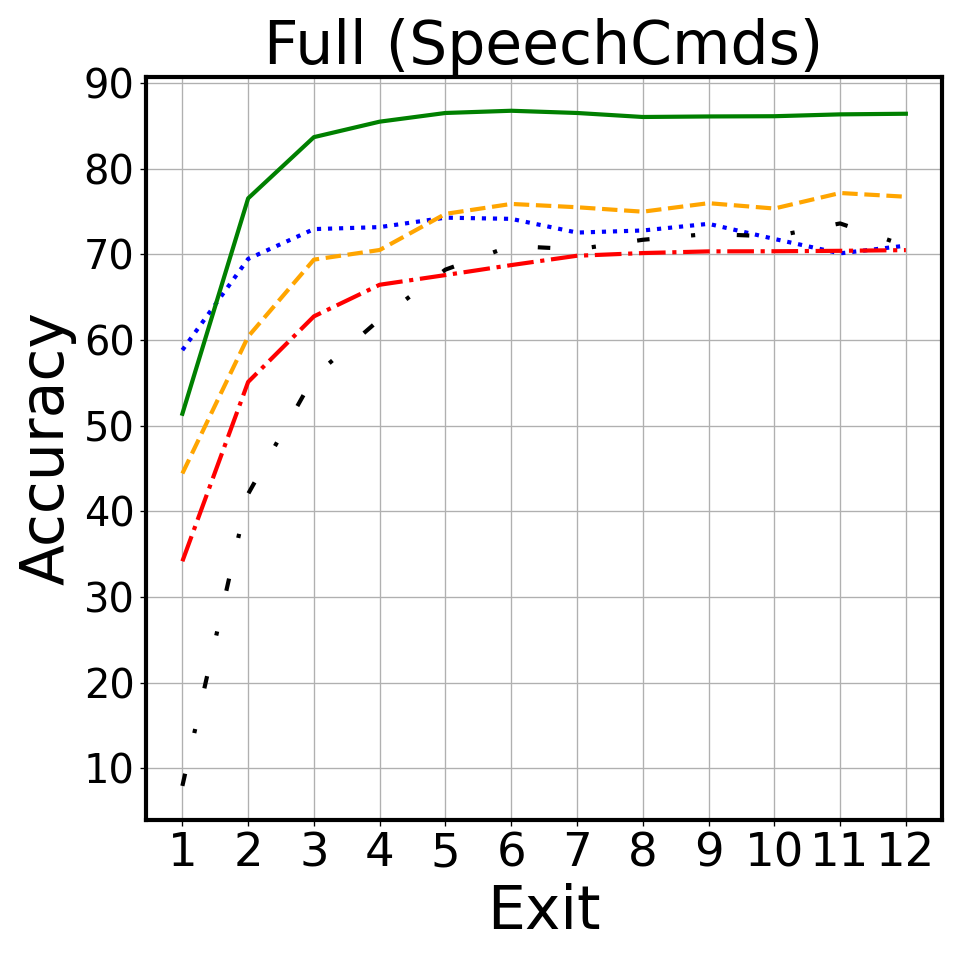}
\end{subfigure}
\begin{subfigure}{0.32\columnwidth}
    \includegraphics[trim=0 0 0 0, clip, width=0.97\columnwidth]{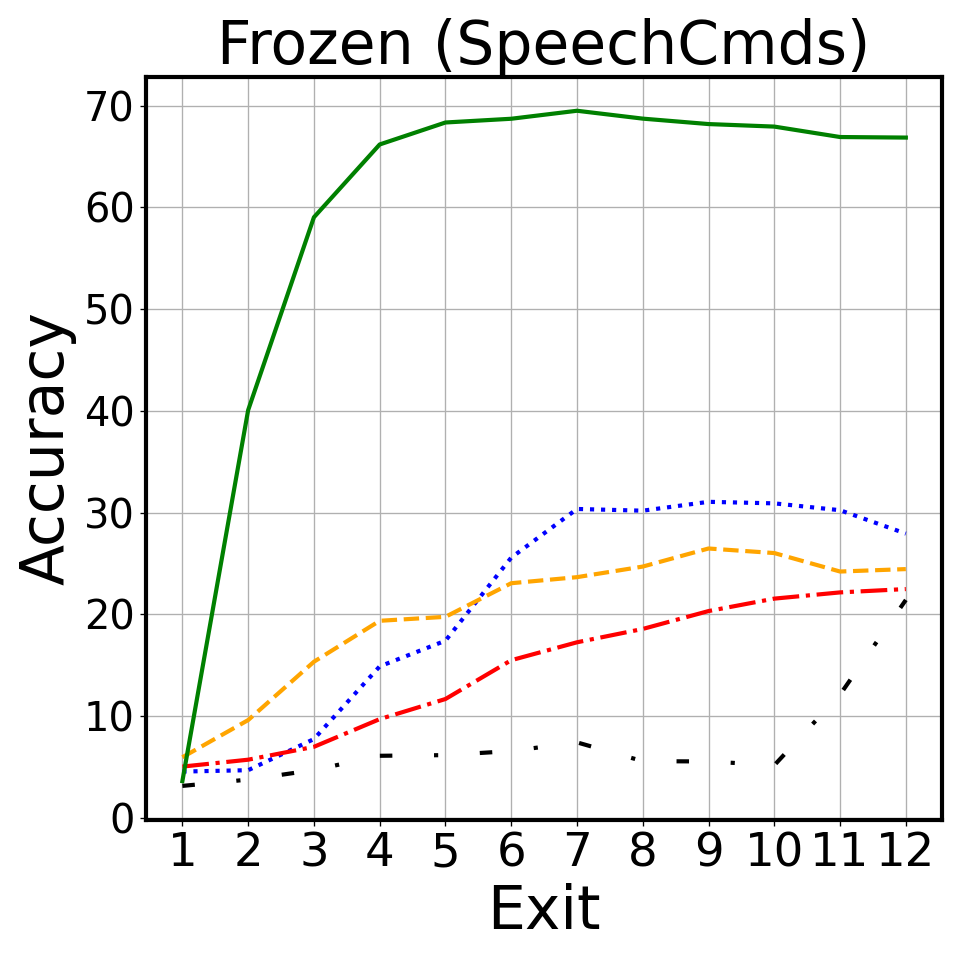}
\end{subfigure}
\begin{subfigure}{0.32\columnwidth}
    \includegraphics[trim=0 0 0 0, clip, width=0.97\columnwidth]{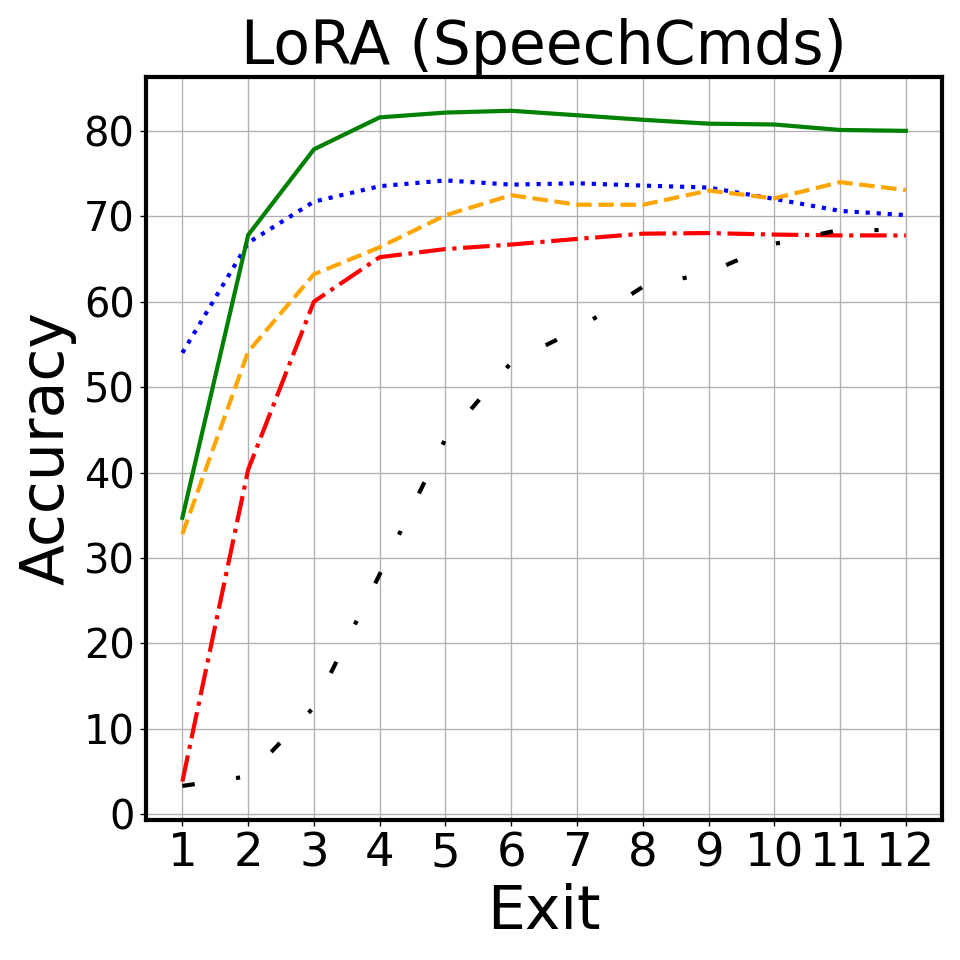}
\end{subfigure}
\begin{subfigure}{0.32\columnwidth}
    \includegraphics[trim=0 0 0 0, clip, width=0.97\columnwidth]{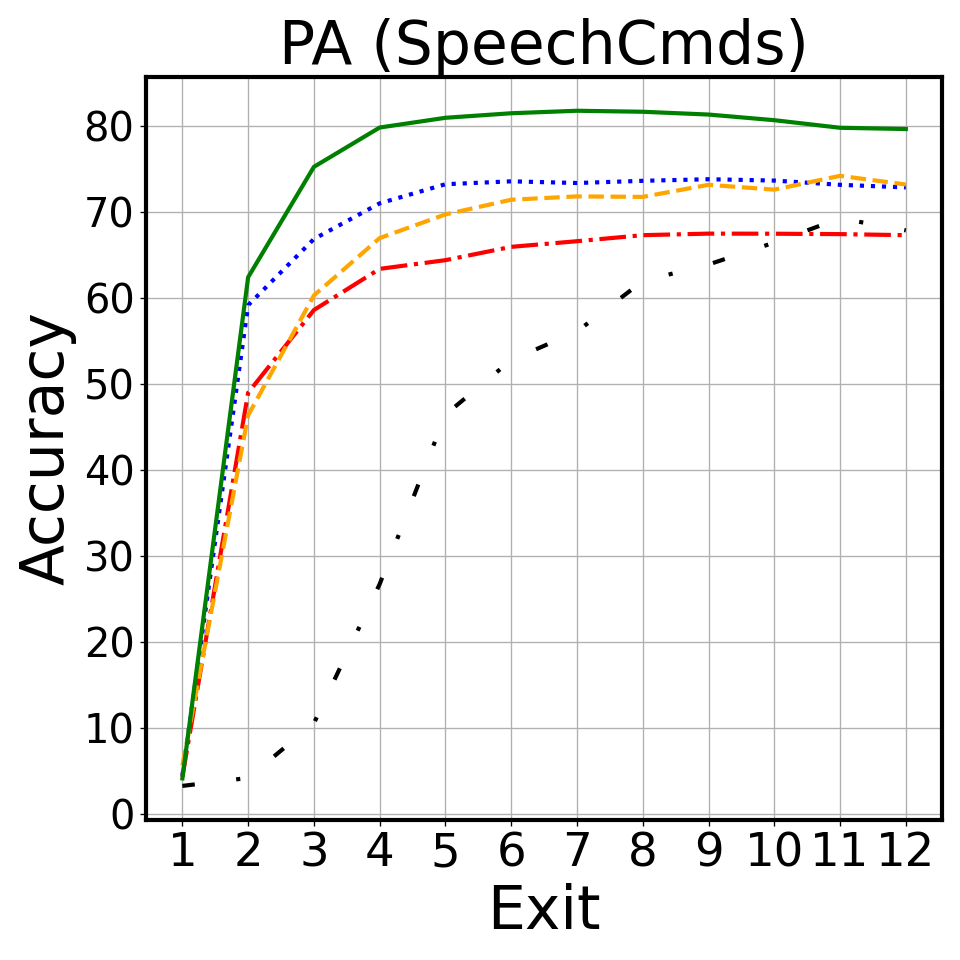}
\end{subfigure}
\begin{subfigure}{0.32\columnwidth}
    \includegraphics[trim=0 0 0 0, clip, width=0.97\columnwidth]{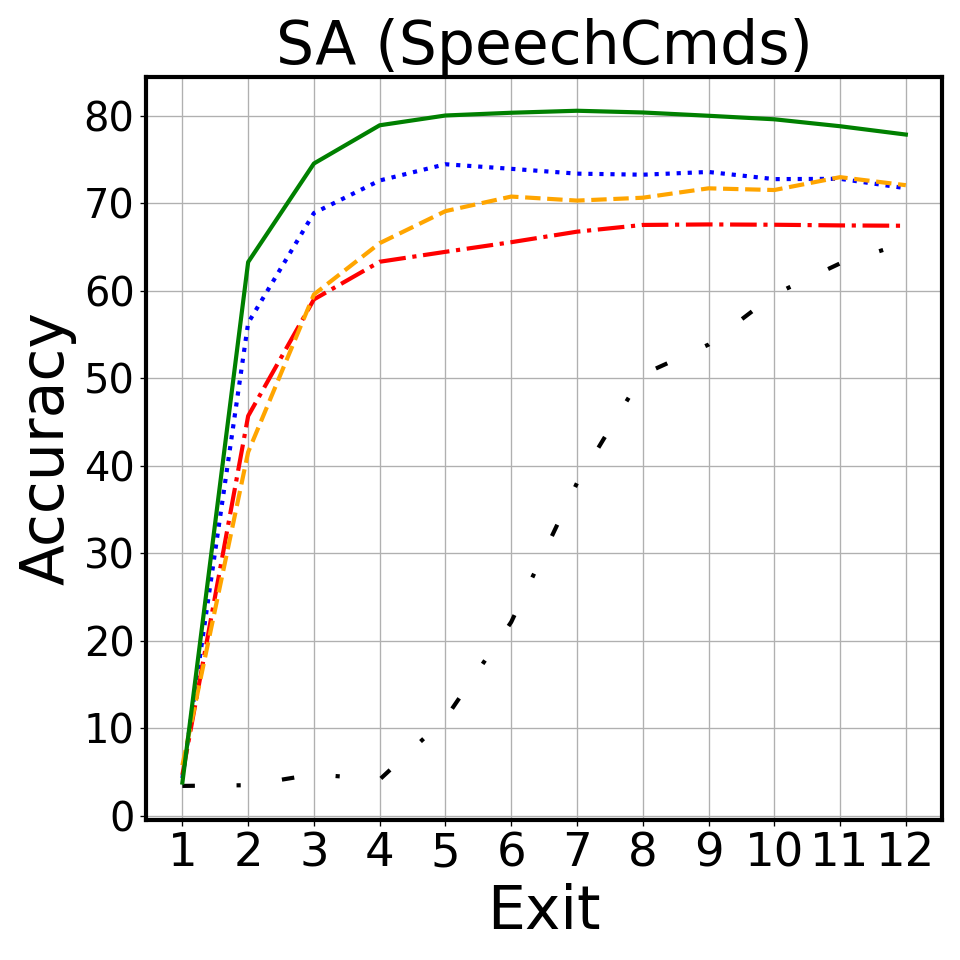}
\end{subfigure}
\begin{subfigure}{0.32\columnwidth}
    \includegraphics[trim=0 0 0 0, clip, width=0.97\columnwidth]{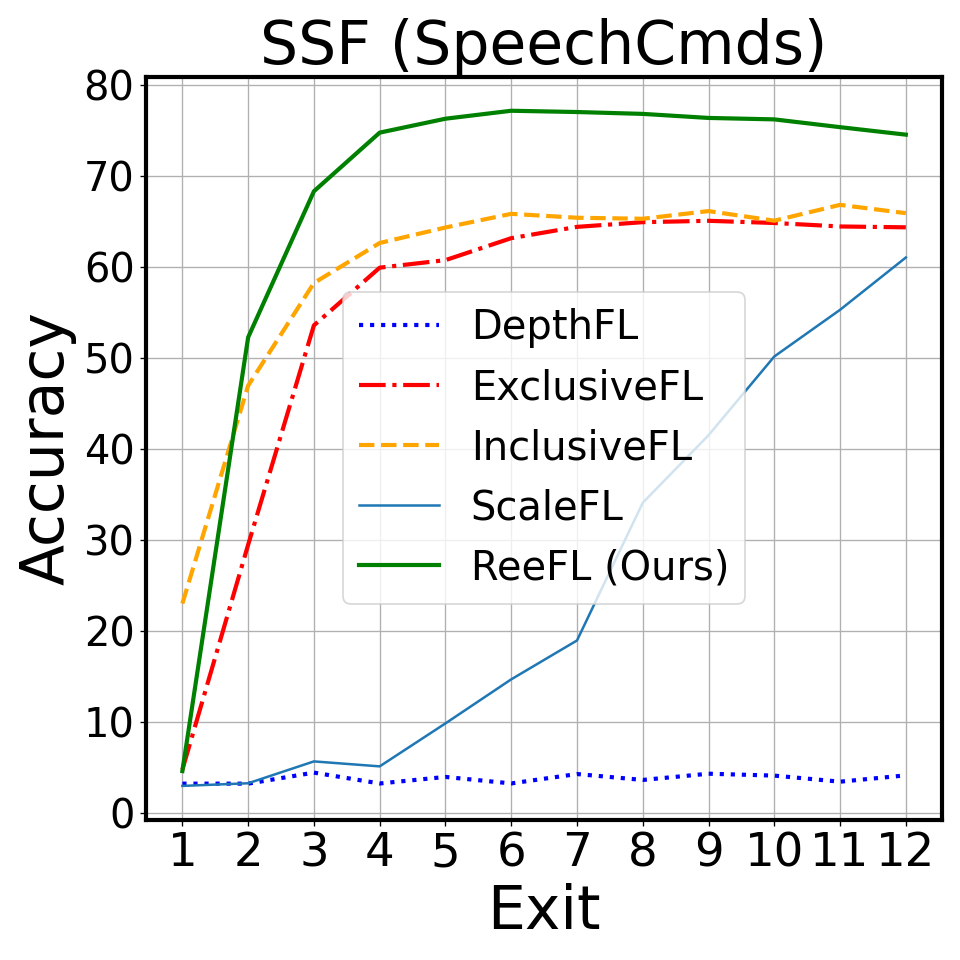}
\end{subfigure} 
\
\caption{Mean accuracy of each exit across 3 runs on SpeechCommands. More results can be found in the Appendix.}
\label{fig:ee_e12}
\end{figure*}
\begin{figure*}[h]
\centering

\begin{subfigure}{0.36\columnwidth}
    \includegraphics[trim=0 0 0 0, clip, width=0.97\columnwidth]{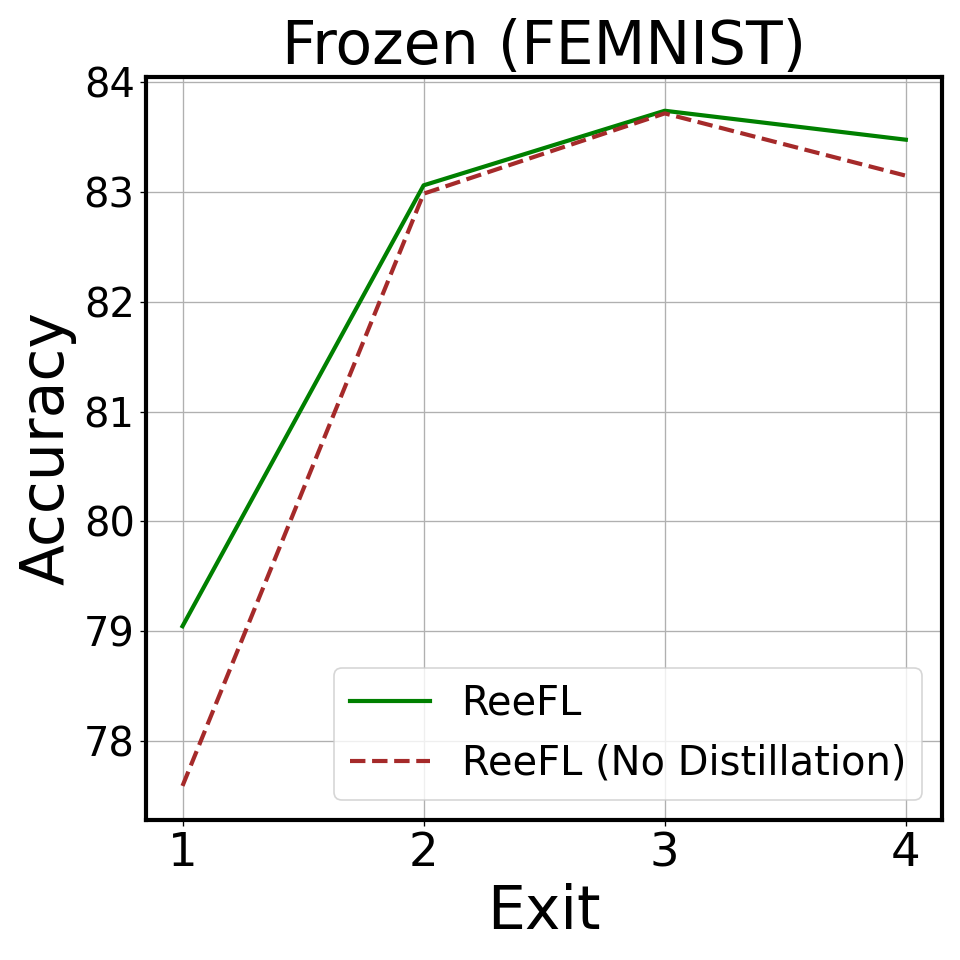}
\end{subfigure}
\begin{subfigure}{0.36\columnwidth}
    \includegraphics[trim=0 0 0 0, clip, width=0.97\columnwidth]{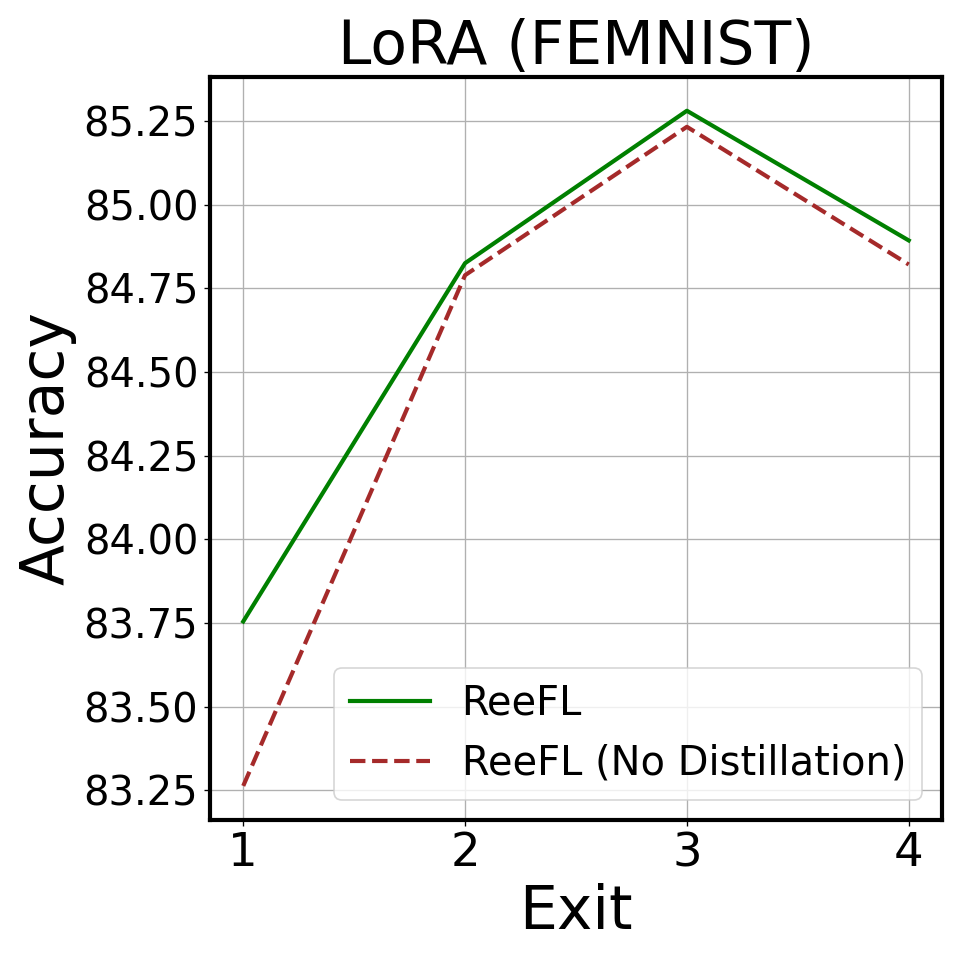}
\end{subfigure}
\begin{subfigure}{0.36\columnwidth}
    \includegraphics[trim=0 0 0 0, clip, width=0.97\columnwidth]{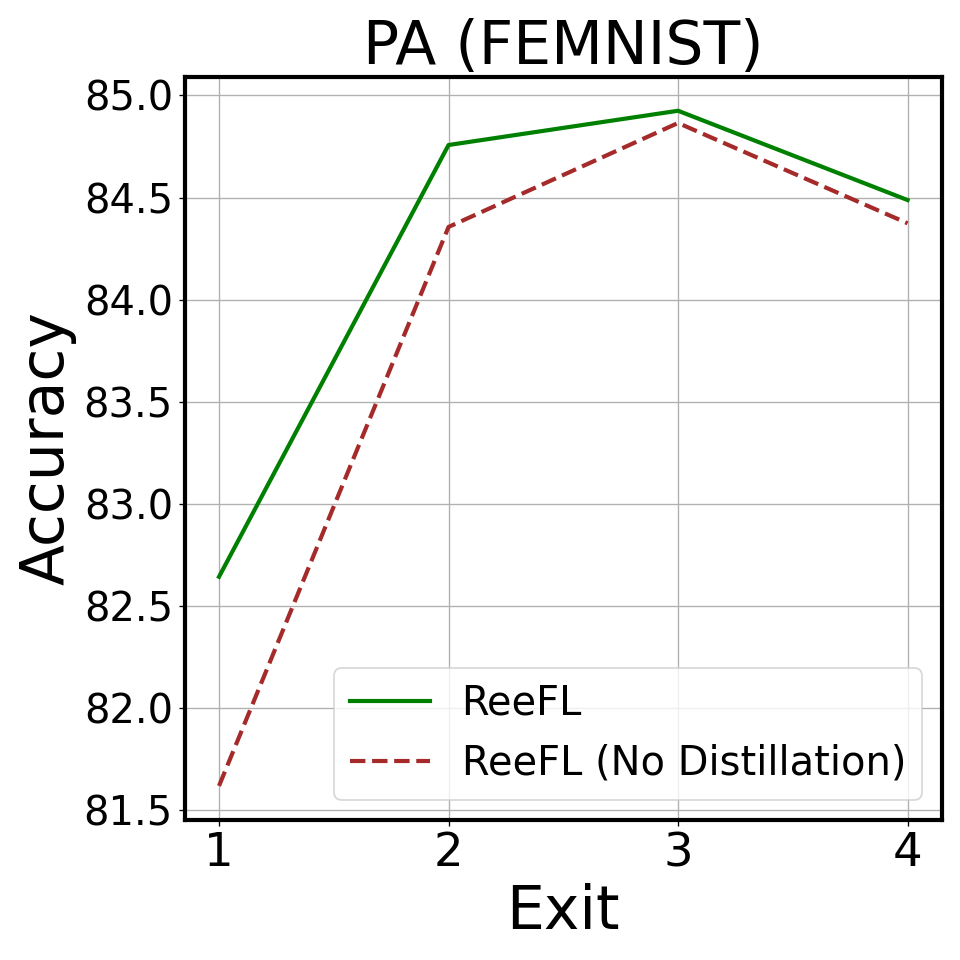}
\end{subfigure}
\begin{subfigure}{0.36\columnwidth}
    \includegraphics[trim=0 0 0 0, clip, width=0.97\columnwidth]{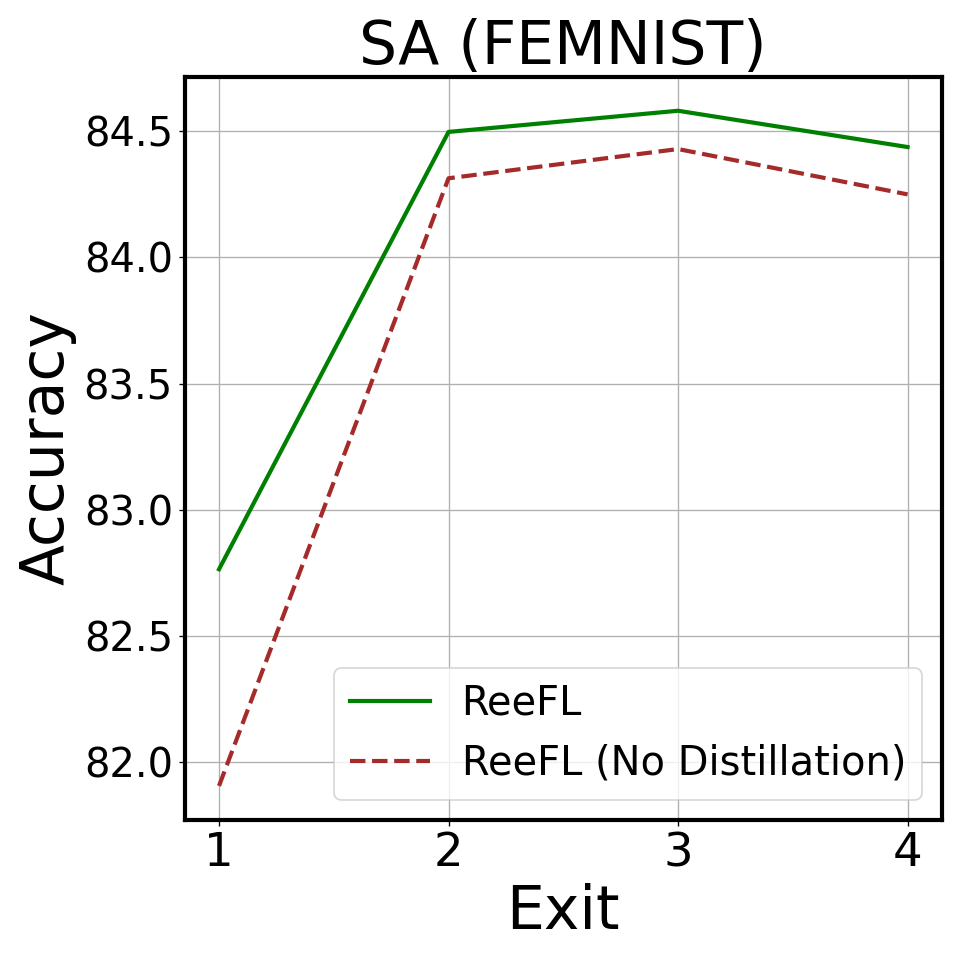}
\end{subfigure}
\begin{subfigure}{0.36\columnwidth}
    \includegraphics[trim=0 0 0 0, clip, width=0.97\columnwidth]{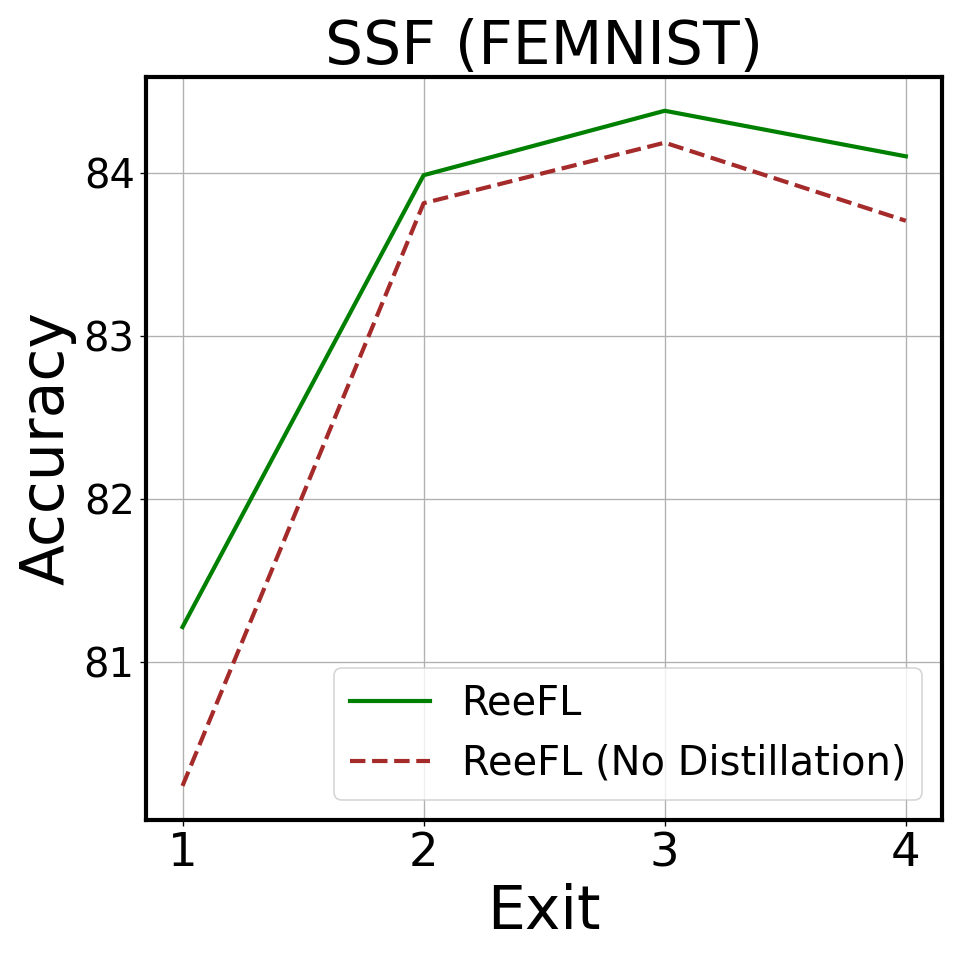}
\end{subfigure} \\


\
\vspace{-1.9em}
\caption{Impact of \method{}'s proposed knowledge distillation on FEMNIST ($4$ exits). See Appendix for more results.}
\label{fig:ablation1_ee_e4}
\vspace{-0.5em}
\end{figure*}

\subsubsection{Hyperparameters}

We run each experiment $3$ times for 1k rounds, sampling $10\%$ of the total number of clients per round, and report the mean performance of each exit, as well as the mean and standard deviation (SD) of the mean performance of all exits\footnote{We also include the ensemble performance in Appendix Section.~\ref{app:sec:ensemble}.}, on the full test dataset. 
Each sampled client in a FL round trains its local parameters with its local dataset using SGD for a single epoch using batch size of $32$. We ran a simple grid search to pick the highest performing learning rate (LR) $[1e^{-1},5e^{-2},1e^{-2},5e^{-3},1e^{-3}]$, weight decay $[0,1e^{-2},1e^{-3},1e^{-4}]$, minimum LR after LR decay using a cosine annealing LR schedule $[1e^{-2},1e^{-3},1e^{-4},1e^{-5}]$, for each baseline. More details, along with the hyperparameters of different baselines, aggregation, and PEFT methods, can be found in Appendix Section~\ref{app:sec:training_deets}. 


\subsection{Comparison with Baselines}\label{sec:main_performance}

\textbf{Performance Evaluation.}~Tables~\ref{tab:mainres_e4} \& \ref{tab:mainres_e12} show the mean and SD of the mean test accuracy of all 4 \& 12 exits respectively across 3 runs. \method{} outperforms all baselines in all scenarios except one particular scenario where we use SA for finetuning with the CIFAR-100 dataset on clients with IID data ($\alpha=1000$) for $4$ exits.
\method{} also outperforms baselines in most exits as shown in Fig.~\ref{fig:ee_e12} and Appendix Fig.~\ref{app:fig:ee_e4} \& \ref{app:fig:ee_e12} where we show the accuracy for each exit for both $4$ and $12$ exits scenarios.

\begin{figure*}[t]
    \centering
    \includegraphics[width=.93\textwidth]{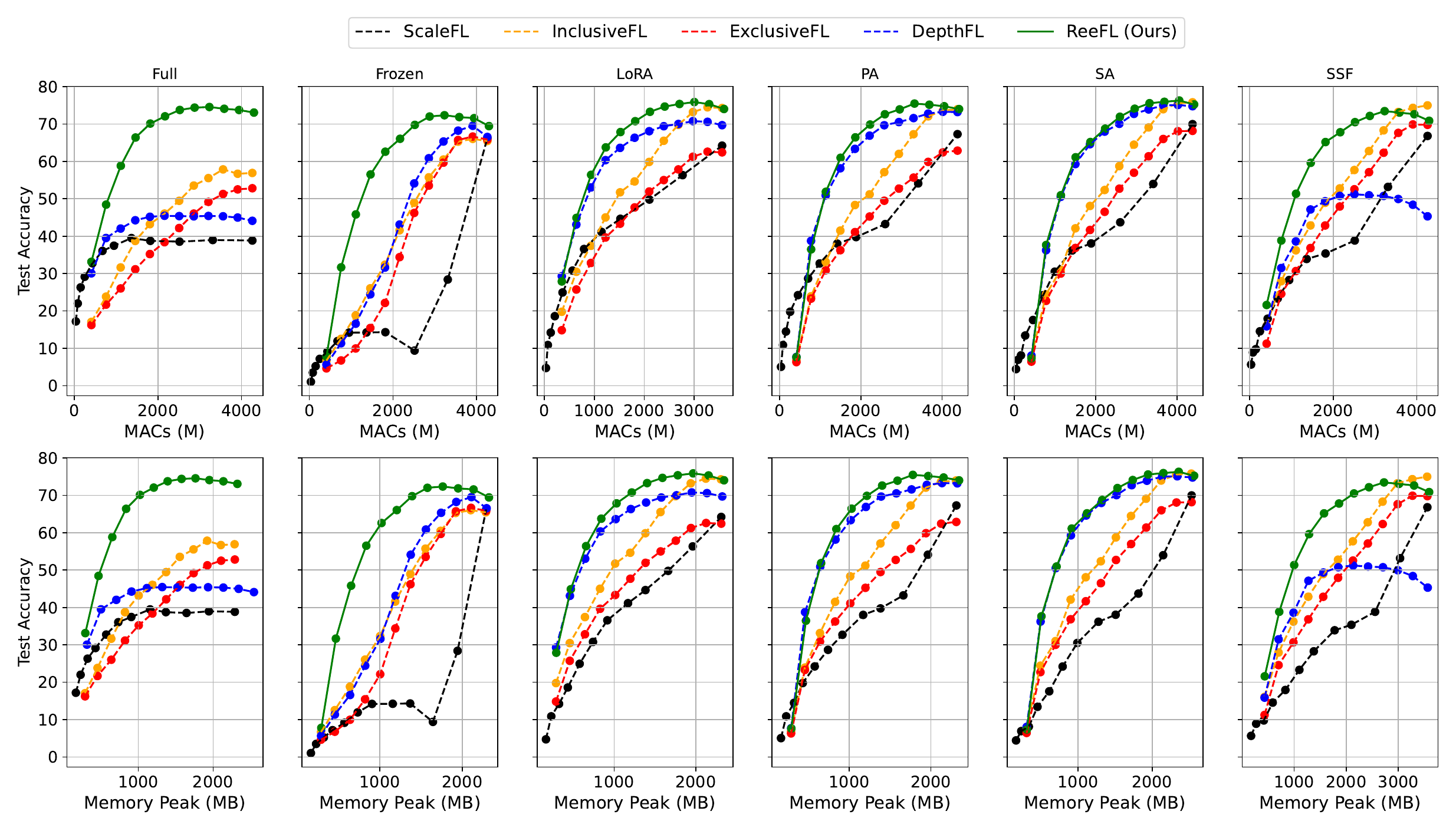}
    \captionsetup{font=small,labelfont=bf}
    \vspace{-1.0em}
    \caption{Quantifying training costs for each exit for CIFAR-100, $\alpha=1.0$ for 12 exits, where each dot along each line represents an exit. Similar results on other datasets and scenarios can be found in the Appendix.}
    \label{fig:profiling_cifar1.0}
    \vspace{-0.5em}
\end{figure*}

We also observe that ScaleFL, although being highly efficient, often struggles to beat the naive baseline ExclusiveFL, especially for Frozen and PEFT scenarios.
With full fine-tuning and PEFT, ScaleFL often performs better than ExclusiveFL and InclusiveFL at lower compute and memory bounds as shown in Figure.~\ref{fig:profiling_cifar1.0} and the Appendix Section.~\ref{app:sec:extended_results}.
However, their performance is inferior in deeper layers, resulting in the low average performance shown in Tables~\ref{tab:mainres_e4} \& \ref{tab:mainres_e12}. As observed in many previous works such as DepthFL and InclusiveFL, width-based scaling requires adequate retraining of these pruned channels. Since deeper layers are fine-tuned with fewer data, ScaleFL fails to beat the other baselines in most higher compute and memory regimes. This observation can also be seen with a frozen backbone, where ScaleFL fails to outperform all baselines in the accuracy and memory trade-off.

Lastly, for both $4$ and $12$ exits scenarios, full finetuning of the backbone model along with DepthFL or InclusiveFL
led to worse performance than using a PEFT method on CIFAR-100 and FEMNIST datasets. Although PEFT outperforming full finetuning is a common phenomenon~\cite{hu2021lora,zhang2023fedpetuning,zhao2023breaking,basu2023strong}, its cause has not been thoroughly investigated. We hypothesize that since the model is pre-trained with ImageNet, the domain gap is smaller as compared to domain gap to SpeechCommands, hence, the model can easily overfit on the clients' small local dataset and lose its generalizability. 

\textbf{\method{}'s Consistent Performance.}~
The baselines in Tables~\ref{tab:mainres_e4} \& \ref{tab:mainres_e12} show different performance depending on the finetuning method. For instance, DepthFL, in most cases, is the second best-performing baseline. Nonetheless, it performs poorly on the SSF PEFT method and fails to converge on more challenging scenarios, \textit{e.g.} $12$ exits with high data heterogeneity on some PEFT methods including SA, PA, and SSF. \method{}, on the other hand, is more consistent across different fine-tuning methods. This is due to four main reasons: 1) \module{} is shared and trained on the full dataset, 2) \module{} utilizes features from multi-layers, 3) \method{}'s knowledge distillation is dynamic, and 4) \method{} uses FedAvg which is more robust than FedDyn in highly heterogeneous data and resource scenarios. In existing baselines, classifiers rely on features from a single layer, deep classifiers are trained on partial data, and exits are manually selected as teacher exits for knowledge distillation. The performance of these baselines is more dependent on the given scenario and is hence less consistent. 

For instance, due to the low data regime at deeper exits, these exits are more sensitive to the fine-tuning method used, e.g. full full-tuning leads to overfitting. \method{} counteracts this drawback by 1) utilizing features from earlier exits in addition to the features of the current exit and 2) learning the fusion, through \module{}, and classification of these features on the full dataset. As another example, as different fine-tuning methods result in different performance for each exit, manually picking the teacher exits is only advantageous in scenarios where the teacher exits are high-performing. \module{}, on the other hand, dynamically selects this teacher exit, resulting in more consistent gains.

Lastly, in the case of DepthFL, FedDyn is the recommended choice for parameter aggregation. FedDyn dynamically modifies the local loss functions such that local models converge to a consensus that is consistent with a stationary point of the global loss. While FedDyn often leads to performance gains over FedAvg as seen in Table.~\ref{tab:ablation1} as well as other existing works that adopt FedDyn, we observe that in highly heterogeneous scenarios, e.g. CIFAR-100 $\alpha=0.1$, this consensus either leads to a sub-optimal stationary point or is not found, leading to divergence, as seen in Table.~\ref{tab:mainres_e12}. A more detailed discussion on the performance gap among these methods can be found in our ablation study (Section.~\ref{sec:ablation}) and in Appendix Section.~\ref{app:sec:detailed_discussion}.

\textbf{Training Costs.}~
A comparison between different FL algorithms remains incomplete unless metrics accounting for communication, compute, and memory costs undergone by the clients during training are collected. To this end, we measure the average communication cost per round shown in Tables.~\ref{tab:mainres_e4} \& \ref{tab:mainres_e12} for $4$ \& $12$ exits respectively. Additionally, we measure MACs (multiply-accumulate), using a single $224\!\times\!224$ input, as proxy metric of the amount of compute needed to train a model; and, we measure the memory peak at training time, using a batch size of $32$, to quantify the minimum amount of memory needed by a client to train the largest sub-model within its resource budget. 

As seen in Fig.~\ref{fig:profiling_cifar1.0}, \method{} outperforms existing baselines in most cases, achieving the best accuracy given similar MACs and peak memory constraints. Regarding communication costs, \method{} incurs, on average, an additional $\sim$1MB more than the second best performing baseline, DepthFL. Although \method has marginally fewer number of parameters, $\sim$2K parameters fewer, in the max resource budget scenario compared to the other depth-based scaling baselines, including InclusiveFL, DepthFL, and ExclusiveFL, its communication cost is constant and does not scale with the number of exits. As a result, every client has to send the shared \module{} \& classifier every round, leading to the slight increase in communication costs. However, we argue that this additional cost is marginal in most real-world scenarios, taking a small fraction of the average bandwidth supported by both broadband and mobile networks globally~\cite{ciscowhitepaper}. Moreover, in realistic FL deployment scenarios, the eligible clients are connected to power and to an unmetered connection (i.e. WiFi), hence the main bottleneck is no longer the upstream communication, but the device resource capacity~\cite{sysdesign_fl2019mlsys}. Hence, we place a greater emphasis on the benefit-cost ratio of on-device training.

\subsection{Ablation}\label{sec:ablation}

\textbf{Aggregation \& Knowledge Distillation.}~We weight the contribution of the aggregation strategy and knowledge distillation proposed in existing depth-based scaling works, namely DepthFL and InclusiveFL, as well as in our approach, \method{}. We show results for Frozen and the best-performing PEFT approach, LoRA, in Table.~\ref{tab:ablation1} and Appendix Table.~\ref{app:tab:ablation1} for $4$ and $12$ exits respectively. 

InclusiveFL proposed distilling from deeper backbone transformer blocks to shallower backbone blocks and hence not applicable for a frozen backbone; this distillation leads to a boost in performance in most cases for $4$ exits but often lead to a drop in performance in the more challenging setup of $12$ exits. Their proposed aggregation, FedAdam~\cite{fedadam}, outperforms InclusiveFL+FedAvg by a considerable margin in all cases. DepthFL's proposed aggregation, FedDyn~\cite{feddyn}, on the other hand, boosts DepthFL's performance in most cases for $4$ exits and some cases for $12$ exits. Their proposed distillation, deep mutual distillation, however, often leads to a slight drop in performance as weaker sub-models are also selected as teacher models. In contrast, \method{} distillation approach often picks the better performing sub-models as the teacher (Appendix Section~\ref{app:sec:teacher_selection}), resulting in a slight performance boost in most scenarios - also illustrated per-exit in Fig.~\ref{fig:ablation1_ee_e4}. 
Most importantly, our approach still outperforms both DepthFL and InclusiveFL by a significant margin when all three approaches forgo knowledge distillation and use the same aggregation strategy, FedAvg.


\textbf{\method{} Feature Modulation.}~In Section~\ref{sec:reefl}, we detail that \method{} aids feature learning through modulating the feature representations of the backbone model using the modulated class tokens (Eq.~\ref{eq:mod_cls_token}). In this ablation, we further elucidate its benefits by comparing to its unmodulated counterpart where each block in the backbone model uses the class token from the previous block. In Fig.~\ref{fig:overview}(b), we illustrate the effects of feature modulation through visualizing the attention maps between the class token and other tokens before and after applying \module{}. We observe that this helps the backbone model to build visually sensible feature representation especially in early layers. In Table.~\ref{tab:ablation2} and Appendix Table.~\ref{app:tab:ablation2}, we show that using the modulated class tokens as inputs to the backbone model helps to boost performance. This observation is especially evident in the scenario where the backbone model is frozen on the SpeechCommands dataset where the pretrained model fails to reuse pretrained image features for speech-based prediction without \method{}'s feature modulation. The contribution of \method{}'s feature modulation is smaller when using PEFT methods as these methods explicitly learns new features to adapt to the new domain. Nonetheless, our feature modulation technique aids feature learning in most scenarios without additional costs.

\textbf{Injecting \module{} Only at Exit Layers.} Instead of recurrently sharing \module{} among all layers, we inject \module{} only at the exits and present results for $4$ exits, including the mean performance difference with sharing \module{} in all layers, in Table.~\ref{tab:ree_exit_only}. As shown in the table above, sharing \module{} only at exit layers leads to either similar performance or a drop in performance. Notably, if we freeze the backbone model, there is a considerable drop in accuracy, showing that \module{} benefits from utilizing feature representations from all layers.

\vspace{-0.2em}
\section{Conclusion}

In this paper, we propose \method{}, a radically distinct approach that leverages recurrent early exits to better handle client heterogeneity, offering superior performance, training efficiency and scalability in federated fine-tuning. By learning to weight and fuse feature representations from sub-models of varying depth, we can utilize a single shared classifier for all clients. Additionally, we show that these fused feature respresentations can modulate the backbone model to improve feature learning and subsequent predictions. Coupled with our best-teacher distillation, we are able to boost the accuracy of underperforming sub-models.
As a future work, our approach can be extended to different modalities (e.g. language) and/or to include differential privacy noise for efficient private learning.



\newpage
\section*{Acknowledgements}
This work was supported by Samsung AI and the European
Research Council via the REDIAL project (Grant Agreement ID: 805194).

\section*{Impact Statement}

This paper presents work whose goal is to advance the field of federated learning (FL) when applied to heterogeneous devices. The goal of FL is to enable the training of machine learning models while protecting the privacy of end-users. There are many potential threats and attacks that threaten this privacy, none which we feel must be specifically highlighted here as our work does not focus on mitigating these attacks nor introduce additional vulnerabilities and is orthogonal to many of the current security measures in place.
\bibliography{main}
\bibliographystyle{icml2024}

\newpage
\appendix
\clearpage
\newpage

\section{Training \& Implementation Details}\label{app:sec:training_deets}

In Section~\ref{sec:setup}, we provide a summary of our experimental setup. In this section, we detail all hyperparameters and configurations used in all our experiments. Code is available at https://github.com/royson/reefl.

\textbf{Baseline Hyperparameters}~Following the original works, we set $\beta=0.2$ for InclusiveFL's momentum distillation. For DepthFL, we consistently ramp up the weight of the KL loss, $\eta$, for 300 rounds till $\eta=1.0$. For ScaleFL, we use $\eta=0.05$ and set the softmax temperature, $\tau=3.0$ - $\tau=1.0$ for all other works. For 4 exits, we follow ScaleFL's depth-scaling of adding classifiers to the 4-th, 6-th, 9-th and last transformer block and width-scaling ratios of $[0.4, 0.55, 0.75, 1]$. For 12 exits, we add a classifier for every block and scale the width ratio with ${\scriptstyle[0.25,0.3,0.35,0.4,0.48,0.55,0.61,0.69,0.75,0.84,0.92,1.]}$.

\textbf{\method{} Hyperparameters}~We scale \method{} to be similar in parameters than baselines for a fair comparison. The number of multi-attention heads in \module{} is set to $8$. The number of bottleneck features for MSA is set to $16$ and the number of hidden features in the MLP is set to $1.35\times$ the input features. Following DepthFL, we consistently ramp up $\eta$ for 300 rounds. Our running estimate hyperparameter $\zeta$ is set to $0.2$.

\textbf{Local Training Hyperparameters}~Each client trains its local parameter using SGD with momentum set to $0$, batch size set to $32$, for a single epoch. The other hyperparameters are selected using a simple grid search to pick the highest performing learning rate (LR) $[1e^{-1},5e^{-2},1e^{-2},5e^{-3},1e^{-3}]$, weight decay $[0,1e^{-2},1e^{-3},1e^{-4}]$, minimum LR after LR decay using a cosine annealing LR schedule $[1e^{-2},1e^{-3},1e^{-4},1e^{-5}]$. For FedAvg and FedAdam, we pick $5e^{-2}$ as the initial LR with a minimum LR of $1e^{-3}$ after the aforementioned LR decay. For FedDyn, we pick $1e^{-1}$ as the initial LR with a minimum LR of $1e^{-2}$ and a weight decay value of $1e^{-3}$. We also clip gradients by value $[-1,1]$ for all experiments for better stability.

\textbf{Server \& Aggregation Hyperparameters}~We set the total number of rounds to 1K, sampling 10\% of the total number of clients for each round. Following recommended settings, we set FedDyn's $\alpha=0.1$, not to be confused with $\alpha$ in the Dirichlet Distribution used in the main paper. For FedAdam, we set $\beta_1=0.9$, $\beta_2=0.999$, and the learning rate to $1e^{-3}$.

\textbf{Backbone Model Hyperparameters}~We use a vision transformer model, pre-trained DeiT-S, as our backbone for all experiments. Our backbone, hence, has 12 blocks, a hidden dimension of $384$, $6$ multi-attention heads, and divides the input image into $16\times16$ patches. Instead of experimenting with different models, we experiment with different PEFT fine-tuning methods which inserts different learnable adapters to the backbone model. 

\textbf{PEFT Hyperparameters}~Rank of LoRA, Parallel Adapter (PA), Serial Adapter (SA) is set to $32$. LoRA's $\alpha$, not to be confused with $\alpha$ in the Dirichlet Distribution used in the main paper, is set to $64$. Following their original works, LoRA is applied to the query and value projection matrices, PA \& SA are inserted in each MSA and MLP block, and SSF is inserted to every learnable parameter except the last Linear layer in each classifier.

\textbf{Data Hyperparameters}~To utilize the pretrained backbone model, we resize all inputs to $224\times224$ using bilinear interpolation. During training, we augment CIFAR-100 and FEMNIST images by randomly cropping and randomly flipping the images horizontally. Following best practices, we normalize all images with the mean and SD of the respective dataset. 


\section{Baseline Details.}\label{app:sec:baseline_deets}

In this section, we present further details of each baseline that we compare with. ExclusiveFL is a standard baseline commonly used in existing works where clients with insufficient compute or memory are left out. 
InclusiveFL and DepthFL are depth-based approaches, which prune the deeper layers of the model to accommodate resource-constrained clients, showing considerable gains over previous width-based scaling methods. As the model is pruned by depth, these methods deploy a separate classifier at each exit. 

Two key differences between InclusiveFL and DepthFL include how local optimization is performed and how knowledge transfer is utilized. In InclusiveFL, each client trains its transformer-based sub-model and the sub-model's deepest classifier. On the server, InclusiveFL distills knowledge from deeper transformer layers to earlier transformer layers by injecting a gradient momentum distillation term that applies an extent of the average gradients of the deeper layers to the earlier layers. In contrast, each client in DepthFL trains all classifiers in its sub-model, along with the sub-model itself, and employ mutual self-distillation where each classifier acts as a teacher for the other classifiers. DepthFL also manually hand-picked different classifiers for different exits, with earlier exits having larger classifiers. As mentioned in Section.~\ref{sec:baseline_summary}, we keep the classifier architecture fixed for all exits in ReeFL and the considered baselines for fair comparisons; the number of parameters of the global model for ReeFL, DepthFL, and InclusiveFL is similar.

ScaleFL, on the other hand, does both width-based and depth-based scaling, often resulting in models that are more efficient than depth-based scaling approaches. Besides additionally pruning channels, ScaleFL employs a grid-search to decide where to prune the global model for each sub-model as opposed to pruning the global model uniformly in ReeFL, DepthFL, and InclusiveFL. 

\section{Teacher Sub-model Selection}\label{app:sec:teacher_selection}

\begin{figure*}[t]
\centering
\begin{subfigure}{0.47\textwidth}
    \centering
    \includegraphics[trim=0 0 0 0, clip, width=0.3\textwidth]{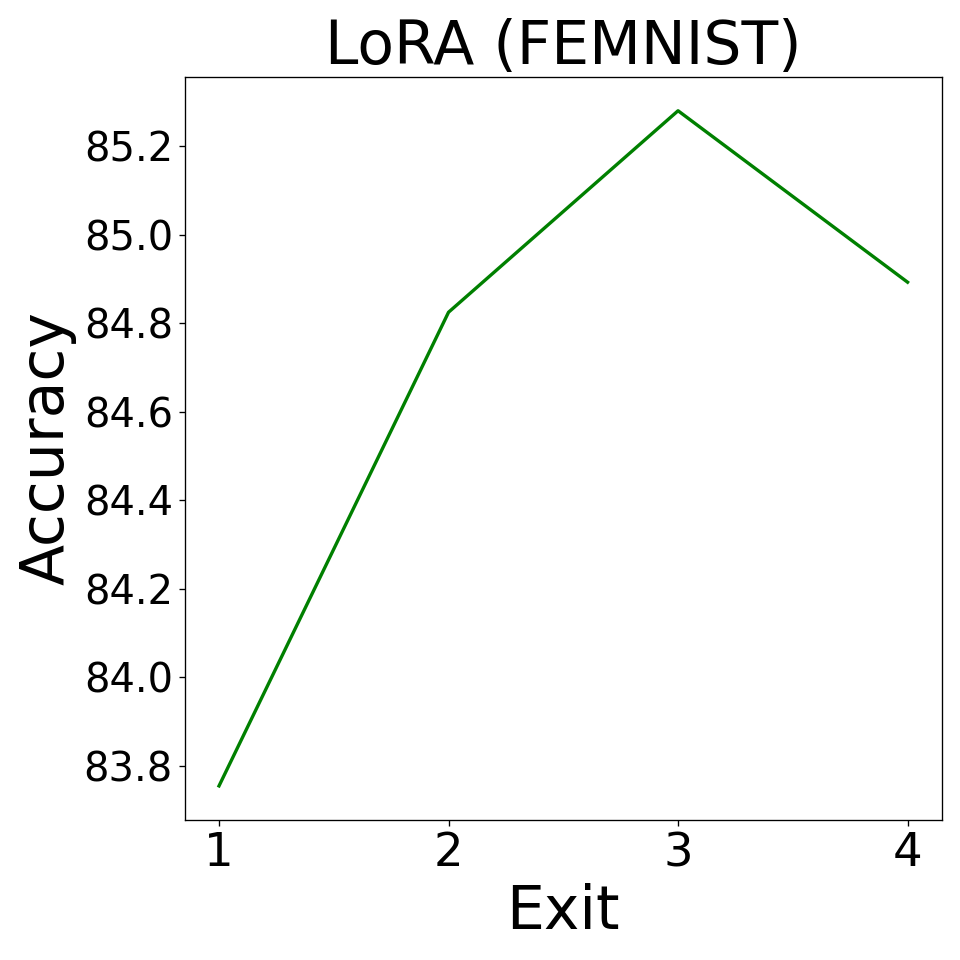}
    \hfill
    \includegraphics[trim=0 0 0 0, clip, width=0.3\textwidth]{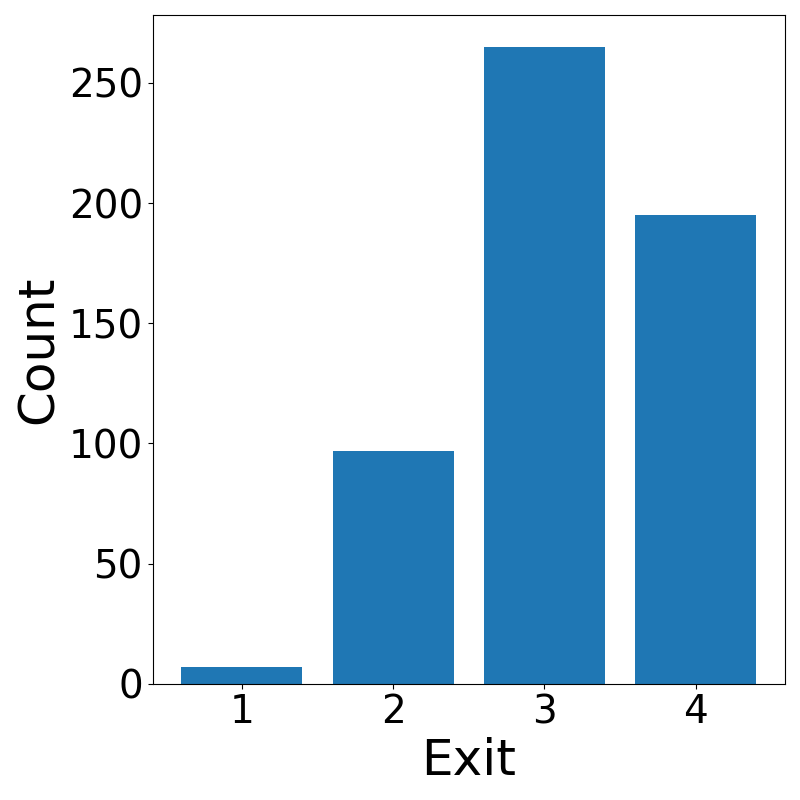}
    \hfill
    \includegraphics[trim=0 0 0 0, clip, width=0.3\textwidth]{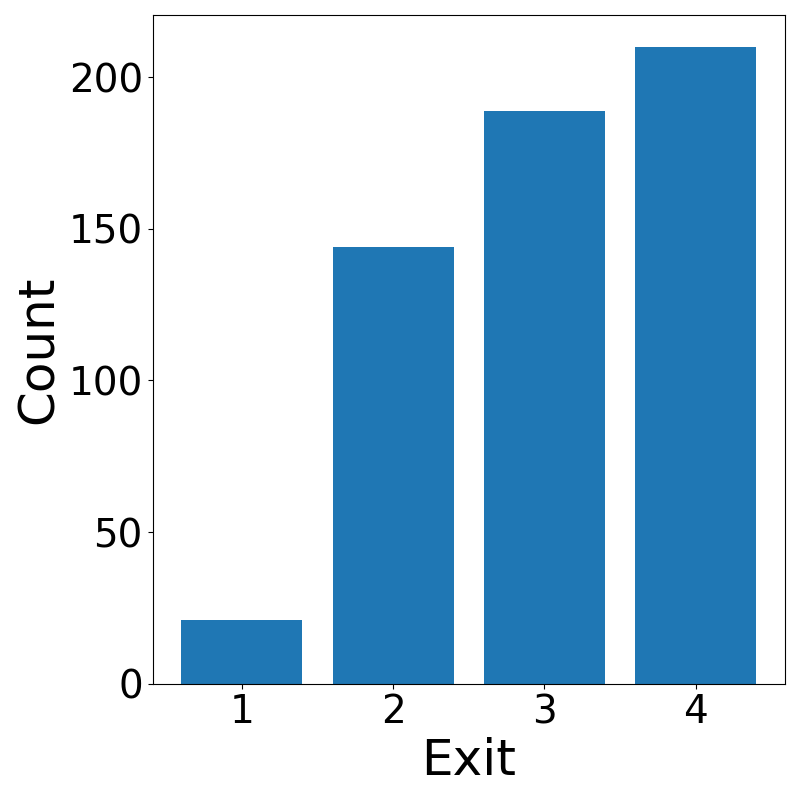}
    \caption{4 Exits}
\end{subfigure}
\hspace{1em}
\begin{subfigure}{0.47\textwidth}
    \centering
    \includegraphics[trim=0 0 0 0, clip, width=0.3\textwidth]{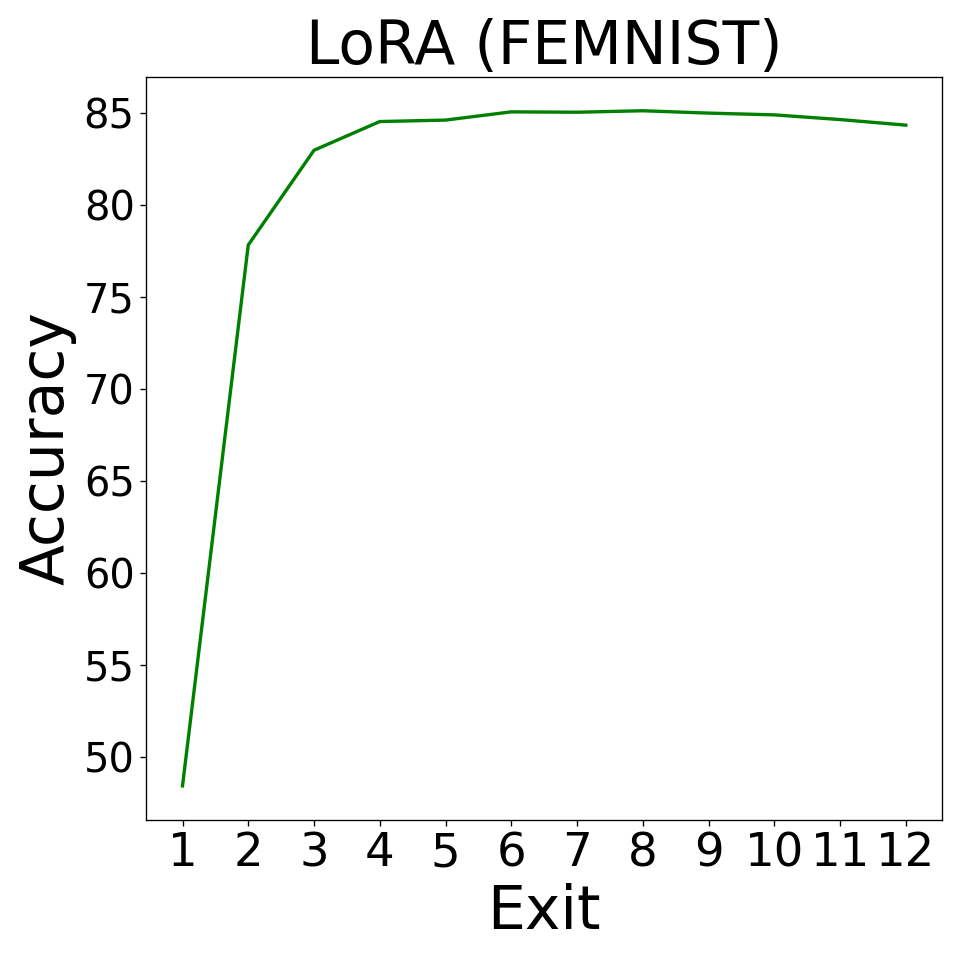}
    \hfill
    \includegraphics[trim=0 0 0 0, clip, width=0.3\textwidth]{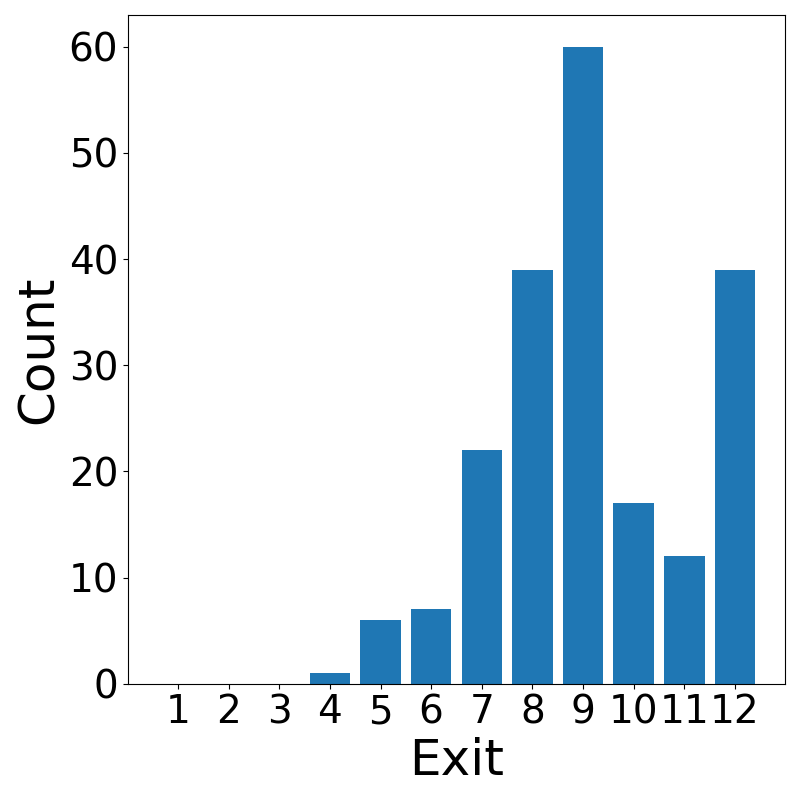}
    \hfill
    \includegraphics[trim=0 0 0 0, clip, width=0.3\textwidth]{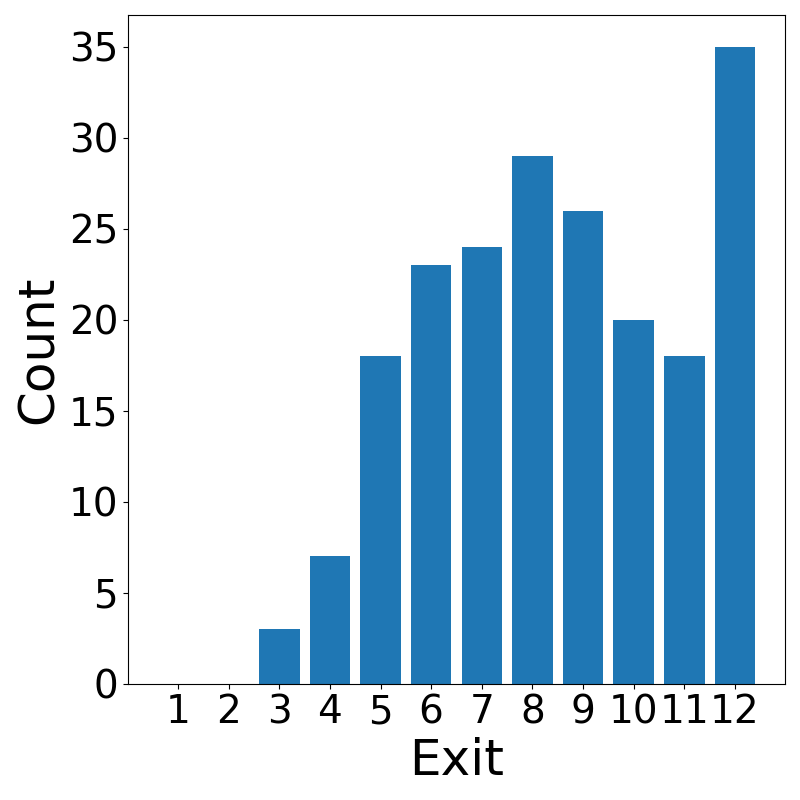}
    \caption{12 Exits}
\end{subfigure}

\
\vspace{-2.0em}
\caption{LoRA finetuning on FEMNIST with \method{} mean accuracy per exit (left), number of times each exit is selected as the teacher sub-model for knowledge distillation across \textbf{max resource budget clients} while taking the \textbf{running estimate} of the training loss for a single round (middle), and selecting the teacher sub-model by simply using the training loss of each mini-batch (right).}
\label{fig:moving_avg}
\end{figure*}

In \method{}, we select the best performing teacher sub-model to distill from per client based on the running estimate of the training loss of the respective client (Section.~\ref{sec:reefl}). In Figure.~\ref{fig:moving_avg}, we show the average test performance of \method{} for each exit and the number of times each exit is selected as the teacher sub-model across all \textbf{max budget} clients for a single round using the running estimate of the training loss versus using the training loss of each mini-batch. We observe that \textit{1)} the best performing exit gets selected the most often across clients and \textit{2)} utilizing the training loss without the running estimate often leads to picking sub-optimal teacher sub-models. 

\section{Extended Results}\label{app:sec:extended_results}

\begin{figure*}[hbtp]
    \centering
    \includegraphics[width=.9\textwidth]{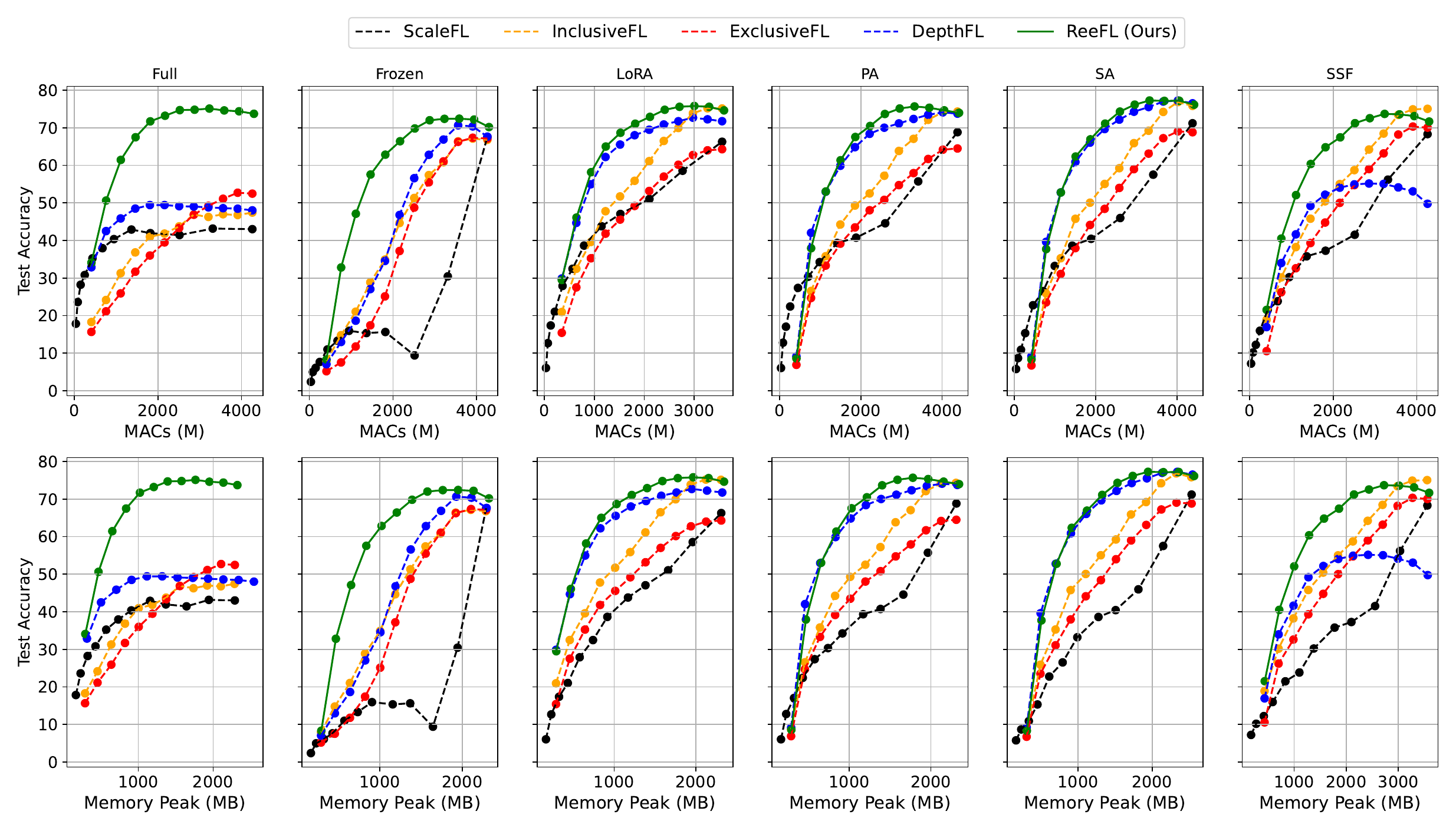}
    \captionsetup{font=small,labelfont=bf}
    \vspace{-1.0em}
    \caption{Quantifying training costs for each exit for CIFAR-100, $\alpha=1000$ for 12 exits, where each dot along each line represents an exit.}
    \label{fig:profiling_cifar1000}
\end{figure*}

\begin{figure*}[hbtp]
    \centering
    \includegraphics[width=.9\textwidth]{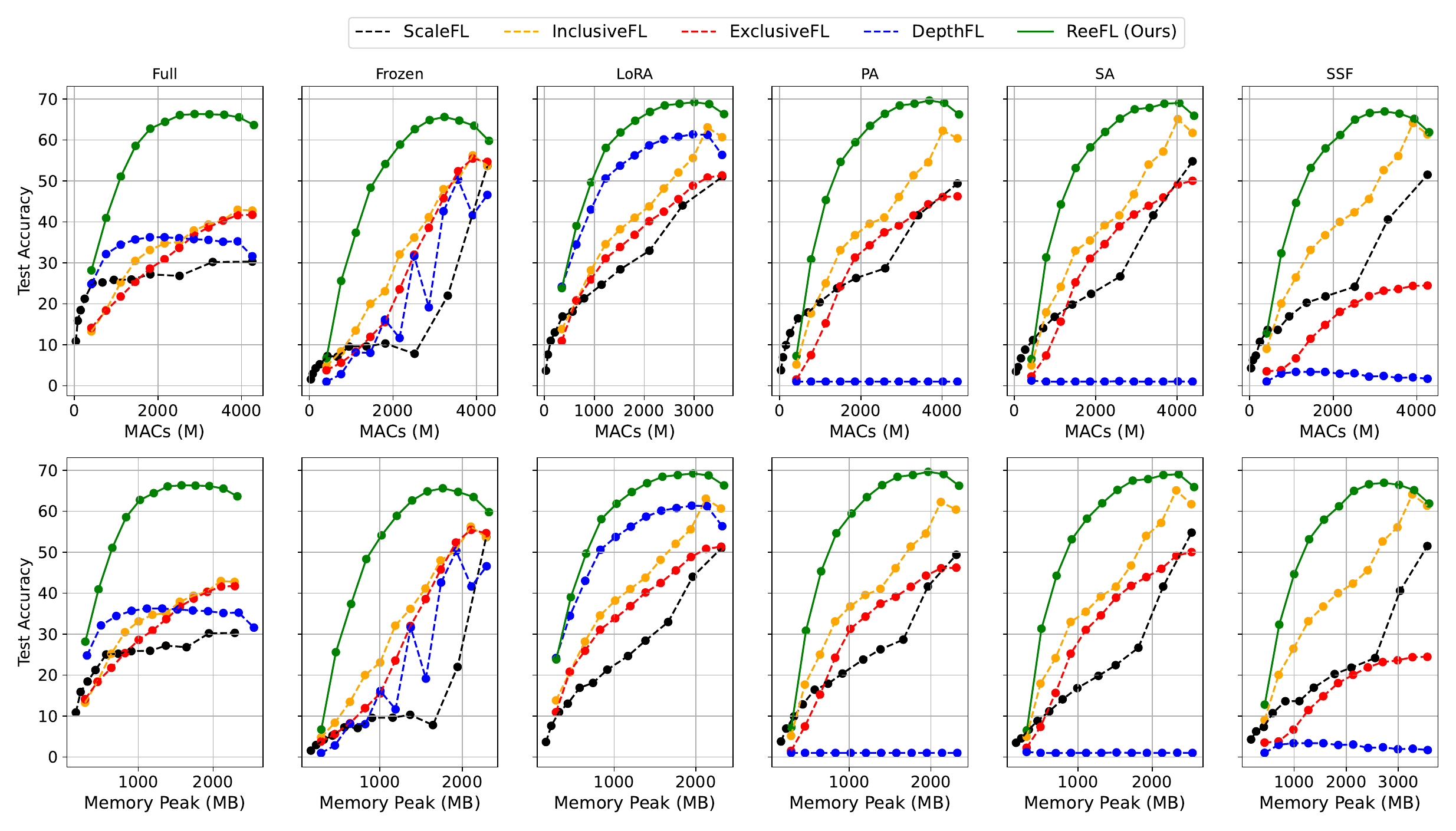}
    \captionsetup{font=small,labelfont=bf}
    \vspace{-1.0em}
    \caption{Quantifying training costs for each exit for CIFAR-100, $\alpha=0.1$ for 12 exits, where each dot along each line represents an exit.}
    \label{fig:profiling_cifar0.1}
\end{figure*}

\begin{figure*}[hbtp]
\centering
\begin{subfigure}{0.3\columnwidth}
    \includegraphics[trim=0 0 0 0, clip, width=0.97\columnwidth]{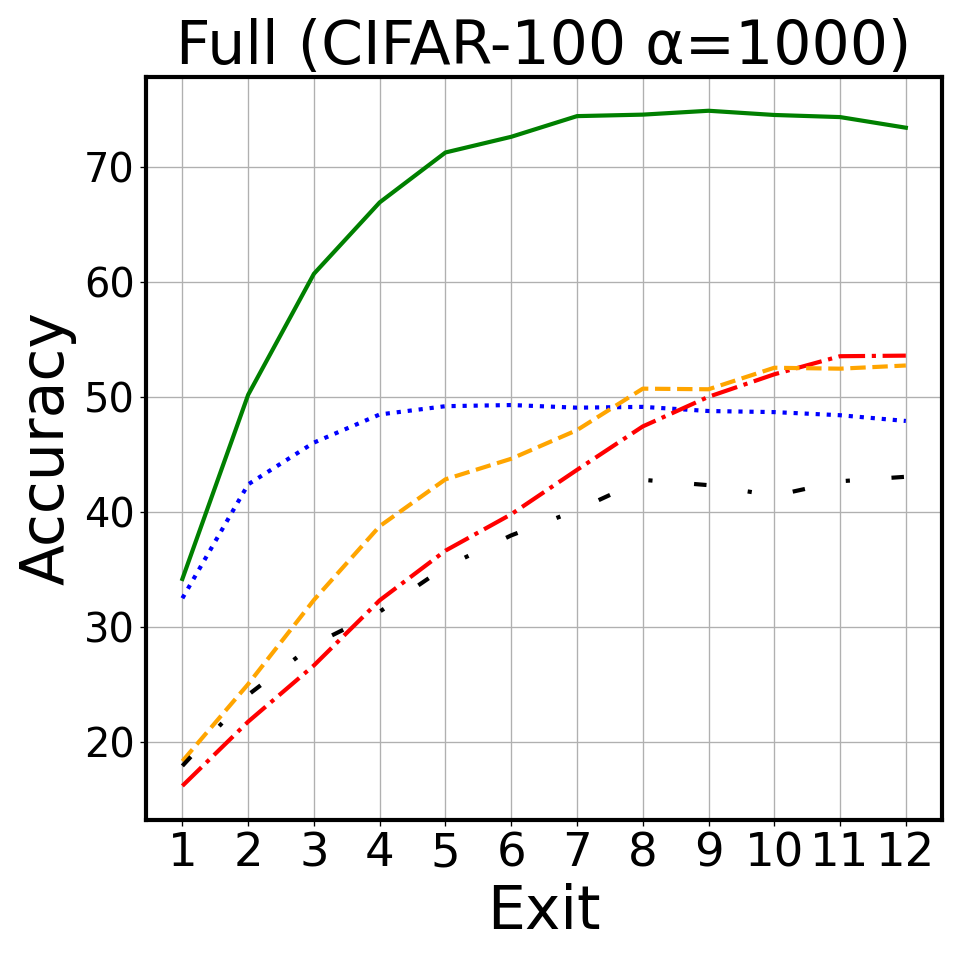}
\end{subfigure}
\begin{subfigure}{0.3\columnwidth}
    \includegraphics[trim=0 0 0 0, clip, width=0.97\columnwidth]{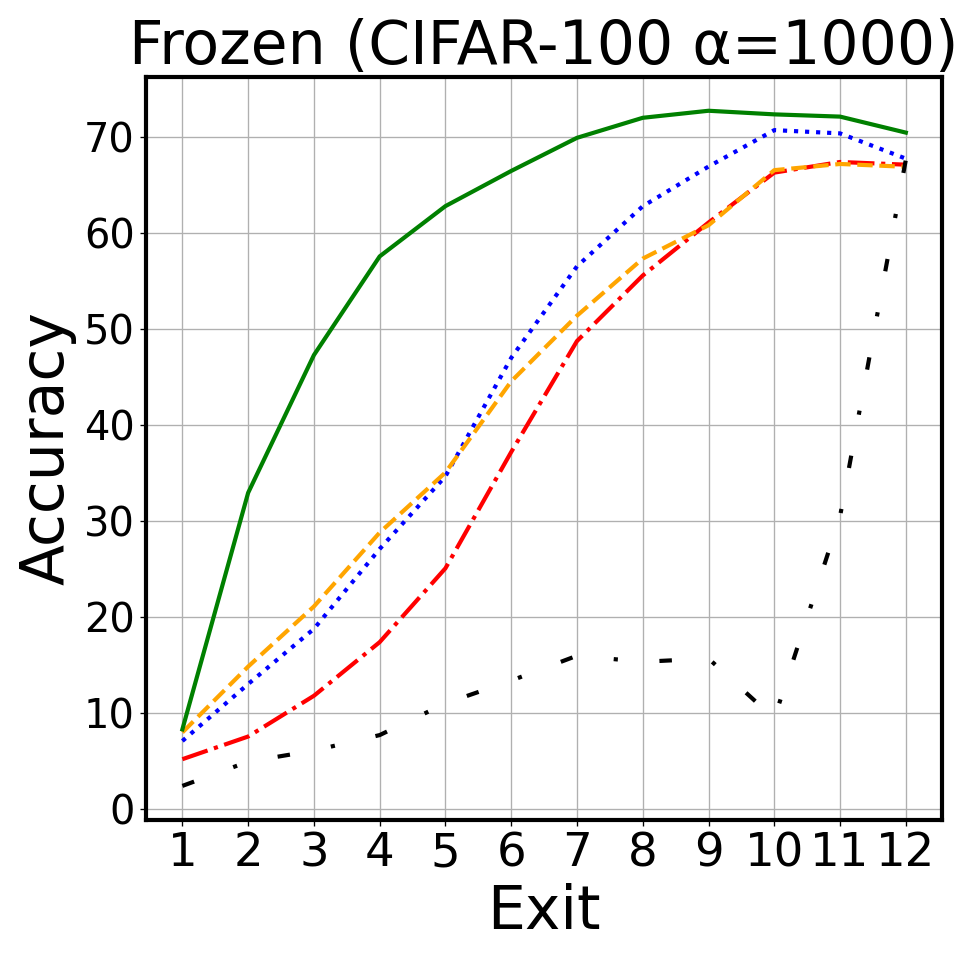}
\end{subfigure}
\begin{subfigure}{0.3\columnwidth}
    \includegraphics[trim=0 0 0 0, clip, width=0.97\columnwidth]{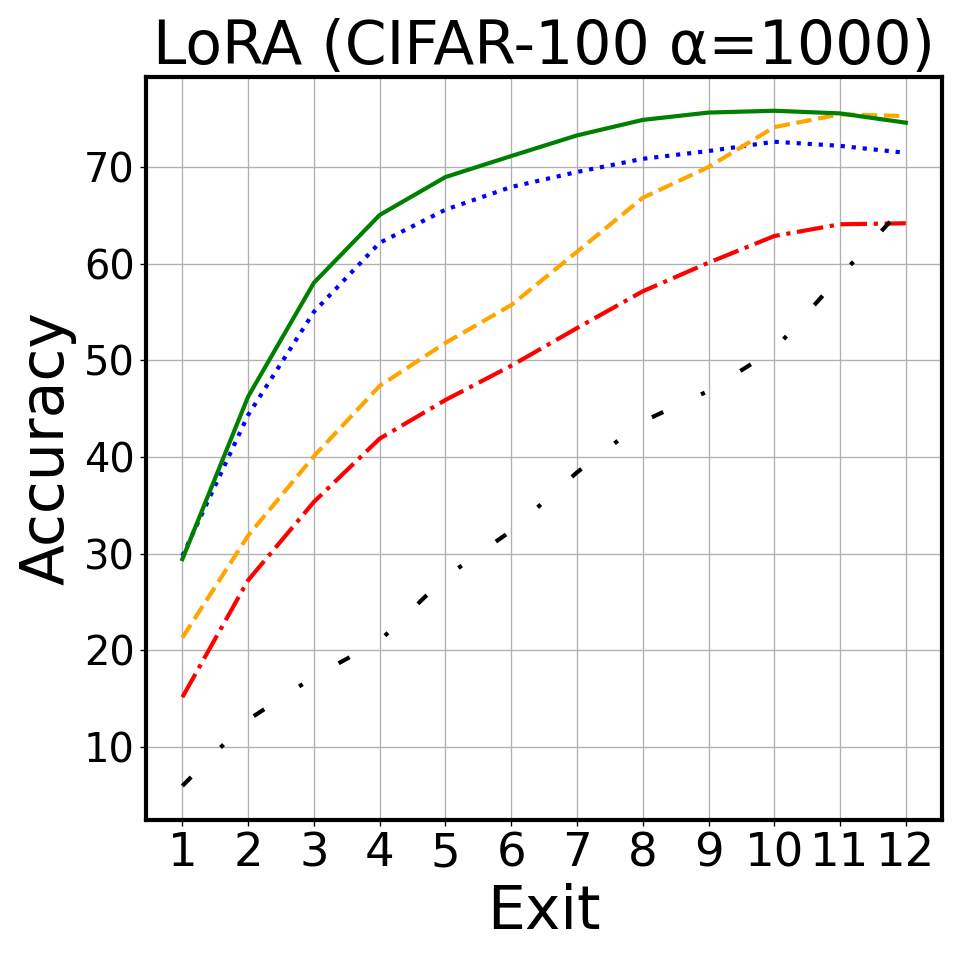}
\end{subfigure}
\begin{subfigure}{0.3\columnwidth}
    \includegraphics[trim=0 0 0 0, clip, width=0.97\columnwidth]{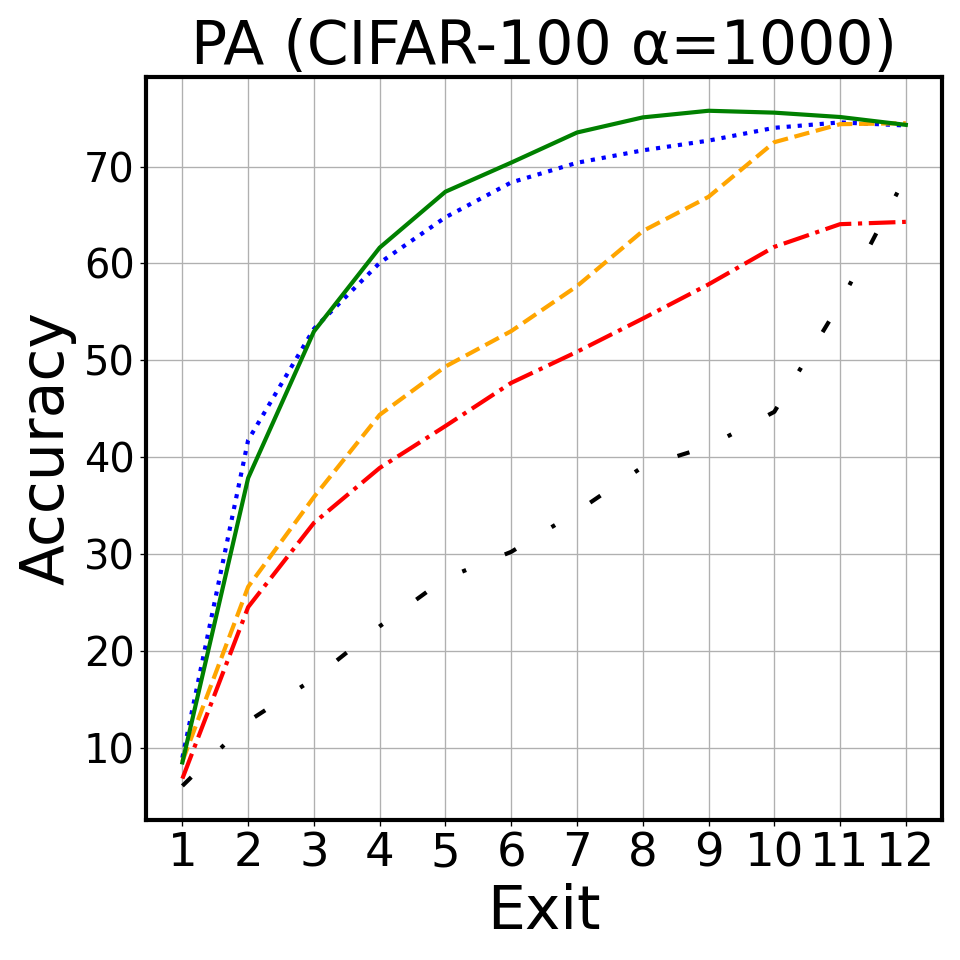}
\end{subfigure}
\begin{subfigure}{0.3\columnwidth}
    \includegraphics[trim=0 0 0 0, clip, width=0.97\columnwidth]{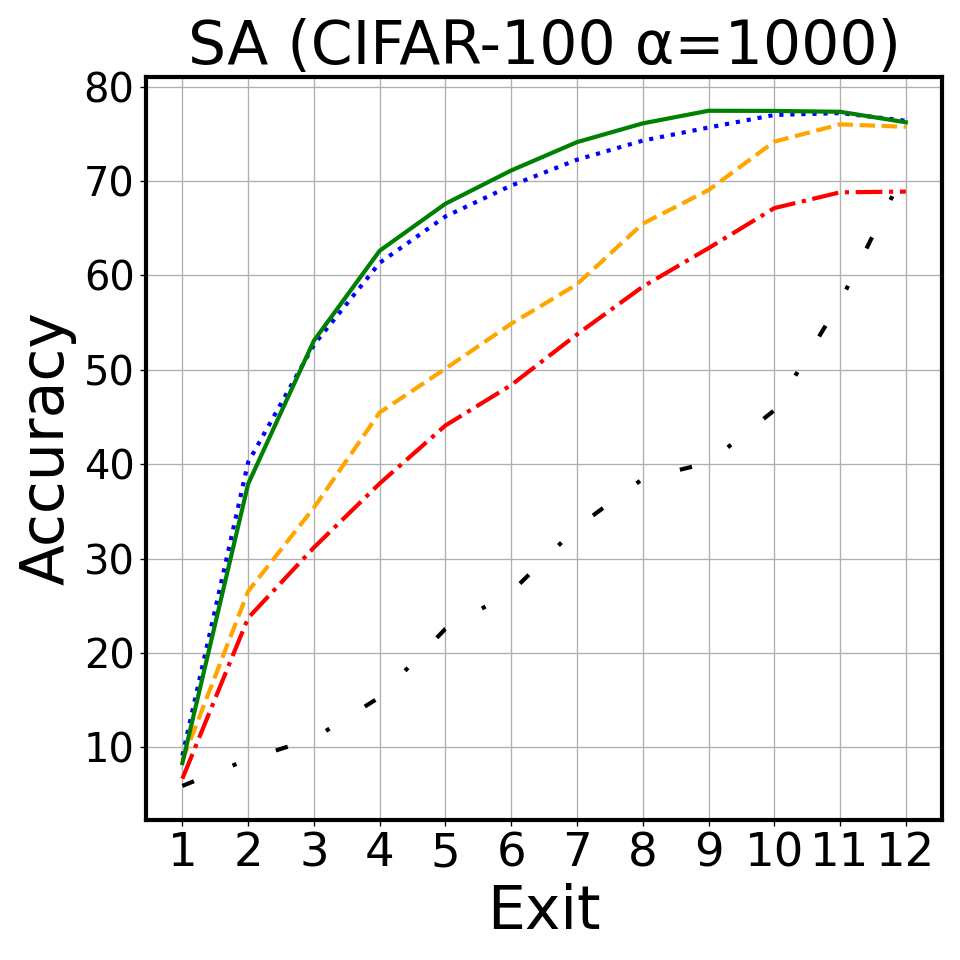}
\end{subfigure}
\begin{subfigure}{0.3\columnwidth}
    \includegraphics[trim=0 0 0 0, clip, width=0.97\columnwidth]{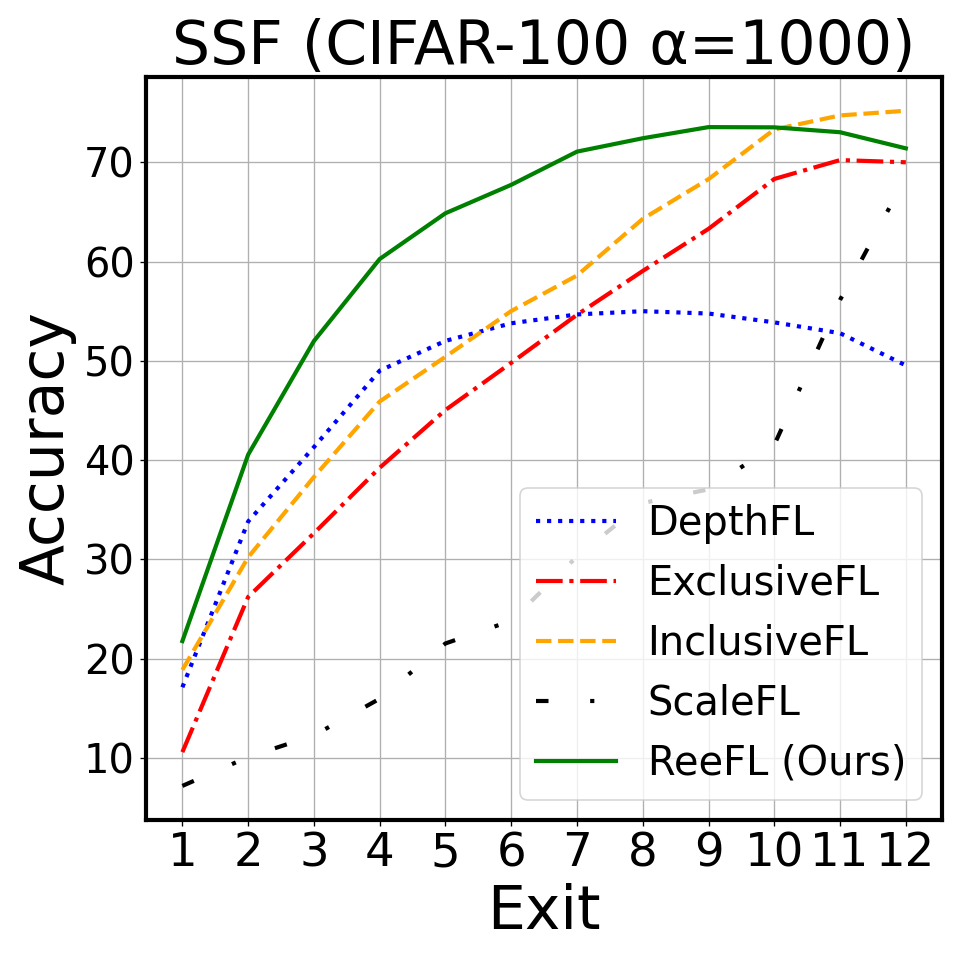}
\end{subfigure} \\

\begin{subfigure}{0.3\columnwidth}
    \includegraphics[trim=0 0 0 0, clip, width=0.97\columnwidth]{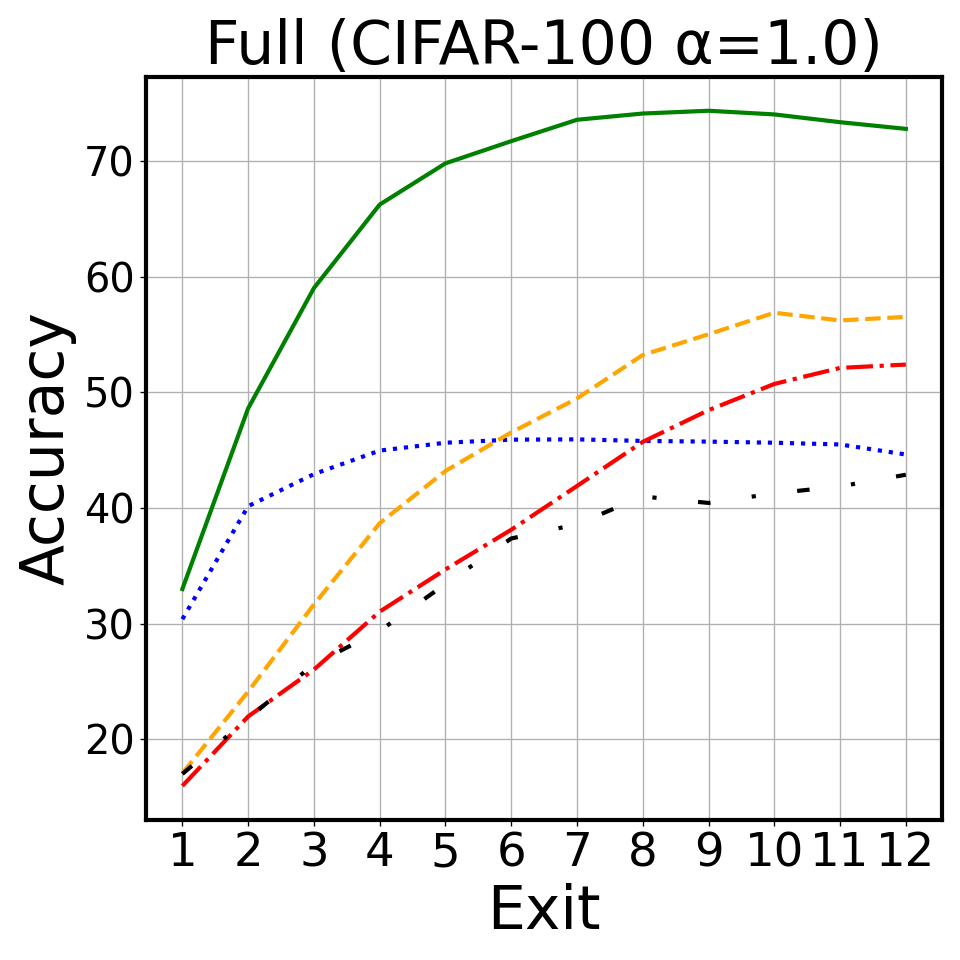}
\end{subfigure}
\begin{subfigure}{0.3\columnwidth}
    \includegraphics[trim=0 0 0 0, clip, width=0.97\columnwidth]{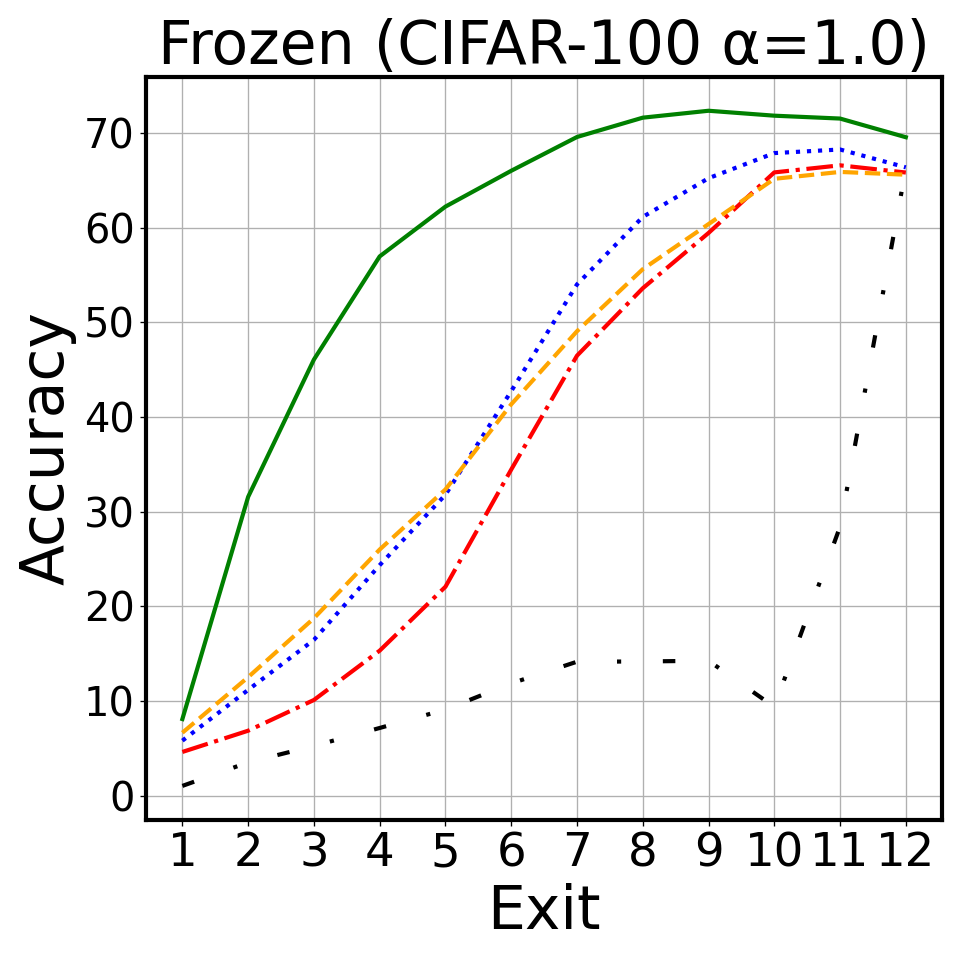}
\end{subfigure}
\begin{subfigure}{0.3\columnwidth}
    \includegraphics[trim=0 0 0 0, clip, width=0.97\columnwidth]{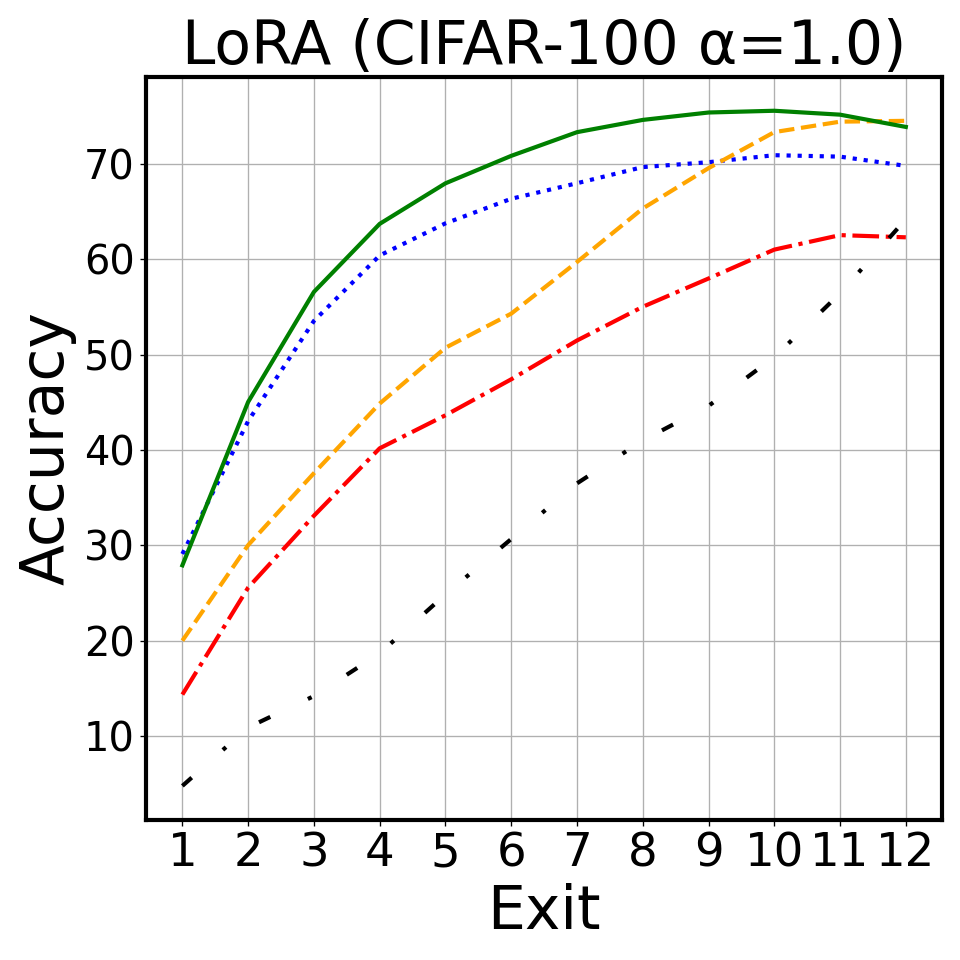}
\end{subfigure}
\begin{subfigure}{0.3\columnwidth}
    \includegraphics[trim=0 0 0 0, clip, width=0.97\columnwidth]{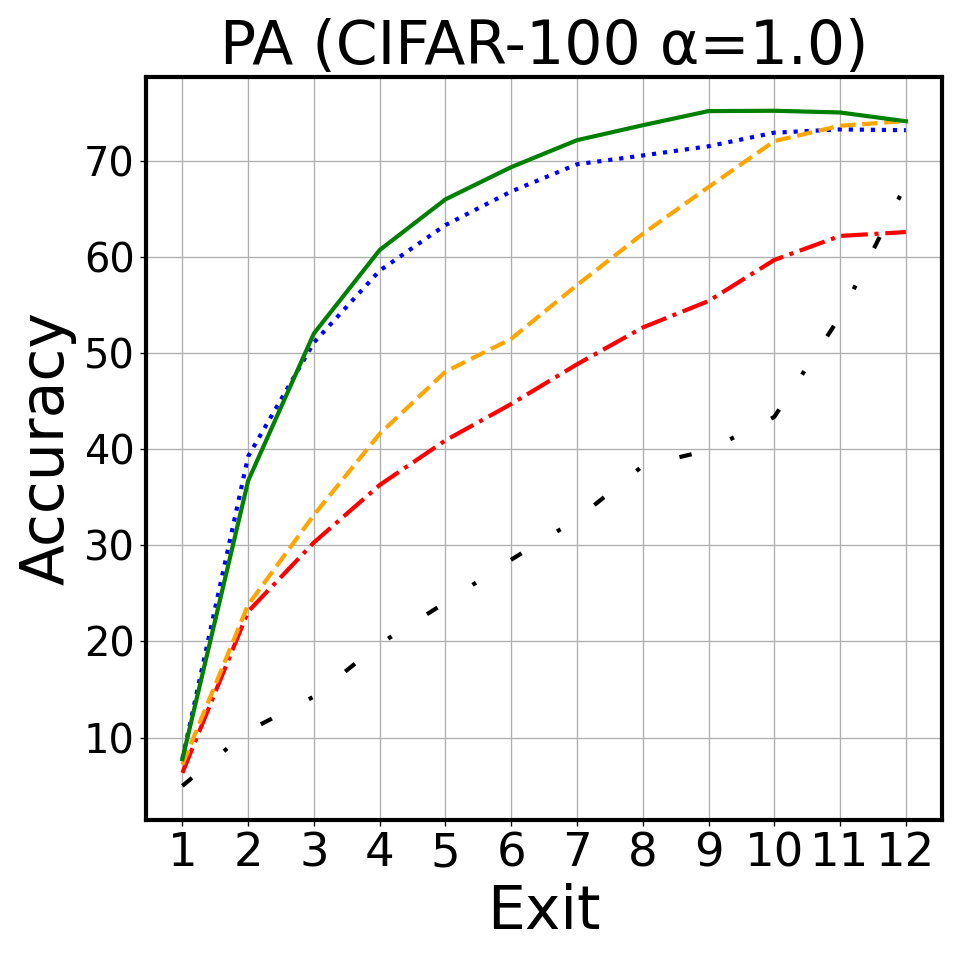}
\end{subfigure}
\begin{subfigure}{0.3\columnwidth}
    \includegraphics[trim=0 0 0 0, clip, width=0.97\columnwidth]{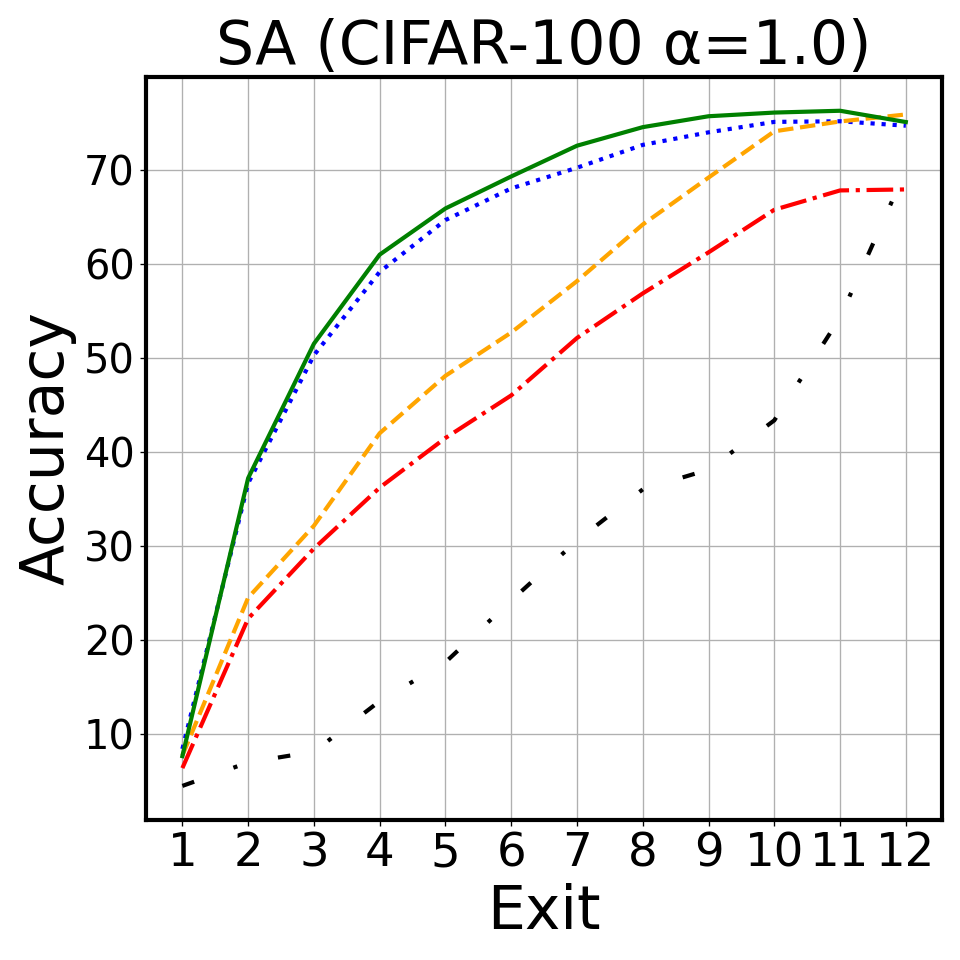}
\end{subfigure}
\begin{subfigure}{0.3\columnwidth}
    \includegraphics[trim=0 0 0 0, clip, width=0.97\columnwidth]{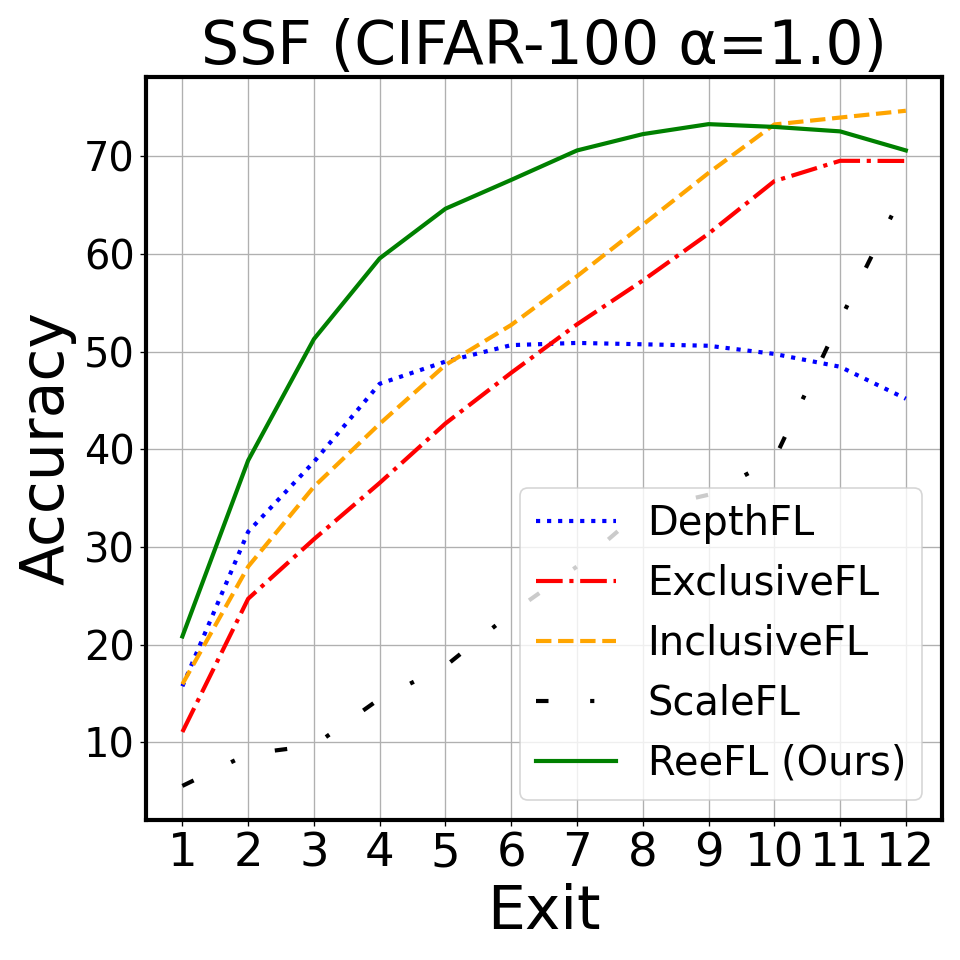}
\end{subfigure} \\

\begin{subfigure}{0.3\columnwidth}
    \includegraphics[trim=0 0 0 0, clip, width=0.97\columnwidth]{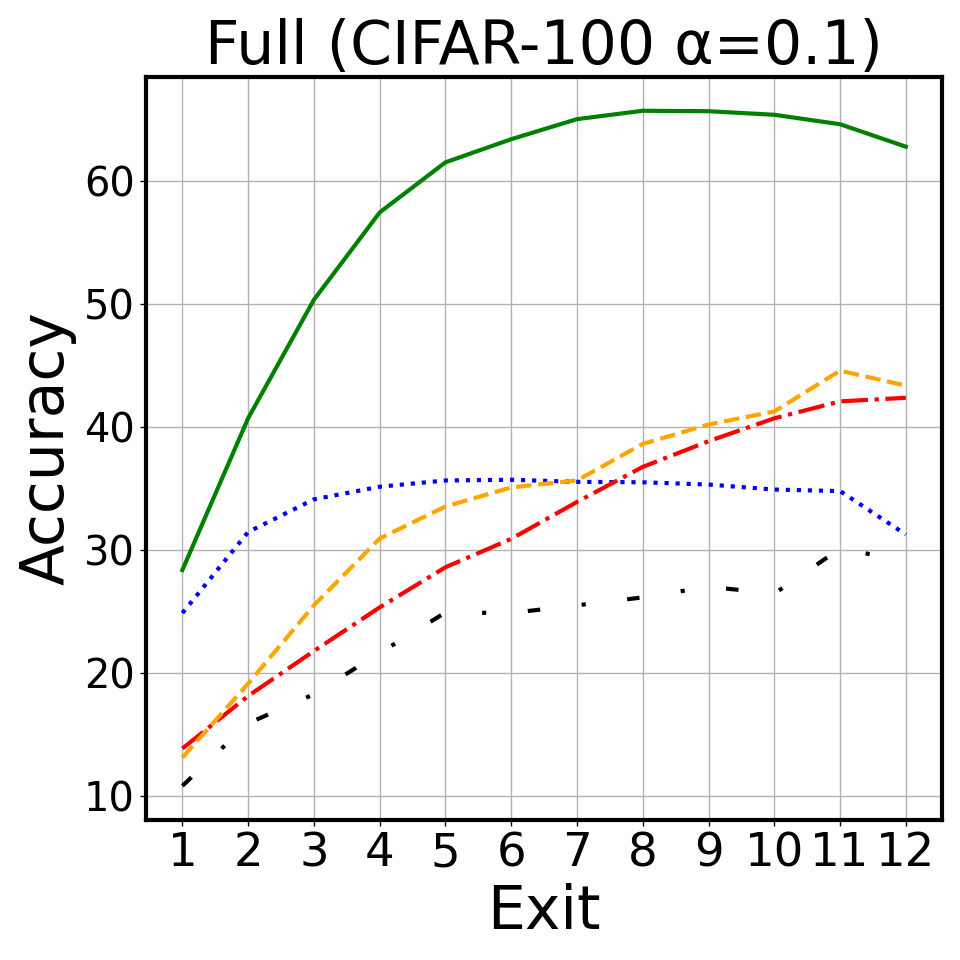}
\end{subfigure}
\begin{subfigure}{0.3\columnwidth}
    \includegraphics[trim=0 0 0 0, clip, width=0.97\columnwidth]{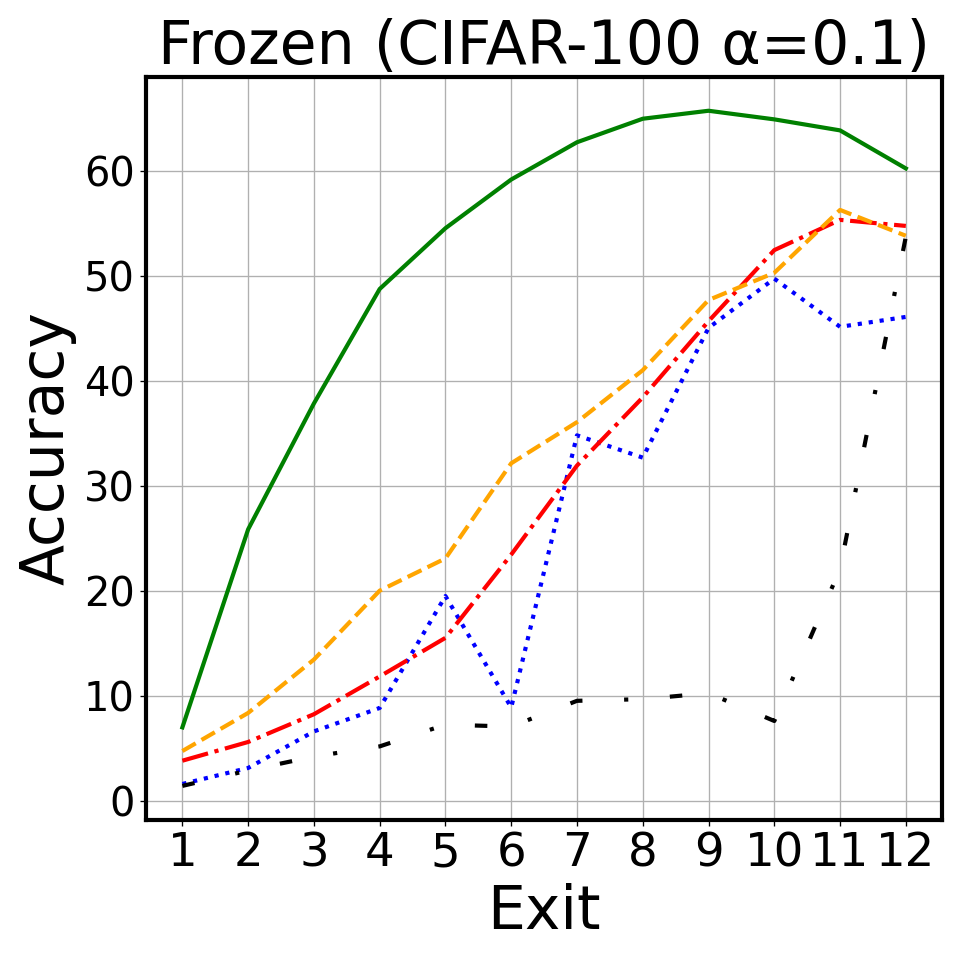}
\end{subfigure}
\begin{subfigure}{0.3\columnwidth}
    \includegraphics[trim=0 0 0 0, clip, width=0.97\columnwidth]{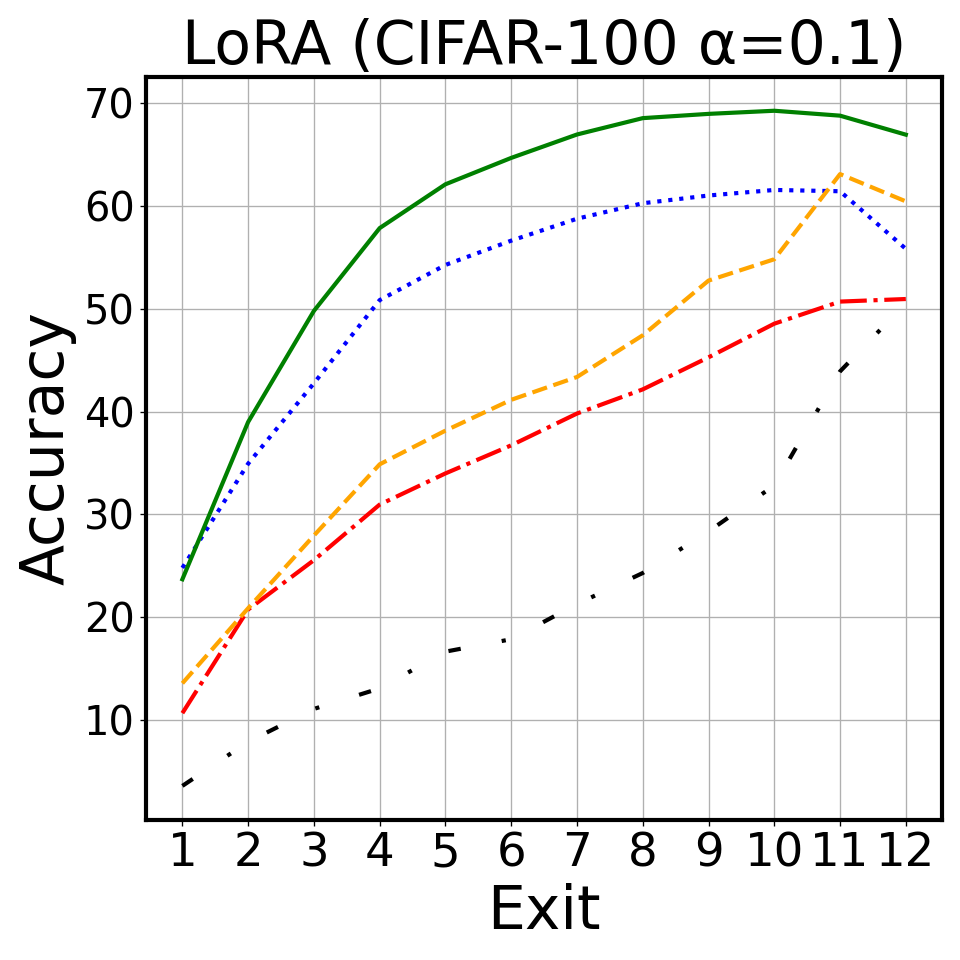}
\end{subfigure}
\begin{subfigure}{0.3\columnwidth}
    \includegraphics[trim=0 0 0 0, clip, width=0.97\columnwidth]{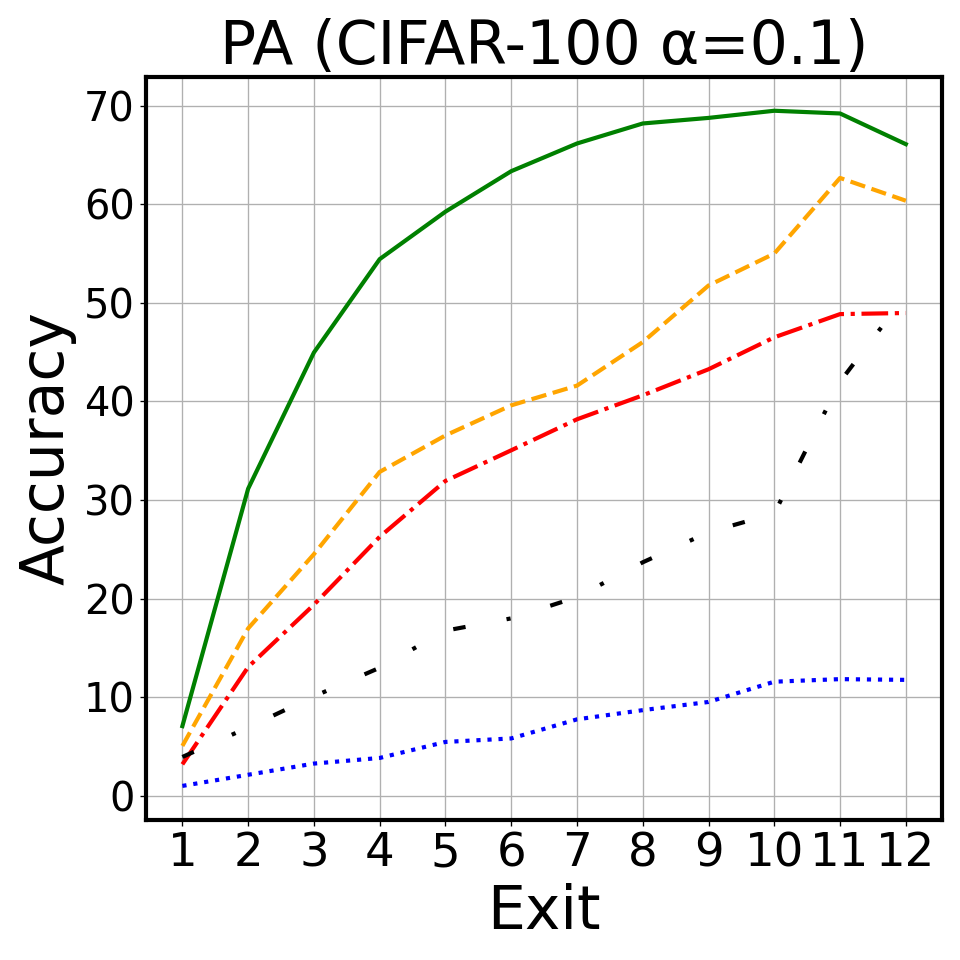}
\end{subfigure}
\begin{subfigure}{0.3\columnwidth}
    \includegraphics[trim=0 0 0 0, clip, width=0.97\columnwidth]{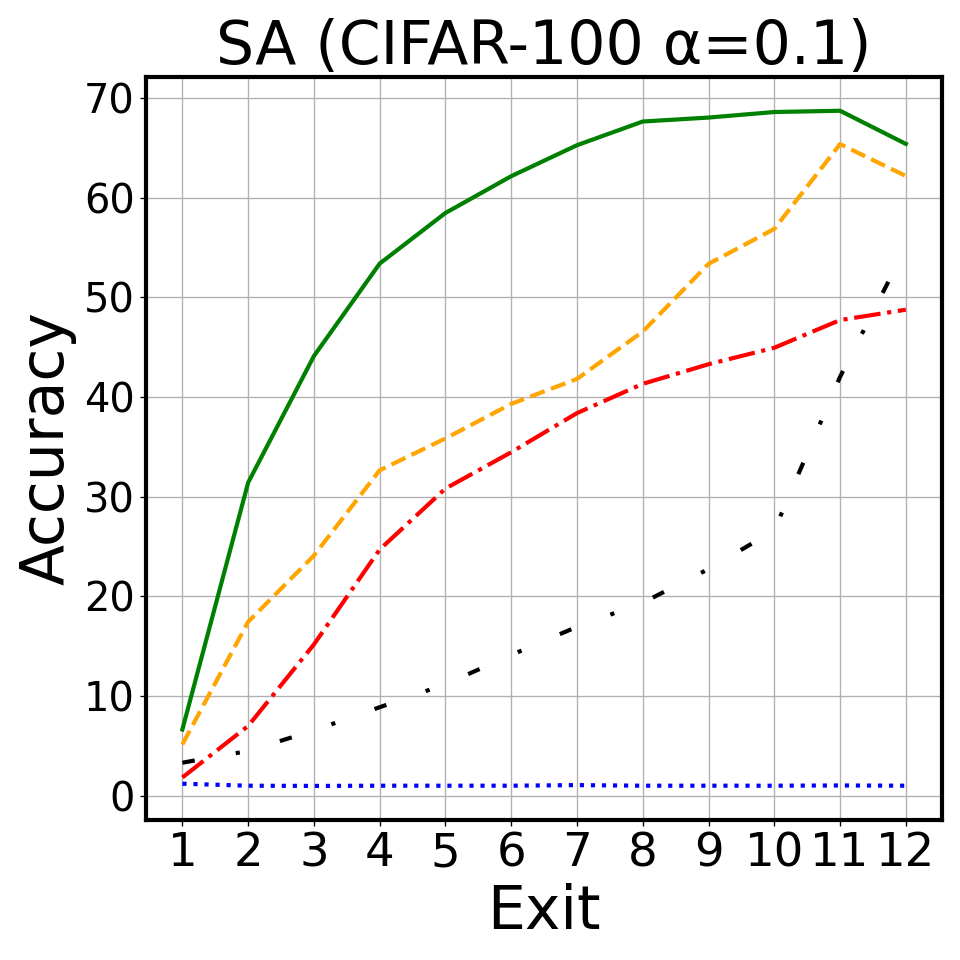}
\end{subfigure}
\begin{subfigure}{0.3\columnwidth}
    \includegraphics[trim=0 0 0 0, clip, width=0.97\columnwidth]{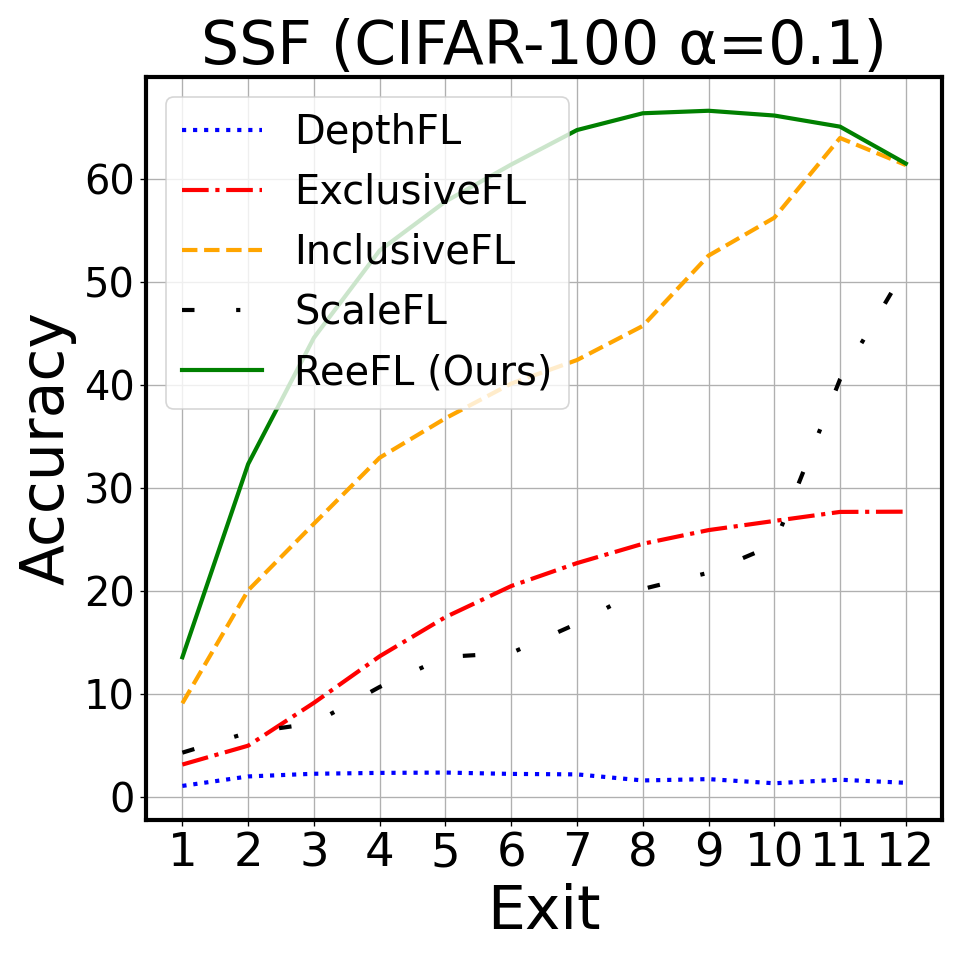}
\end{subfigure} \\

\begin{subfigure}{0.3\columnwidth}
    \includegraphics[trim=0 0 0 0, clip, width=0.97\columnwidth]{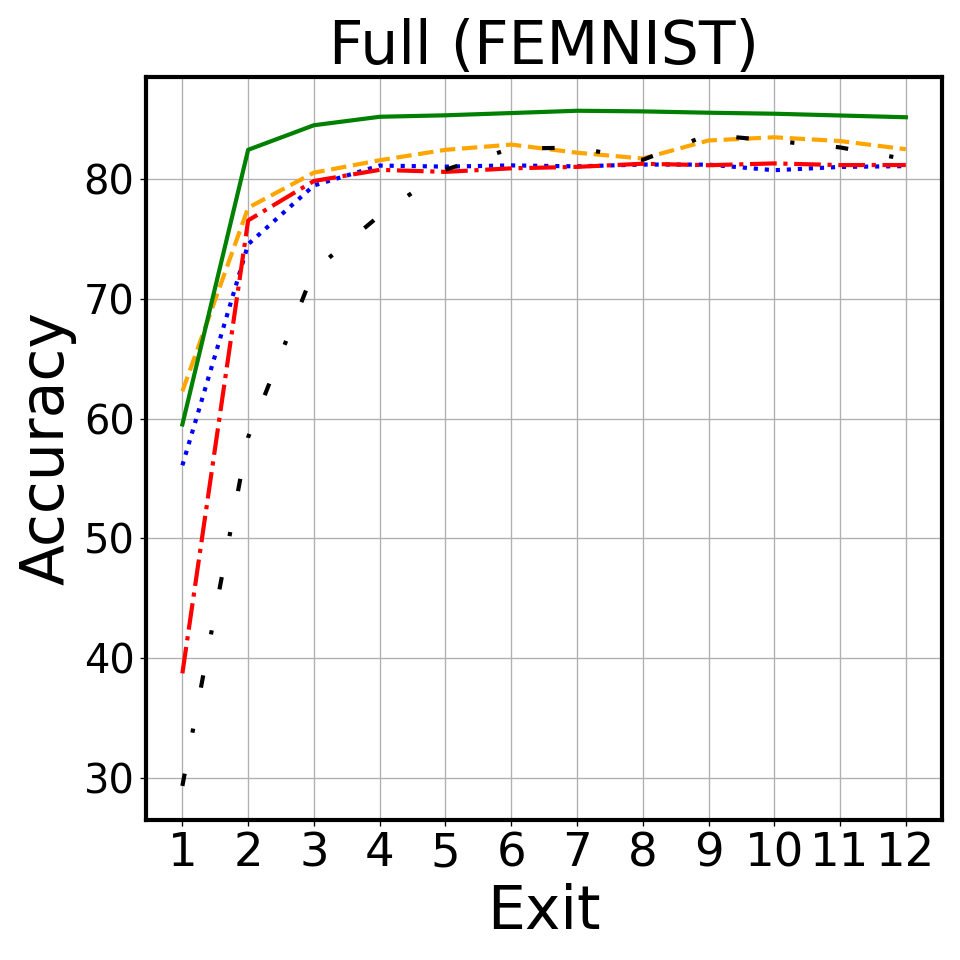}
\end{subfigure}
\begin{subfigure}{0.3\columnwidth}
    \includegraphics[trim=0 0 0 0, clip, width=0.97\columnwidth]{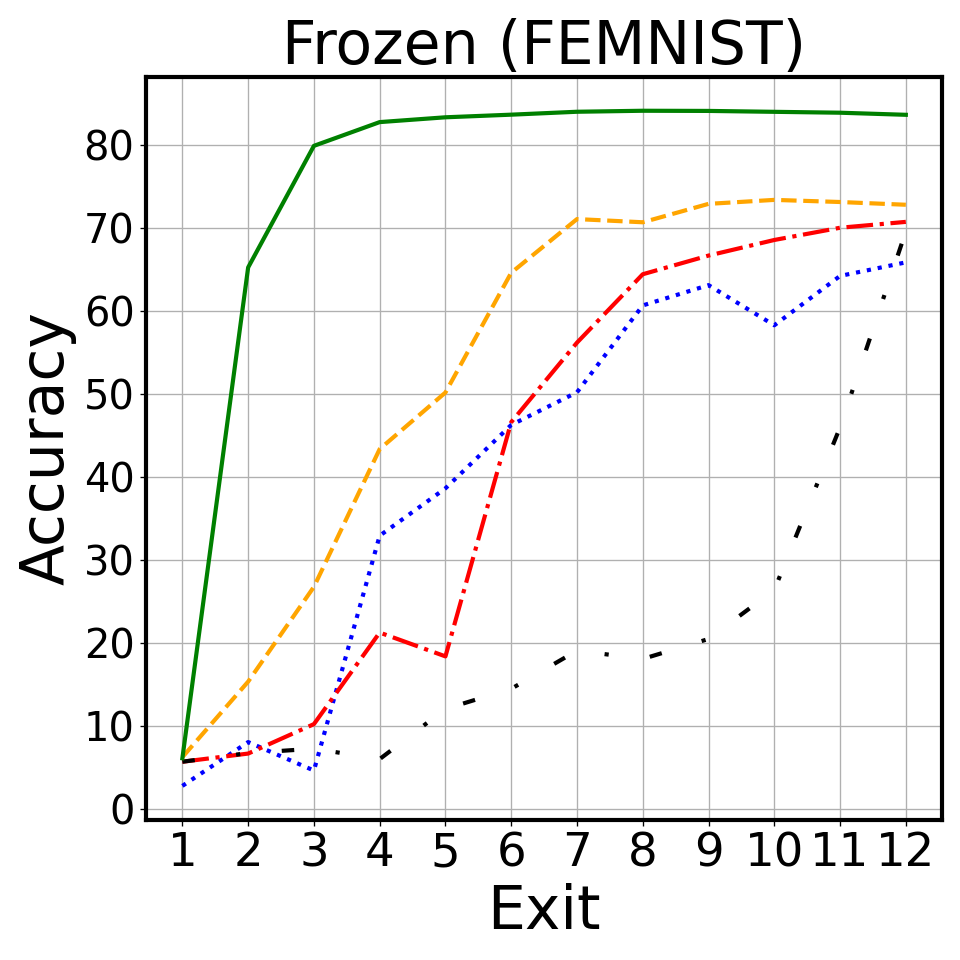}
\end{subfigure}
\begin{subfigure}{0.3\columnwidth}
    \includegraphics[trim=0 0 0 0, clip, width=0.97\columnwidth]{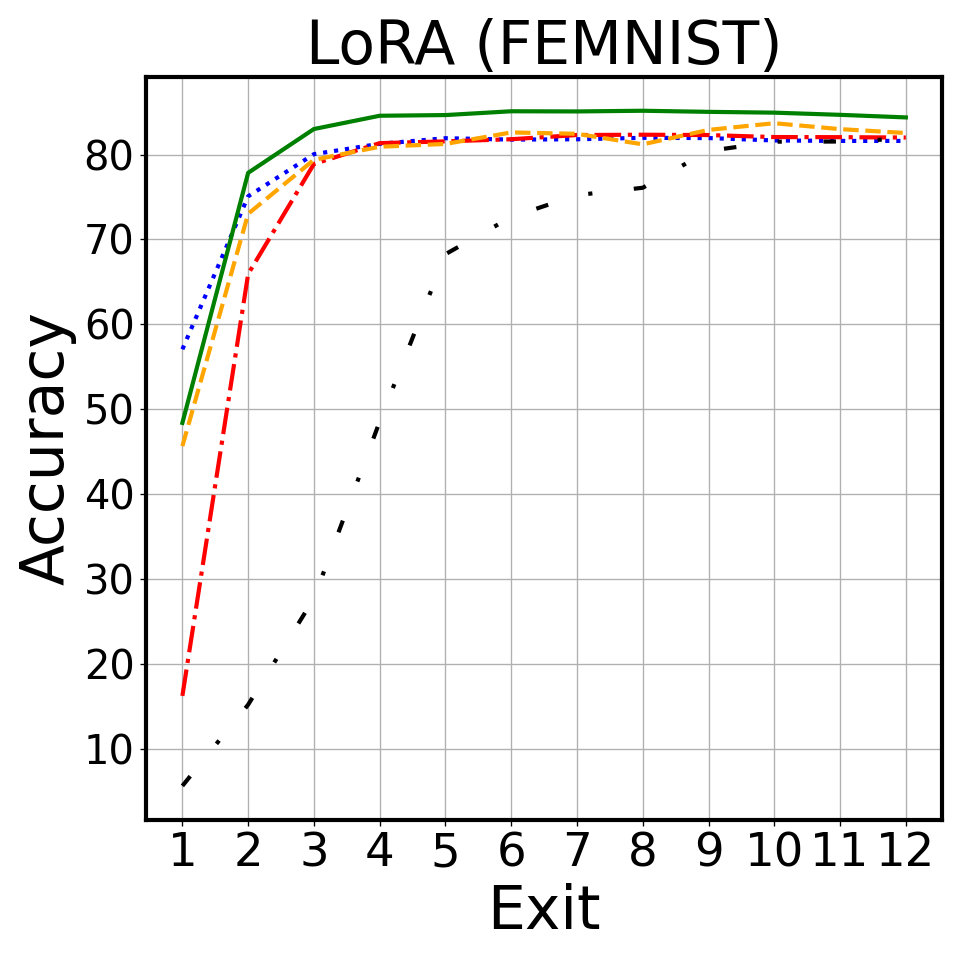}
\end{subfigure}
\begin{subfigure}{0.3\columnwidth}
    \includegraphics[trim=0 0 0 0, clip, width=0.97\columnwidth]{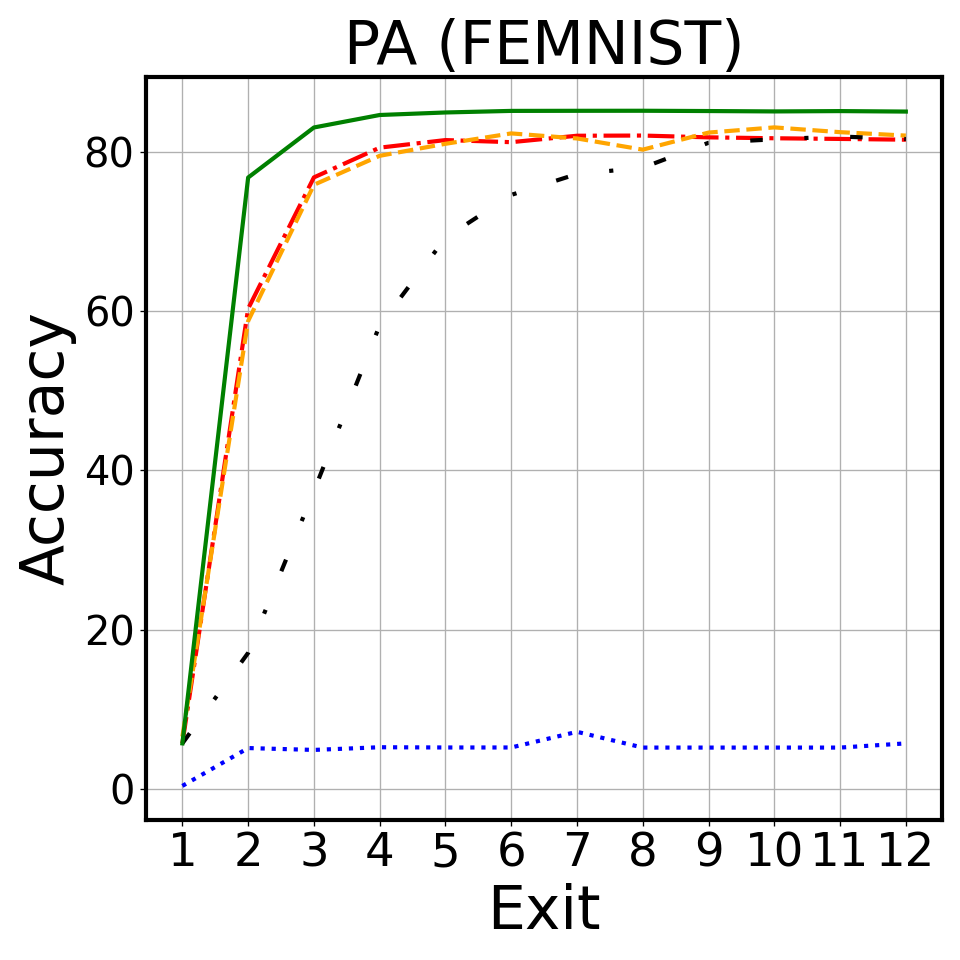}
\end{subfigure}
\begin{subfigure}{0.3\columnwidth}
    \includegraphics[trim=0 0 0 0, clip, width=0.97\columnwidth]{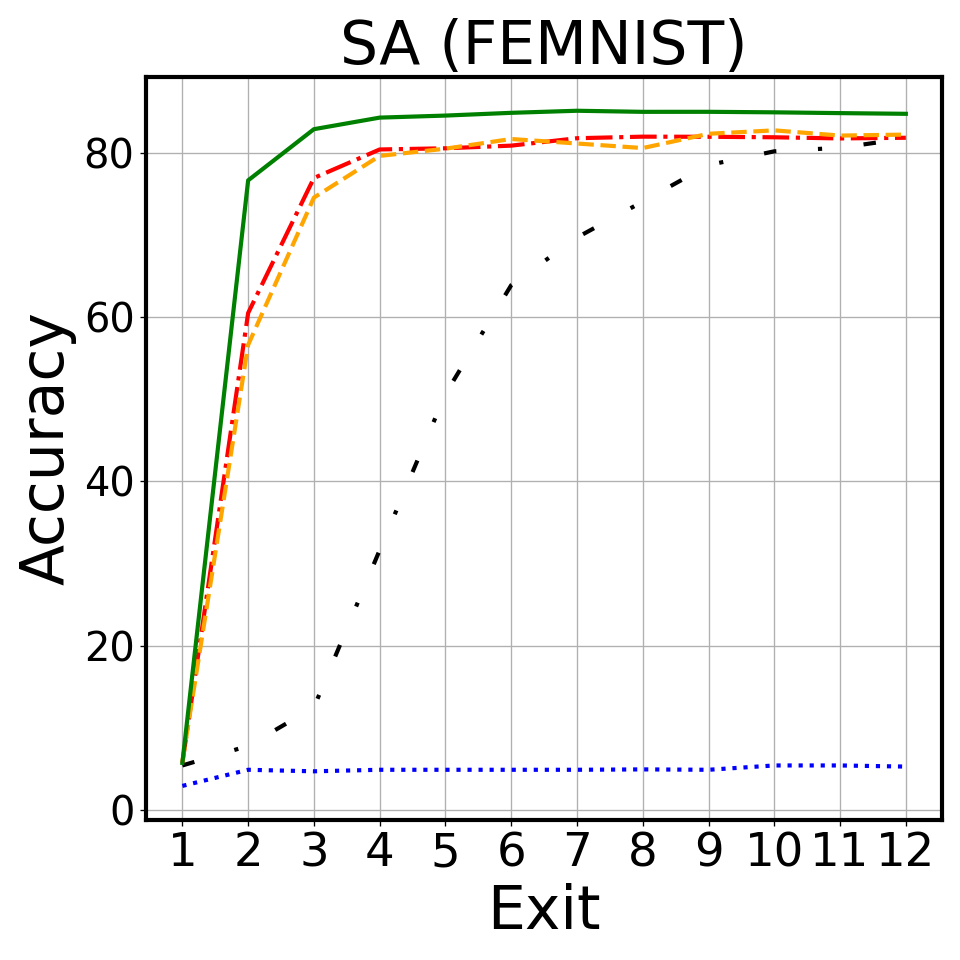}
\end{subfigure}
\begin{subfigure}{0.3\columnwidth}
    \includegraphics[trim=0 0 0 0, clip, width=0.97\columnwidth]{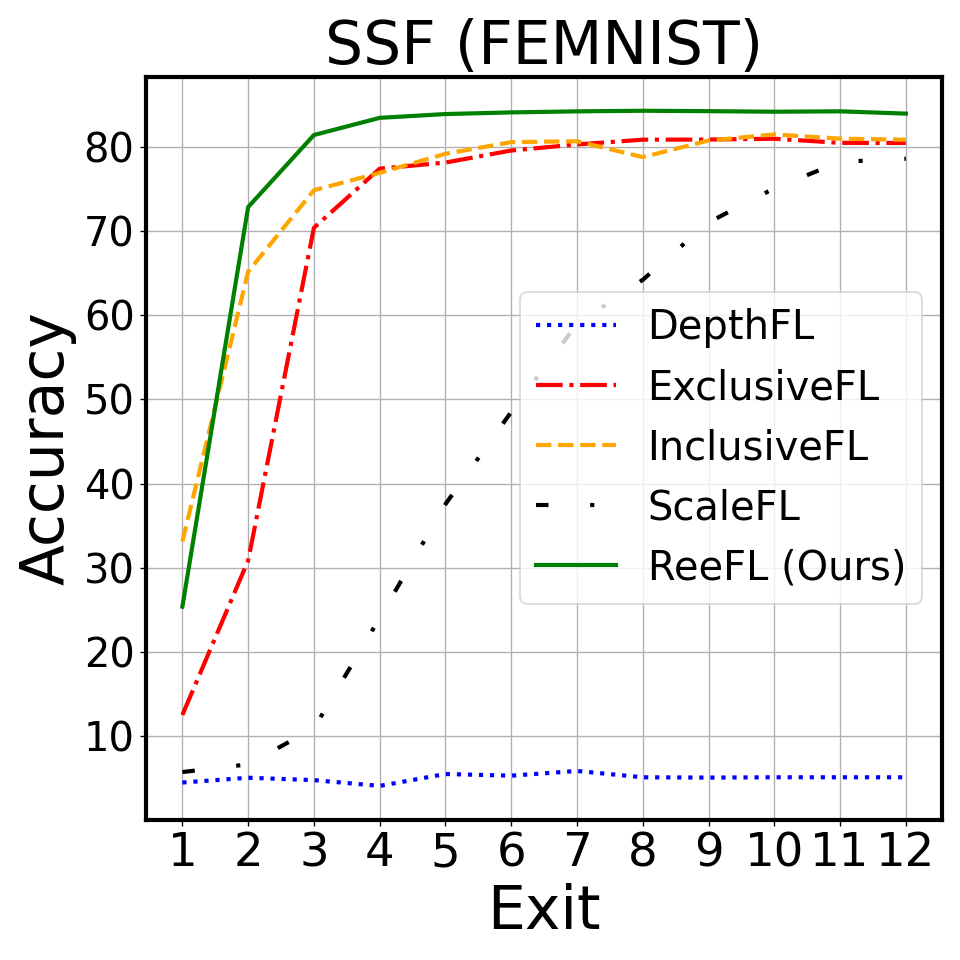}
\end{subfigure} \\

\
\vspace{-1.0em}
\caption{Fig~\ref{fig:ee_e12}'s extended results. Mean accuracy of each exit across 3 runs. }
\label{app:fig:ee_e12}
\end{figure*}

\begin{figure*}[!t]
\centering

\begin{subfigure}{0.36\columnwidth}
    \includegraphics[trim=0 0 0 0, clip, width=0.97\columnwidth]{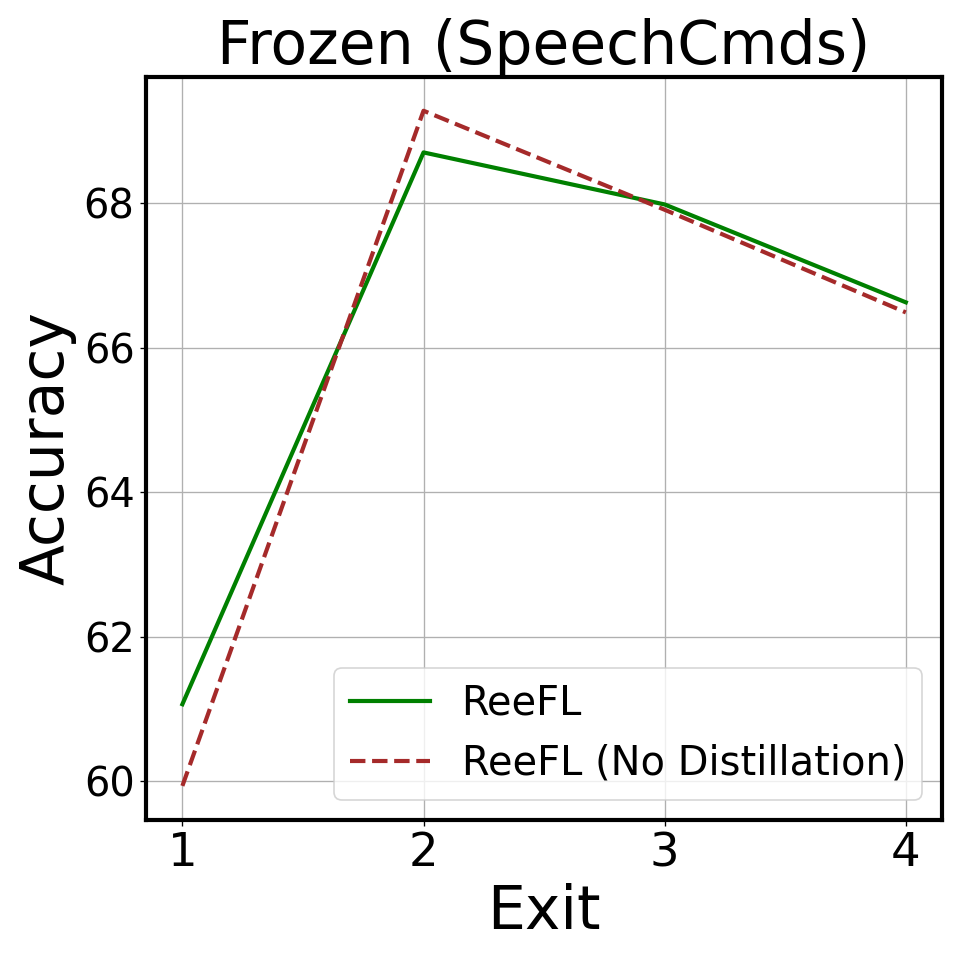}
\end{subfigure}
\begin{subfigure}{0.36\columnwidth}
    \includegraphics[trim=0 0 0 0, clip, width=0.97\columnwidth]{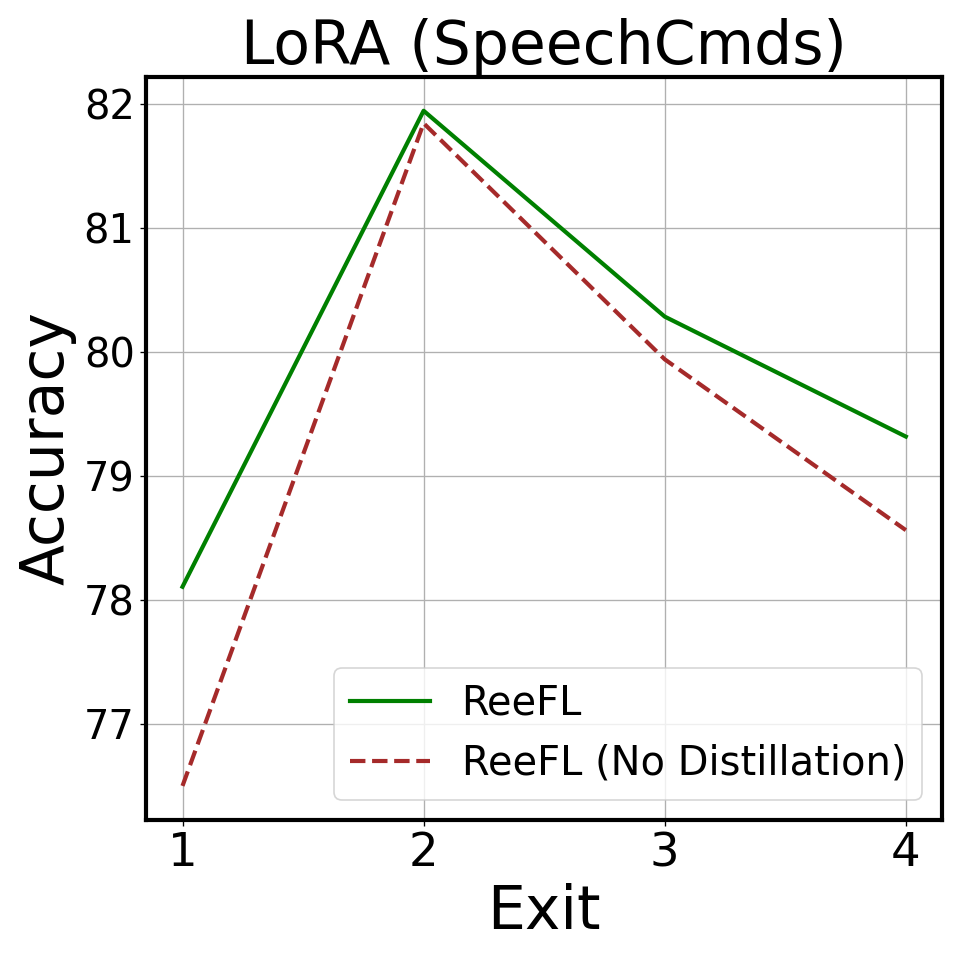}
\end{subfigure}
\begin{subfigure}{0.36\columnwidth}
    \includegraphics[trim=0 0 0 0, clip, width=0.97\columnwidth]{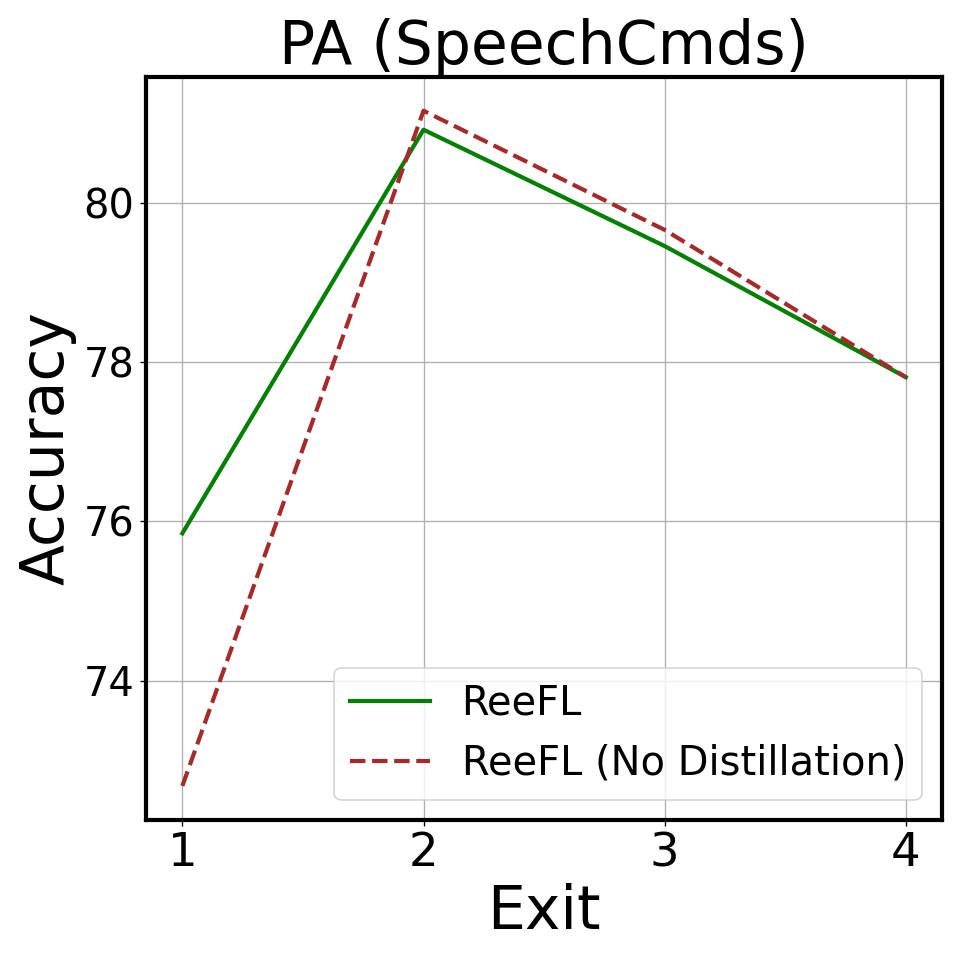}
\end{subfigure}
\begin{subfigure}{0.36\columnwidth}
    \includegraphics[trim=0 0 0 0, clip, width=0.97\columnwidth]{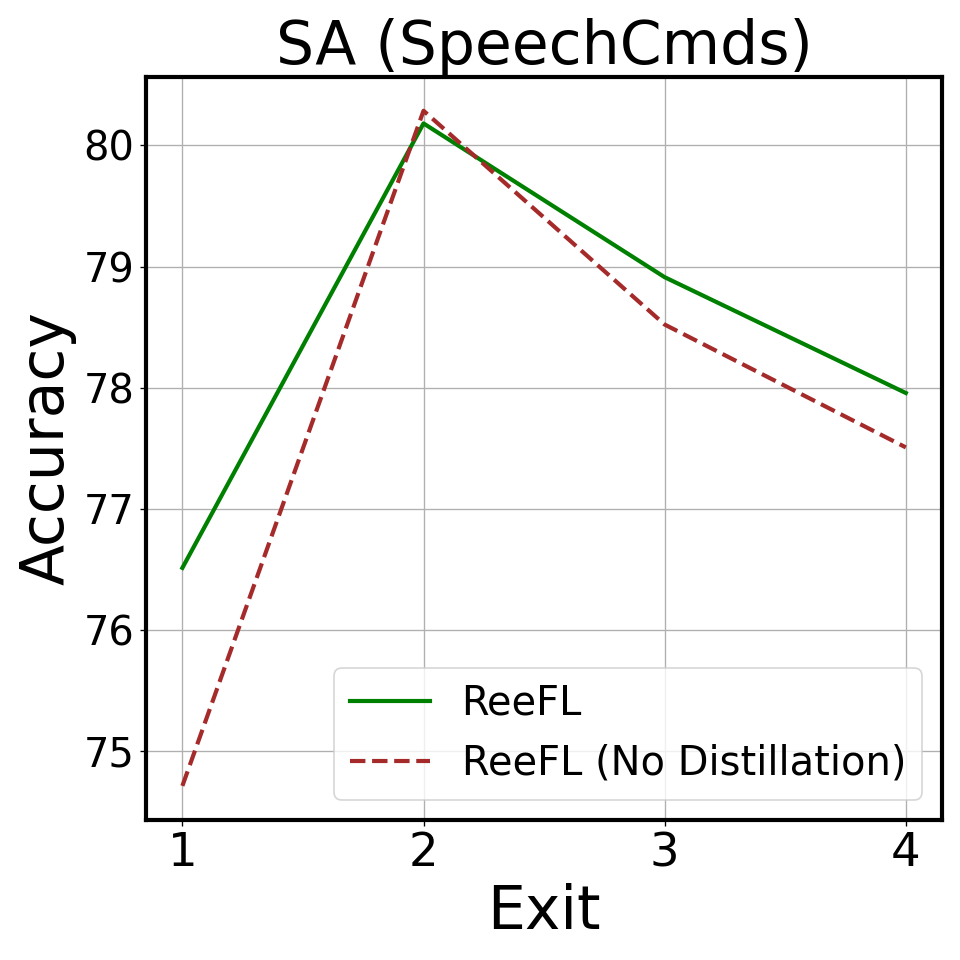}
\end{subfigure}
\begin{subfigure}{0.36\columnwidth}
    \includegraphics[trim=0 0 0 0, clip, width=0.97\columnwidth]{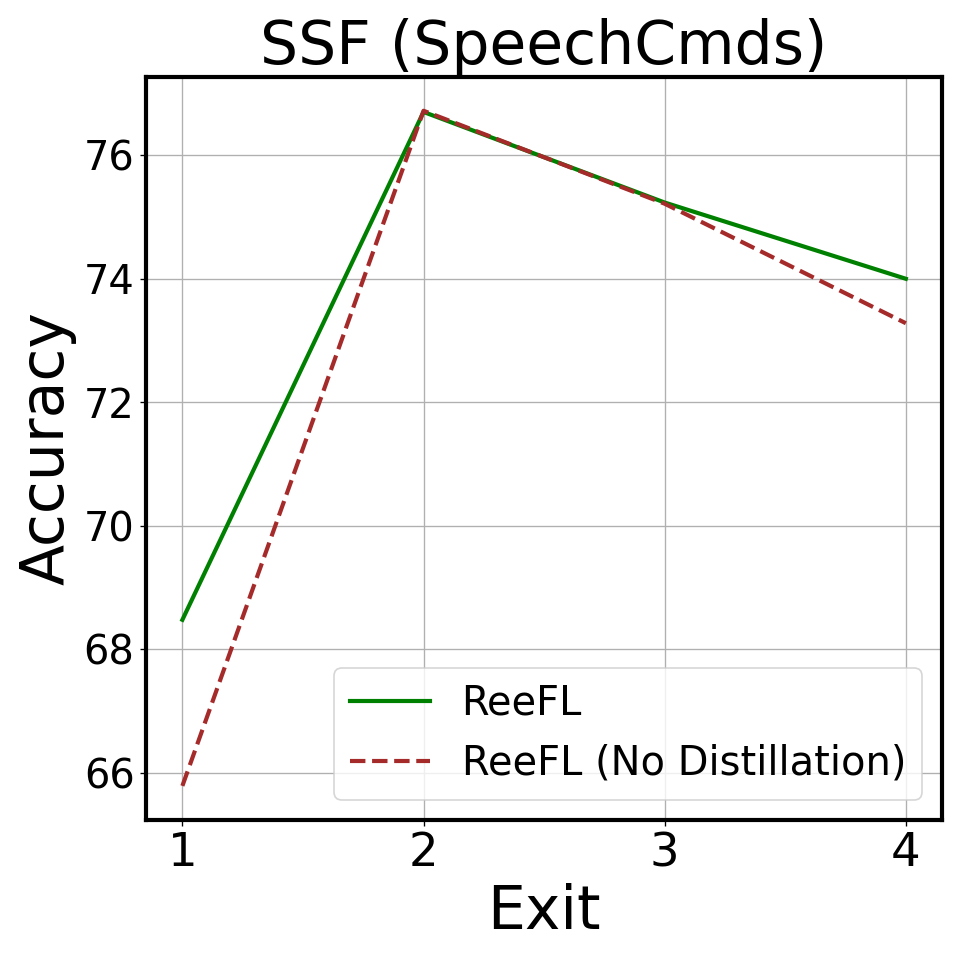}
\end{subfigure} \\

\begin{subfigure}{0.36\columnwidth}
    \includegraphics[trim=0 0 0 0, clip, width=0.97\columnwidth]{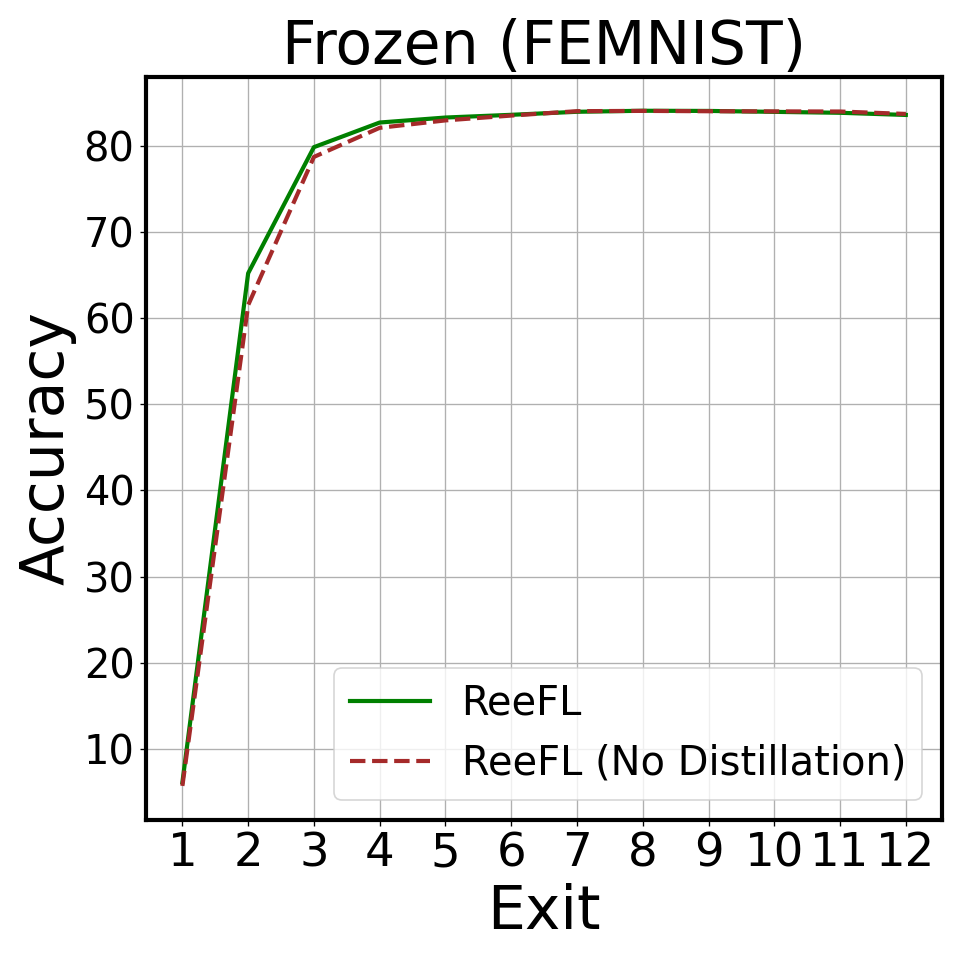}
\end{subfigure}
\begin{subfigure}{0.36\columnwidth}
    \includegraphics[trim=0 0 0 0, clip, width=0.97\columnwidth]{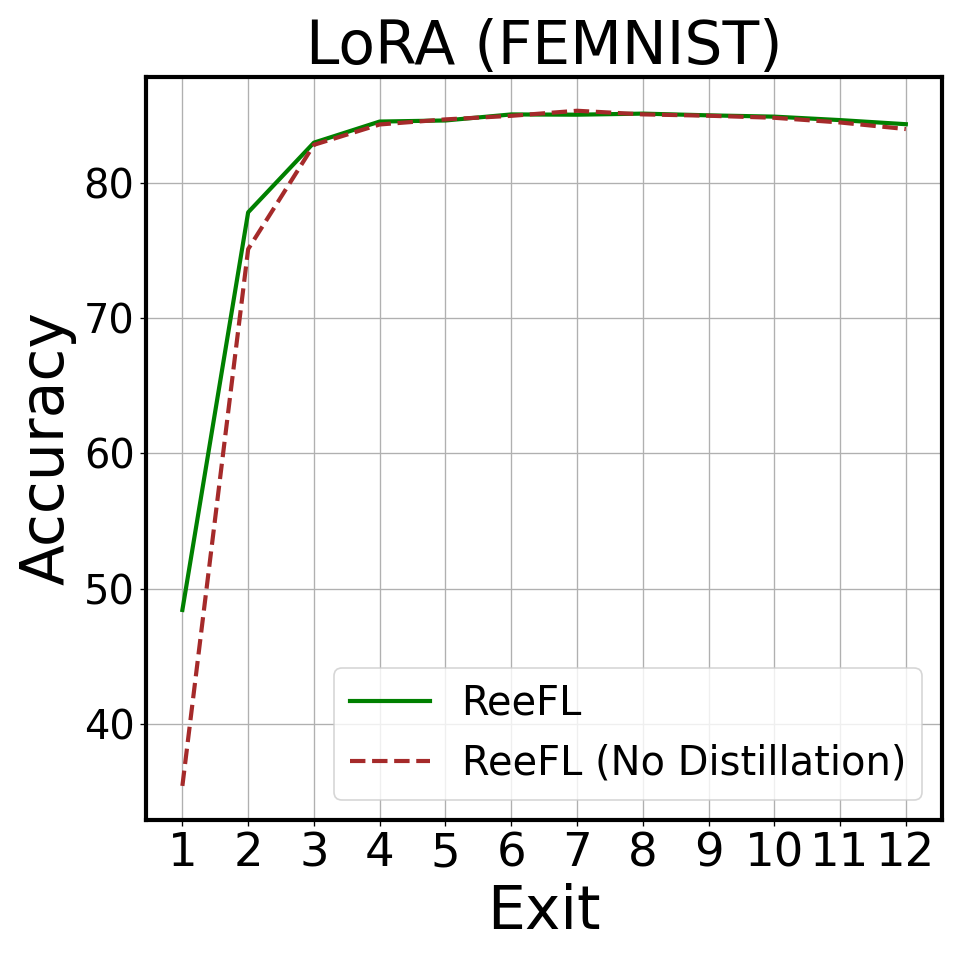}
\end{subfigure}
\begin{subfigure}{0.36\columnwidth}
    \includegraphics[trim=0 0 0 0, clip, width=0.97\columnwidth]{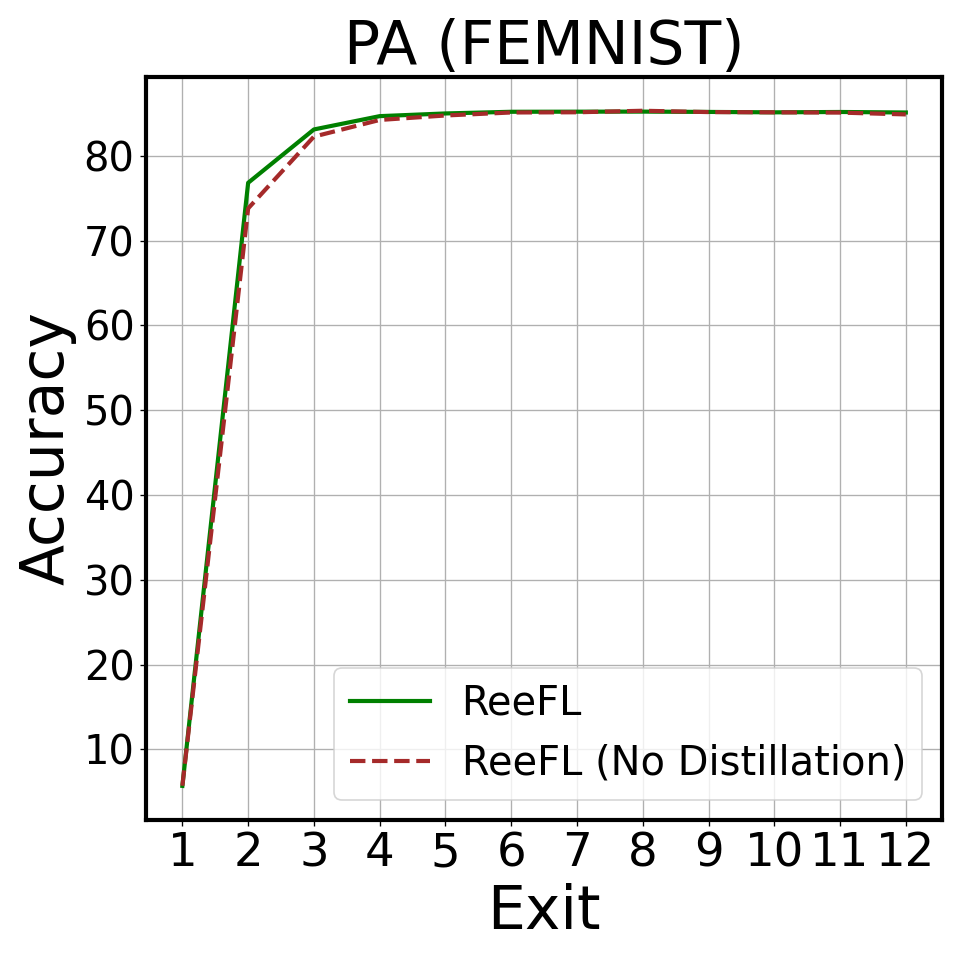}
\end{subfigure}
\begin{subfigure}{0.36\columnwidth}
    \includegraphics[trim=0 0 0 0, clip, width=0.97\columnwidth]{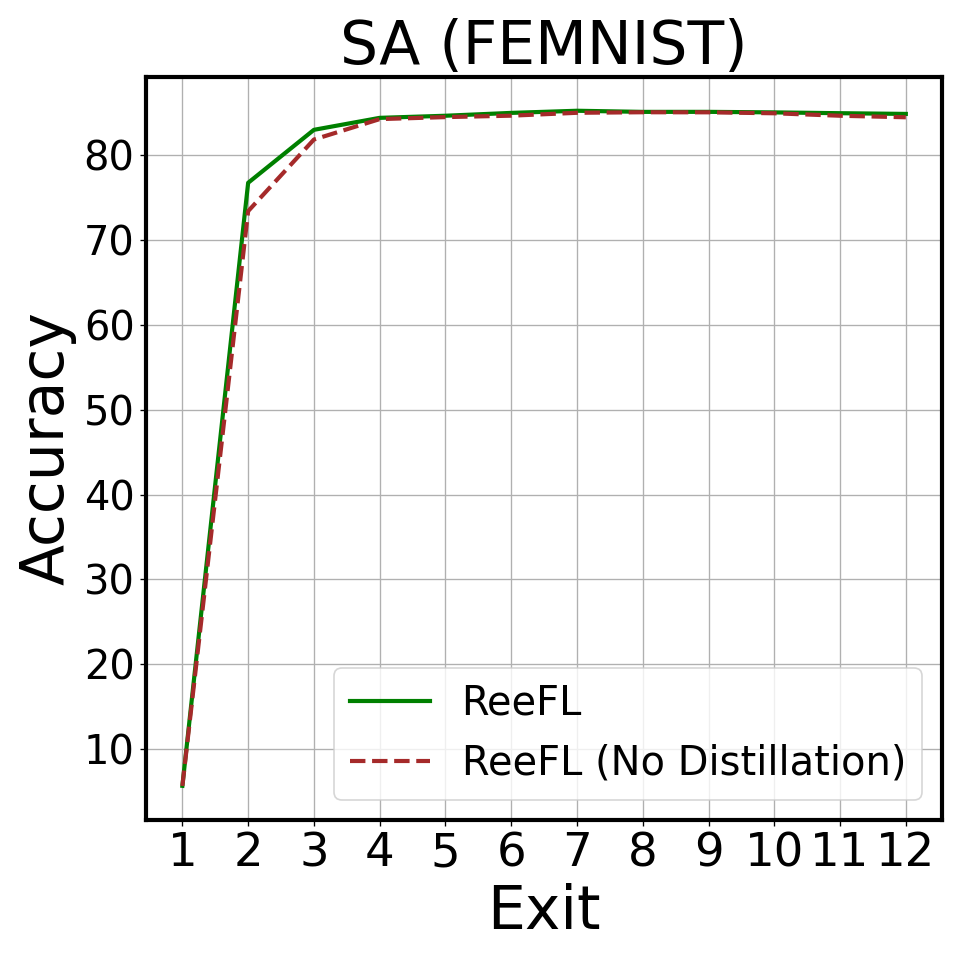}
\end{subfigure}
\begin{subfigure}{0.36\columnwidth}
    \includegraphics[trim=0 0 0 0, clip, width=0.97\columnwidth]{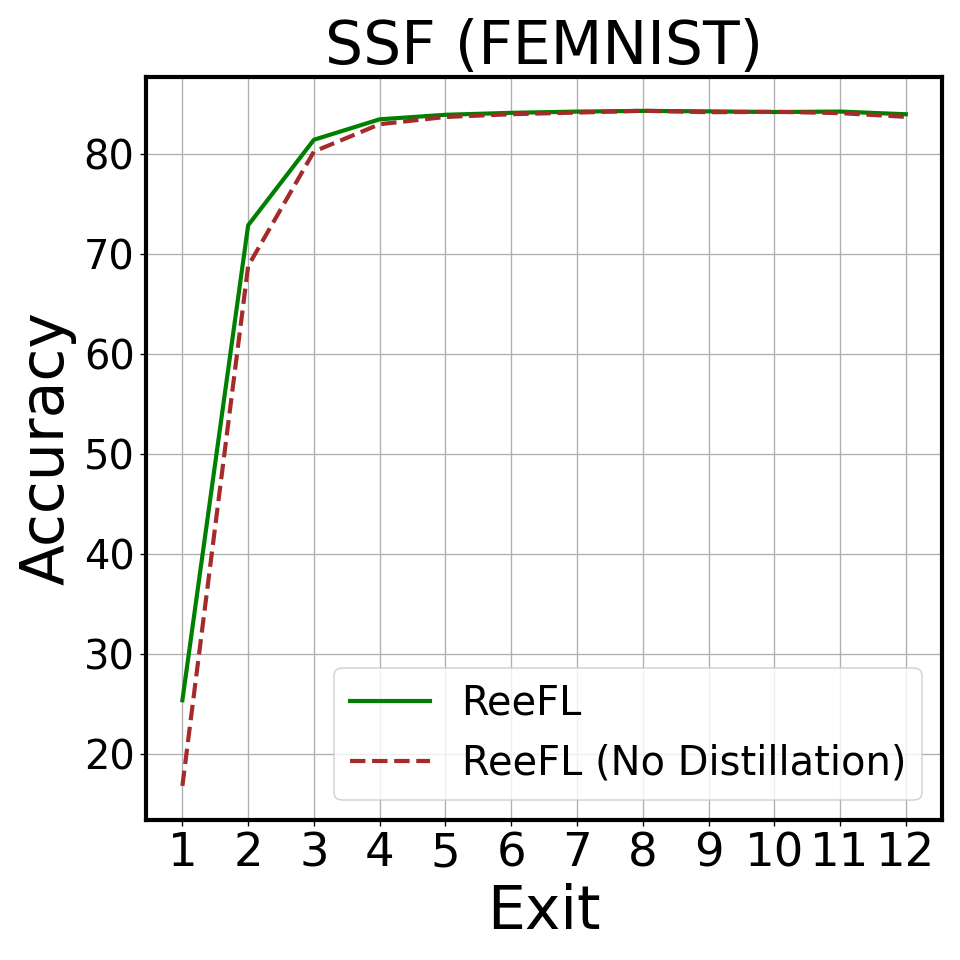}
\end{subfigure} \\

\begin{subfigure}{0.36\columnwidth}
    \includegraphics[trim=0 0 0 0, clip, width=0.97\columnwidth]{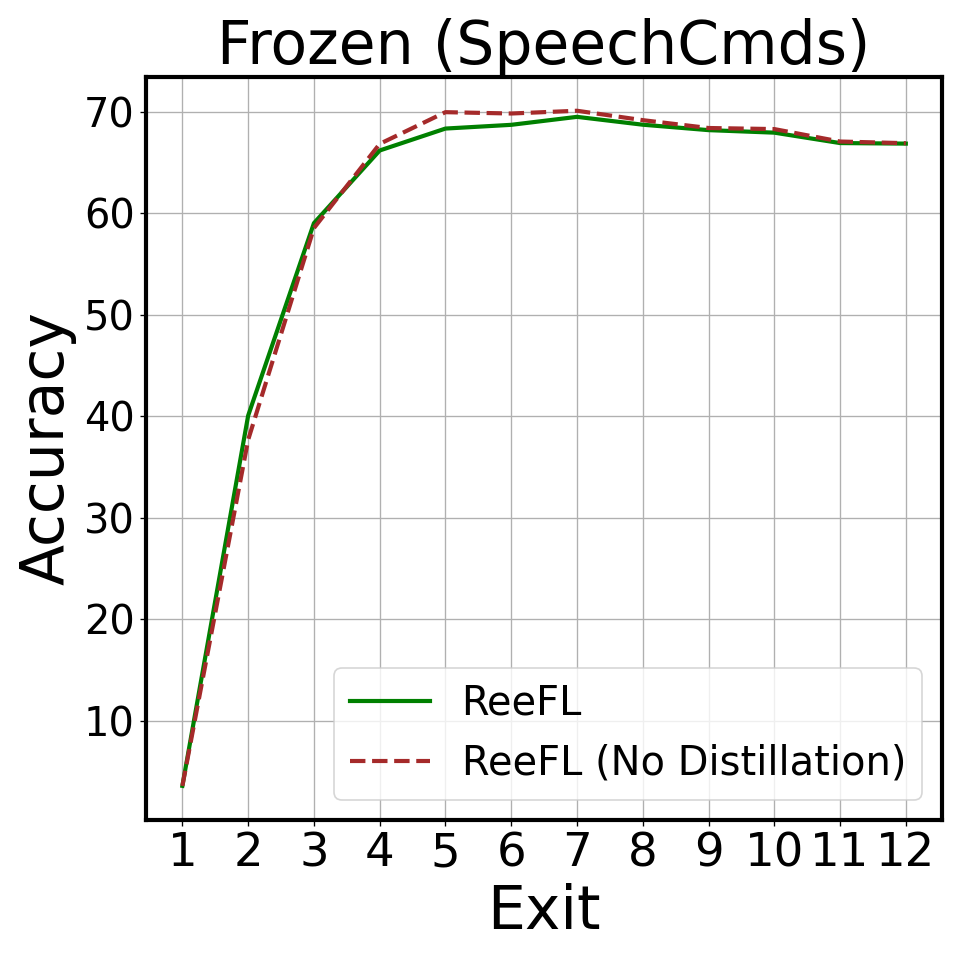}
\end{subfigure}
\begin{subfigure}{0.36\columnwidth}
    \includegraphics[trim=0 0 0 0, clip, width=0.97\columnwidth]{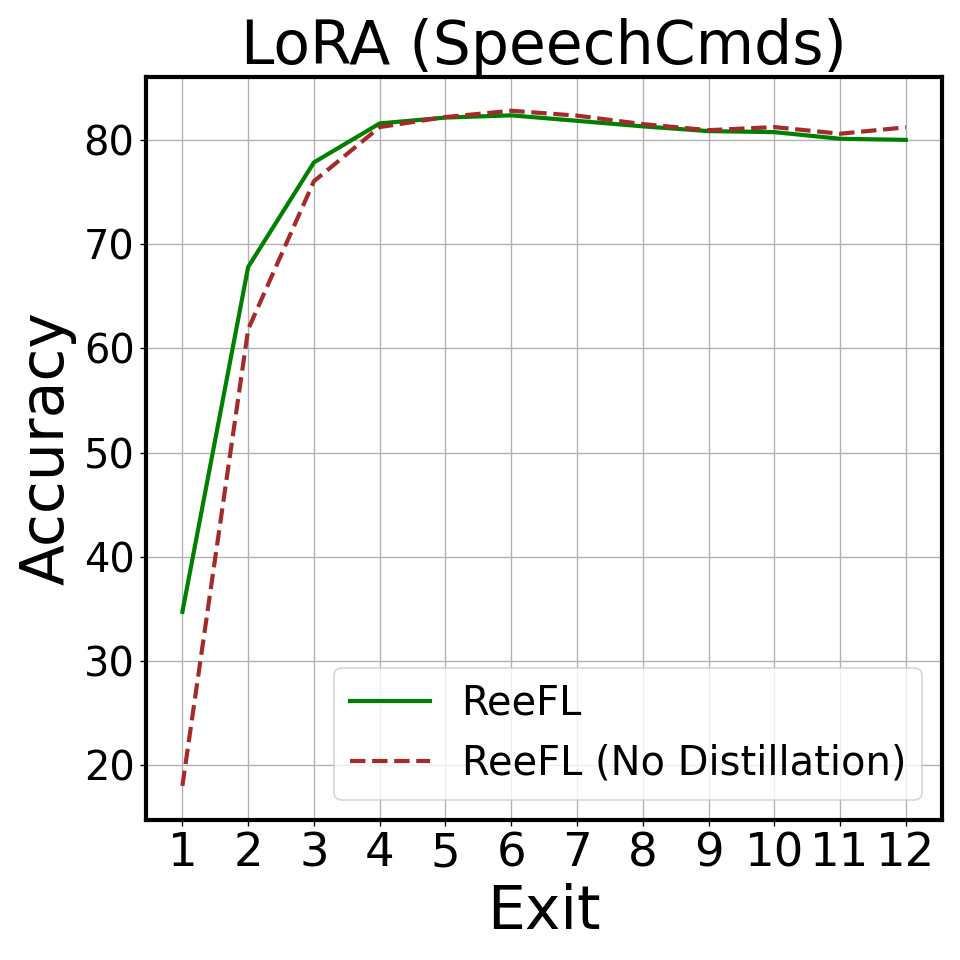}
\end{subfigure}
\begin{subfigure}{0.36\columnwidth}
    \includegraphics[trim=0 0 0 0, clip, width=0.97\columnwidth]{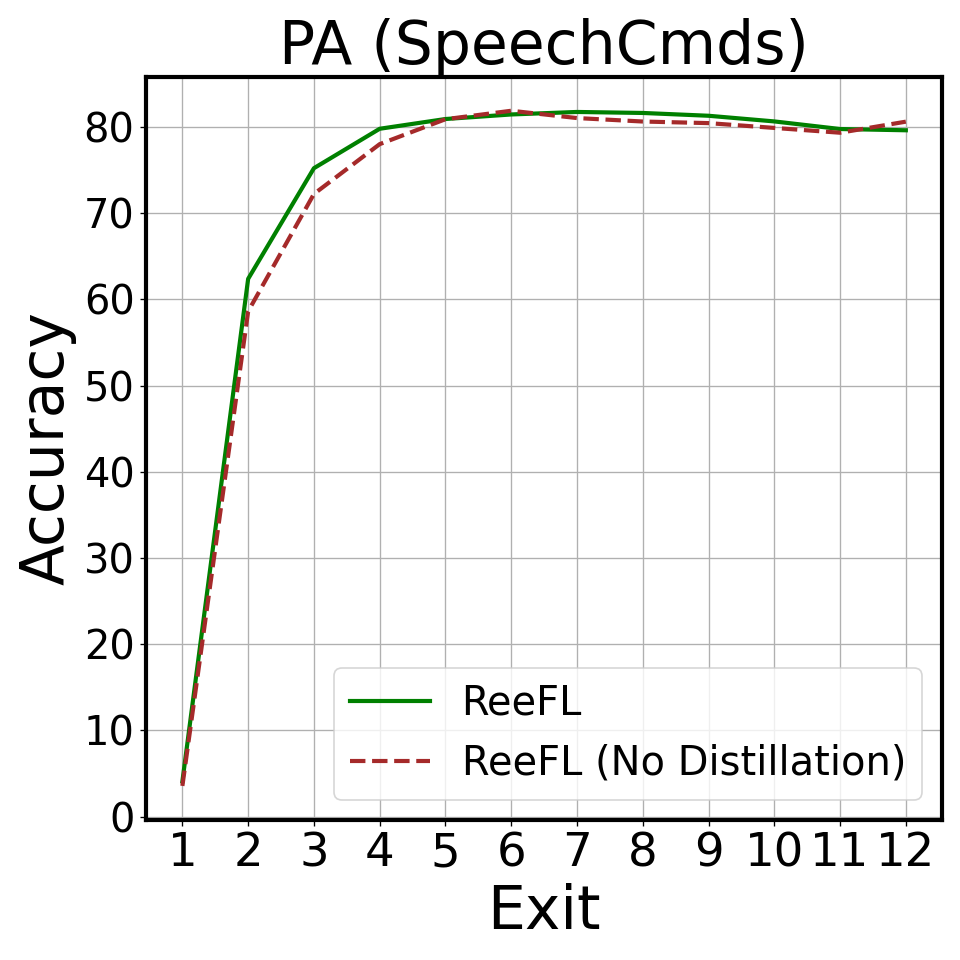}
\end{subfigure}
\begin{subfigure}{0.36\columnwidth}
    \includegraphics[trim=0 0 0 0, clip, width=0.97\columnwidth]{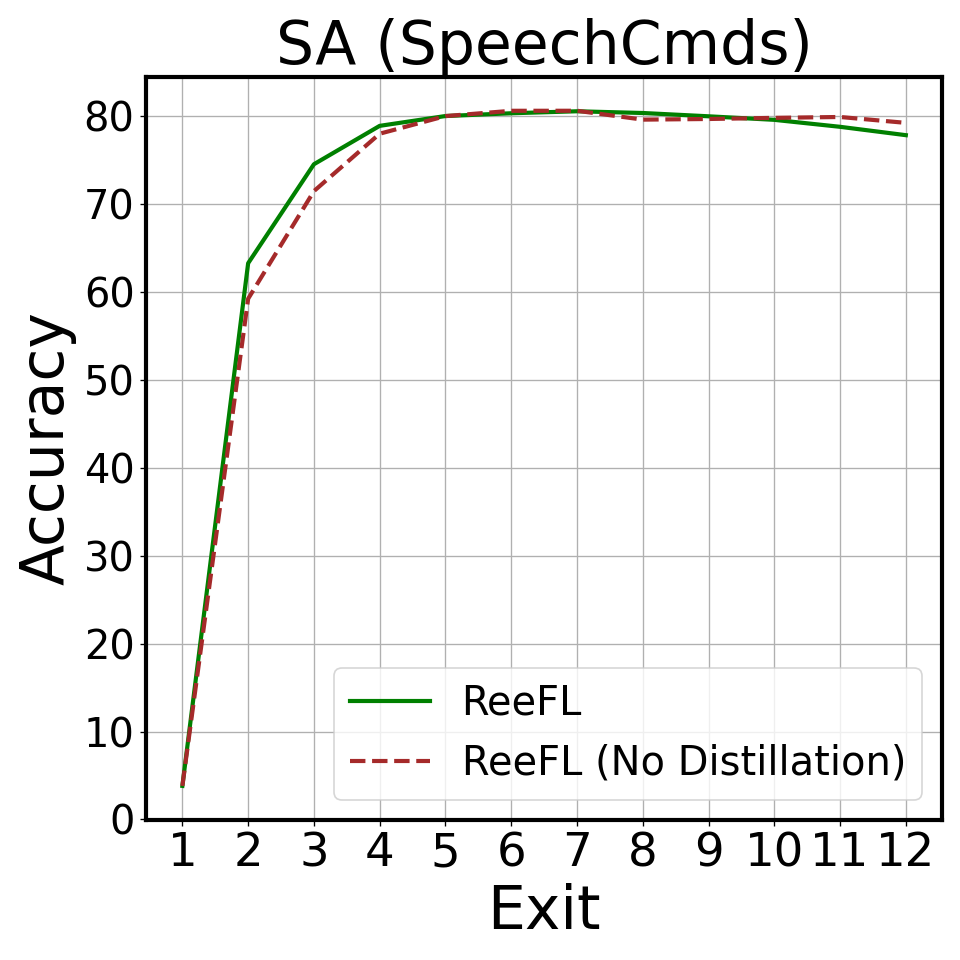}
\end{subfigure}
\begin{subfigure}{0.36\columnwidth}
    \includegraphics[trim=0 0 0 0, clip, width=0.97\columnwidth]{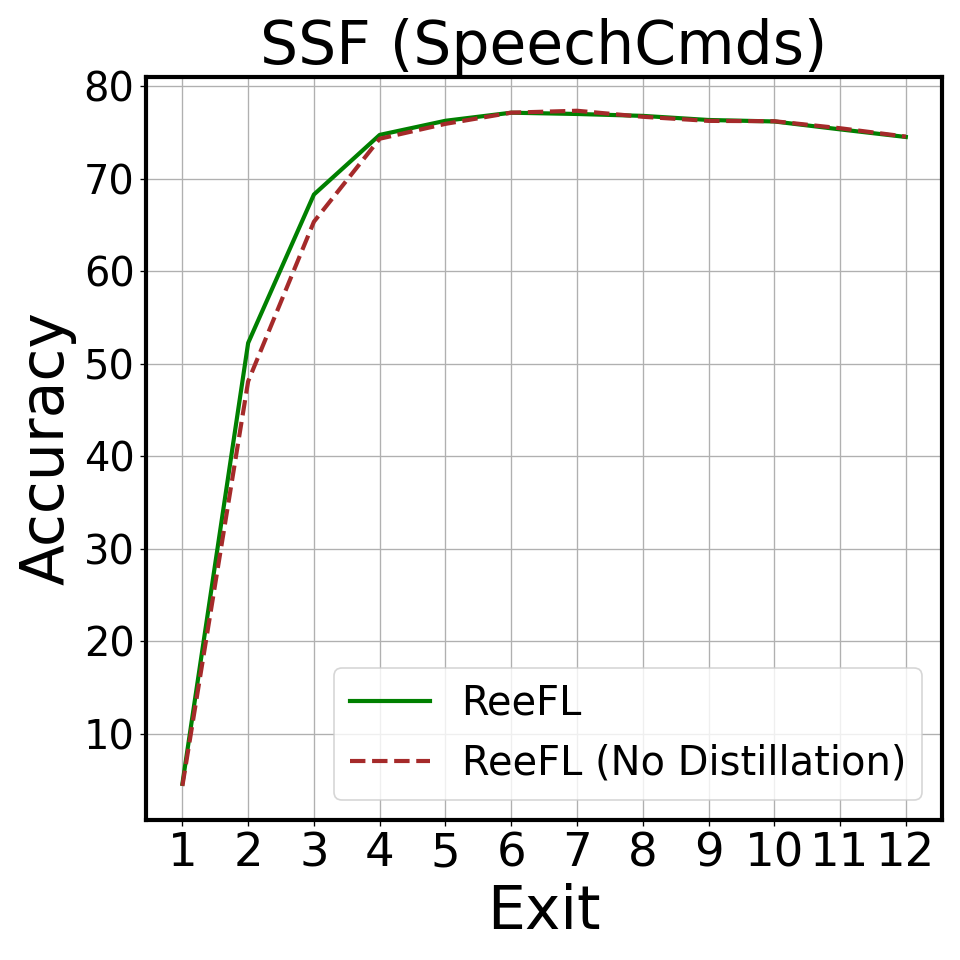}
\end{subfigure} \\
\vspace{-1.0em}
\caption{Impact of \method{}'s proposed knowledge distillation on FEMNIST and SpeechCommands ($4$ \& $12$ exits).}
\label{fig:ablation1_ee_e12}
\end{figure*}

\begin{figure*}[t]
\centering

\begin{subfigure}{0.3\columnwidth}
    \includegraphics[trim=0 0 0 0, clip, width=0.97\columnwidth]{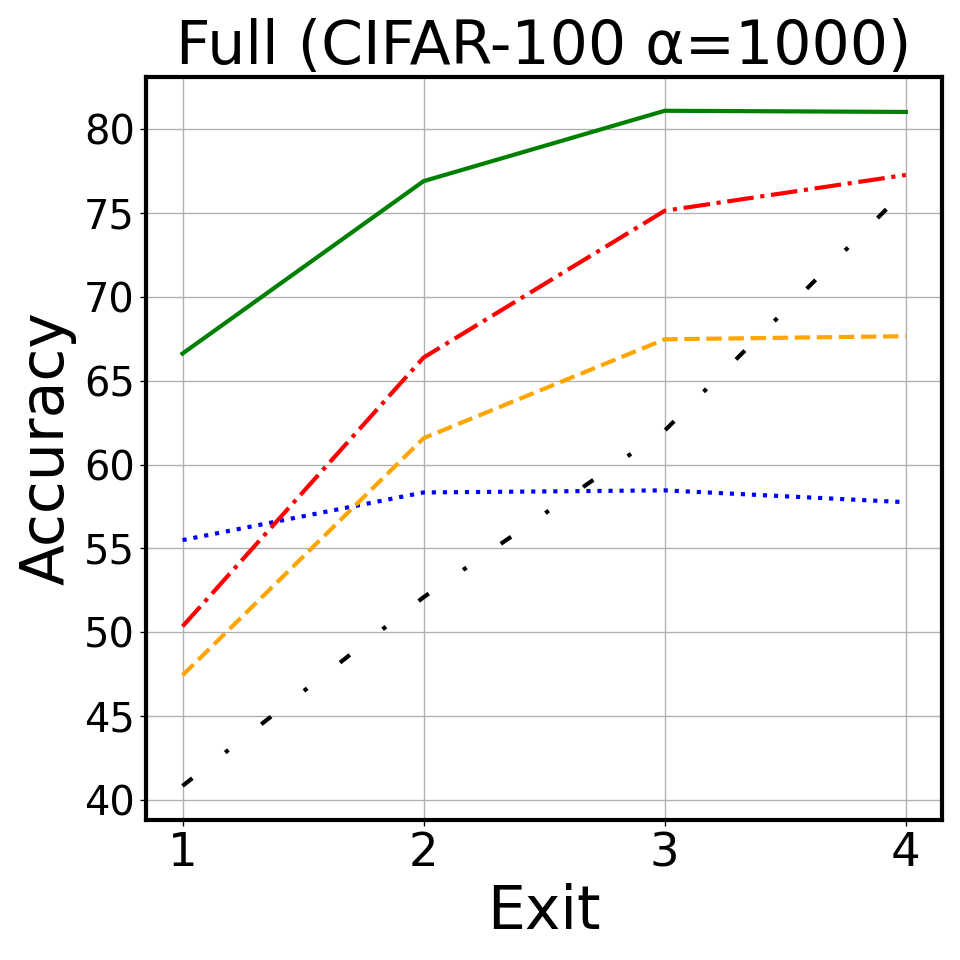}
\end{subfigure}
\begin{subfigure}{0.3\columnwidth}
    \includegraphics[trim=0 0 0 0, clip, width=0.97\columnwidth]{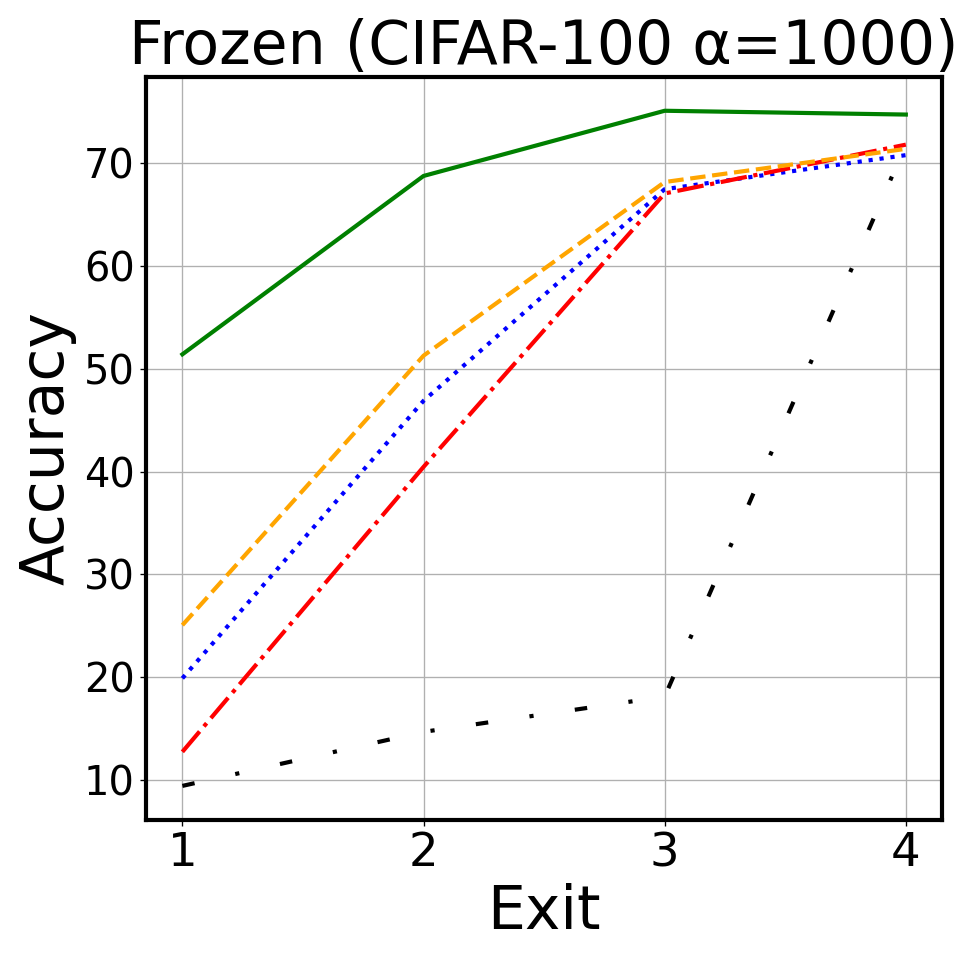}
\end{subfigure}
\begin{subfigure}{0.3\columnwidth}
    \includegraphics[trim=0 0 0 0, clip, width=0.97\columnwidth]{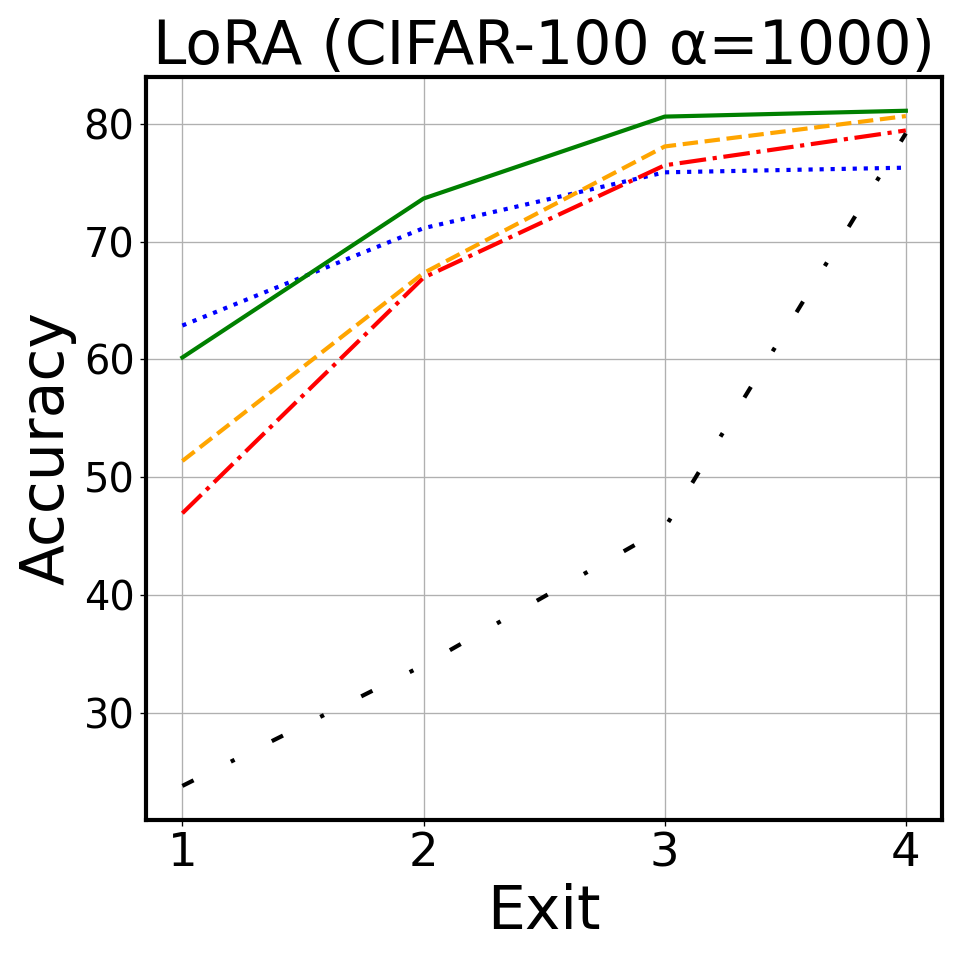}
\end{subfigure}
\begin{subfigure}{0.3\columnwidth}
    \includegraphics[trim=0 0 0 0, clip, width=0.97\columnwidth]{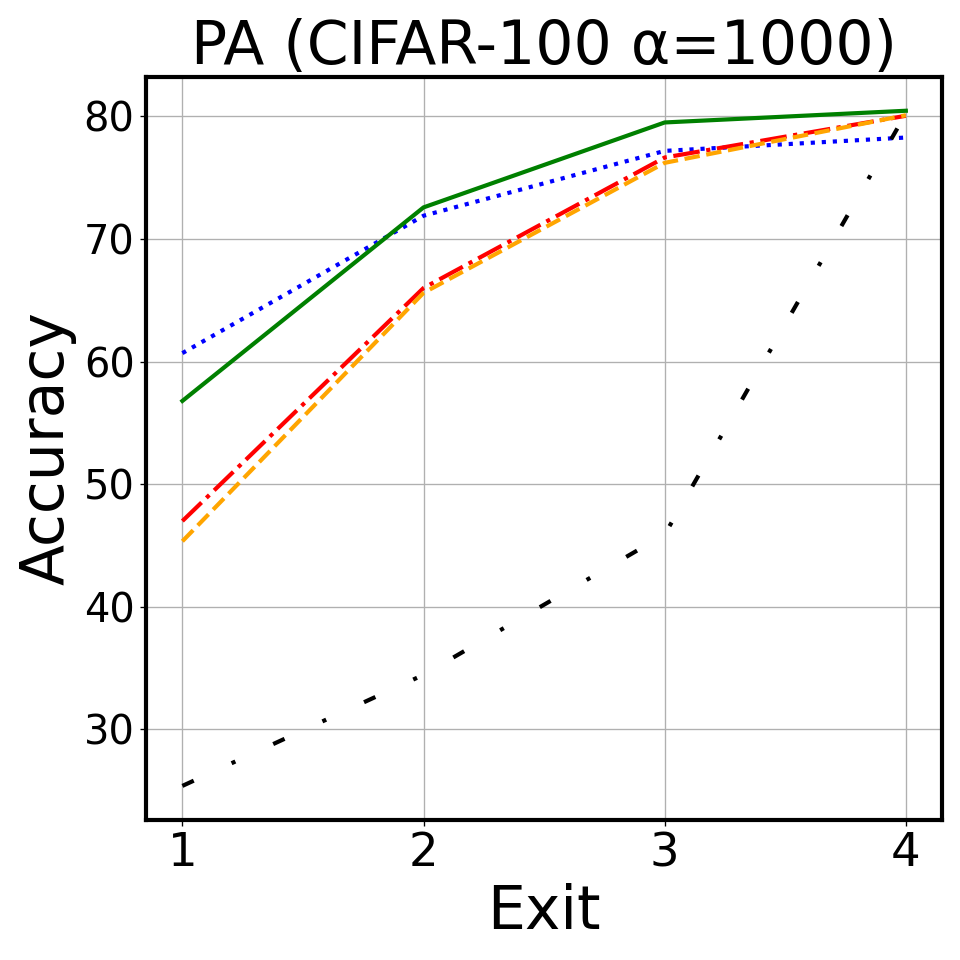}
\end{subfigure}
\begin{subfigure}{0.3\columnwidth}
    \includegraphics[trim=0 0 0 0, clip, width=0.97\columnwidth]{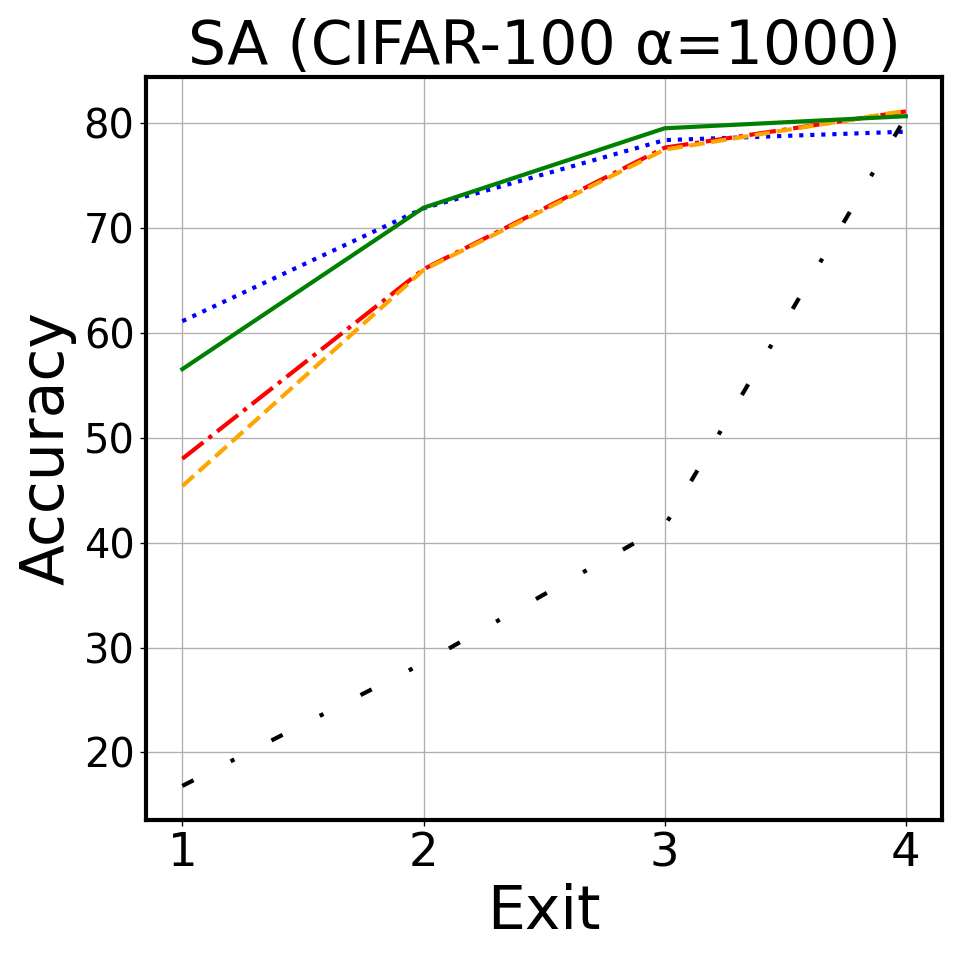}
\end{subfigure}
\begin{subfigure}{0.3\columnwidth}
    \includegraphics[trim=0 0 0 0, clip, width=0.97\columnwidth]{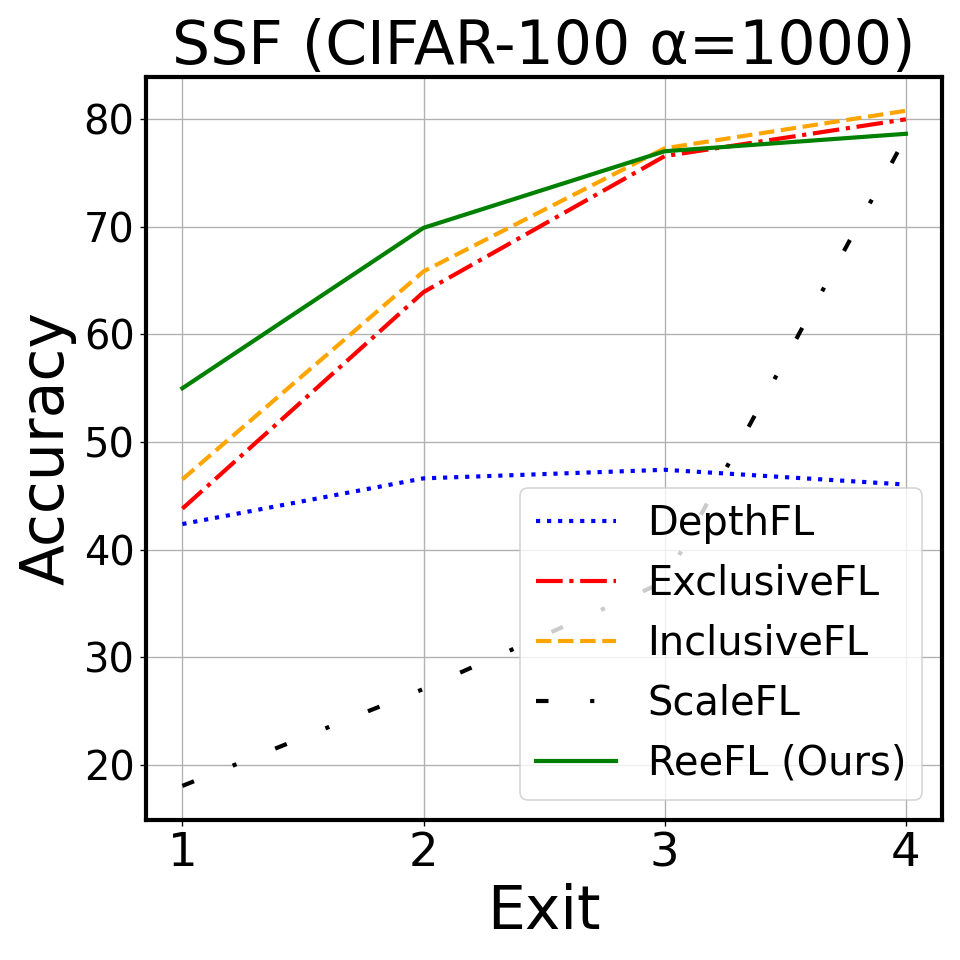}
\end{subfigure} \\

\begin{subfigure}{0.3\columnwidth}
    \includegraphics[trim=0 0 0 0, clip, width=0.97\columnwidth]{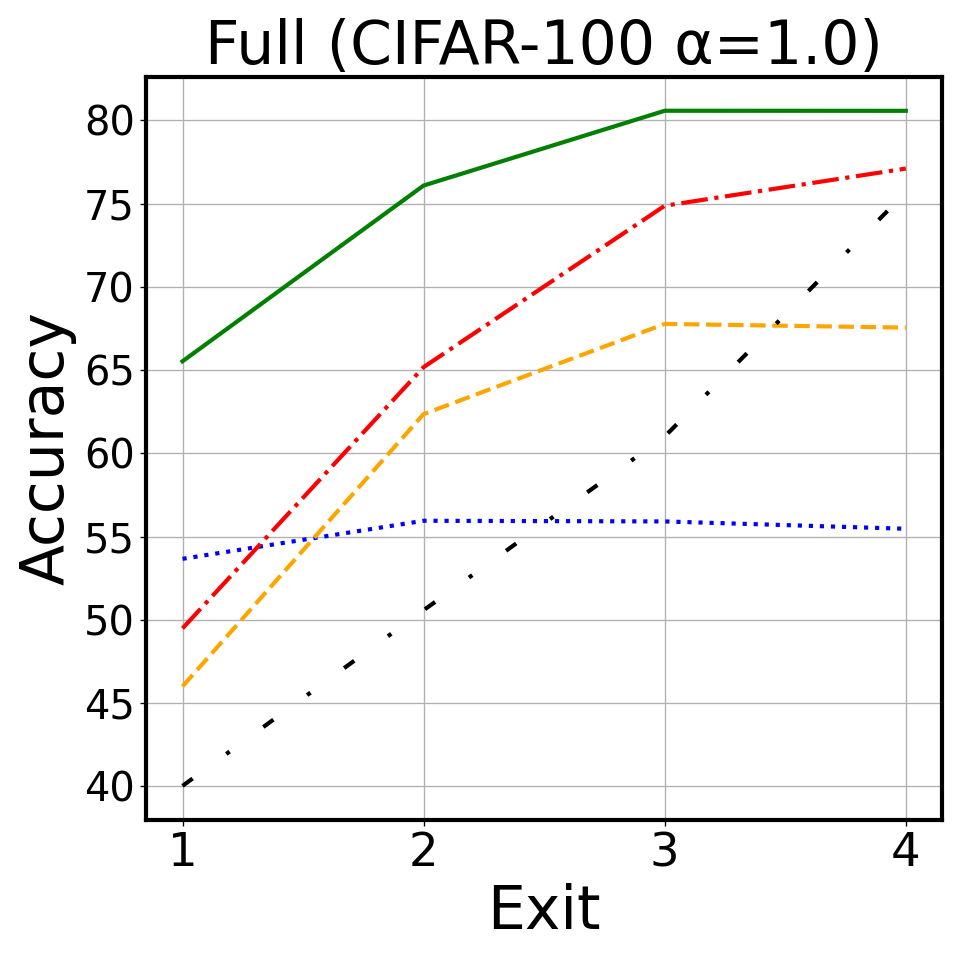}
\end{subfigure}
\begin{subfigure}{0.3\columnwidth}
    \includegraphics[trim=0 0 0 0, clip, width=0.97\columnwidth]{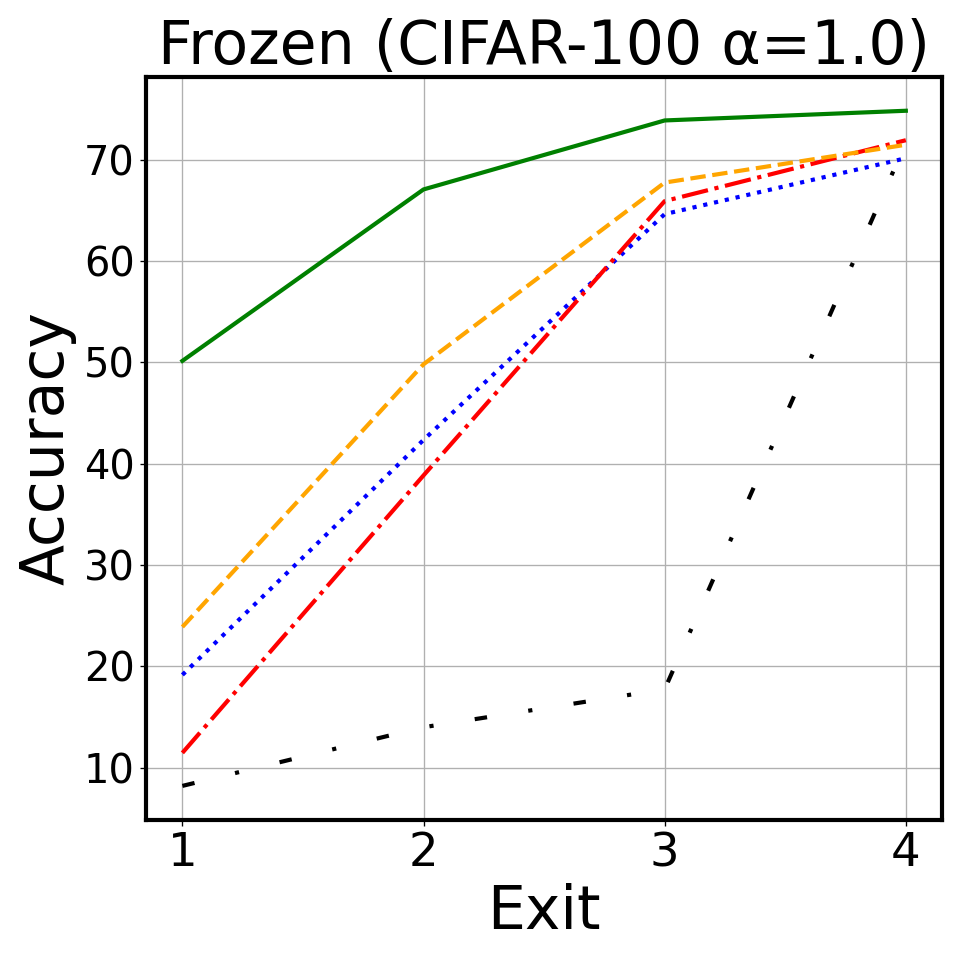}
\end{subfigure}
\begin{subfigure}{0.3\columnwidth}
    \includegraphics[trim=0 0 0 0, clip, width=0.97\columnwidth]{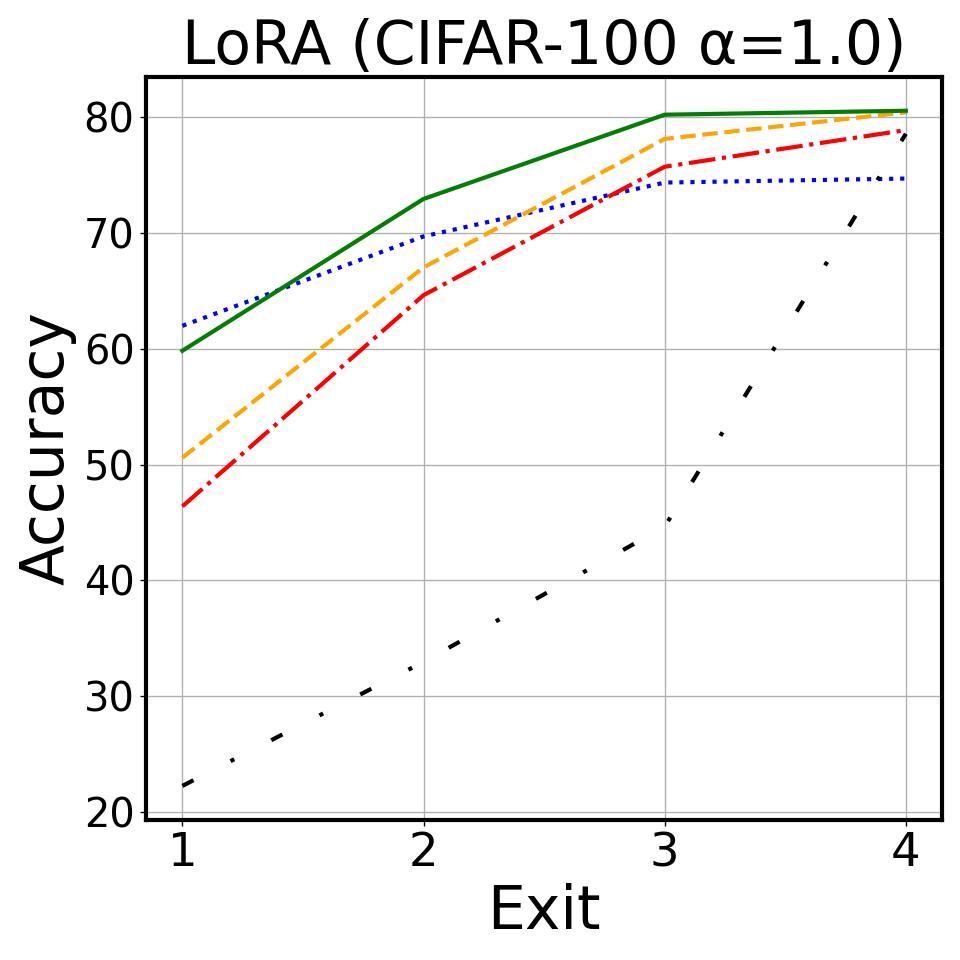}
\end{subfigure}
\begin{subfigure}{0.3\columnwidth}
    \includegraphics[trim=0 0 0 0, clip, width=0.97\columnwidth]{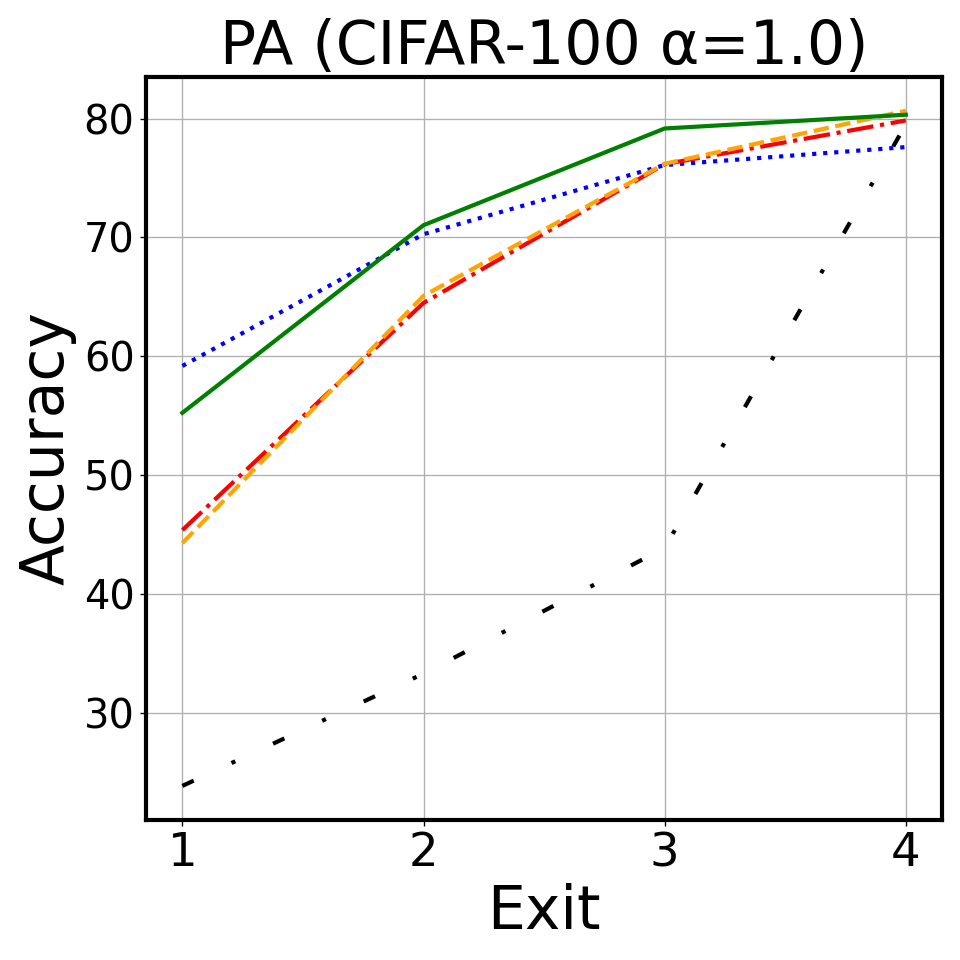}
\end{subfigure}
\begin{subfigure}{0.3\columnwidth}
    \includegraphics[trim=0 0 0 0, clip, width=0.97\columnwidth]{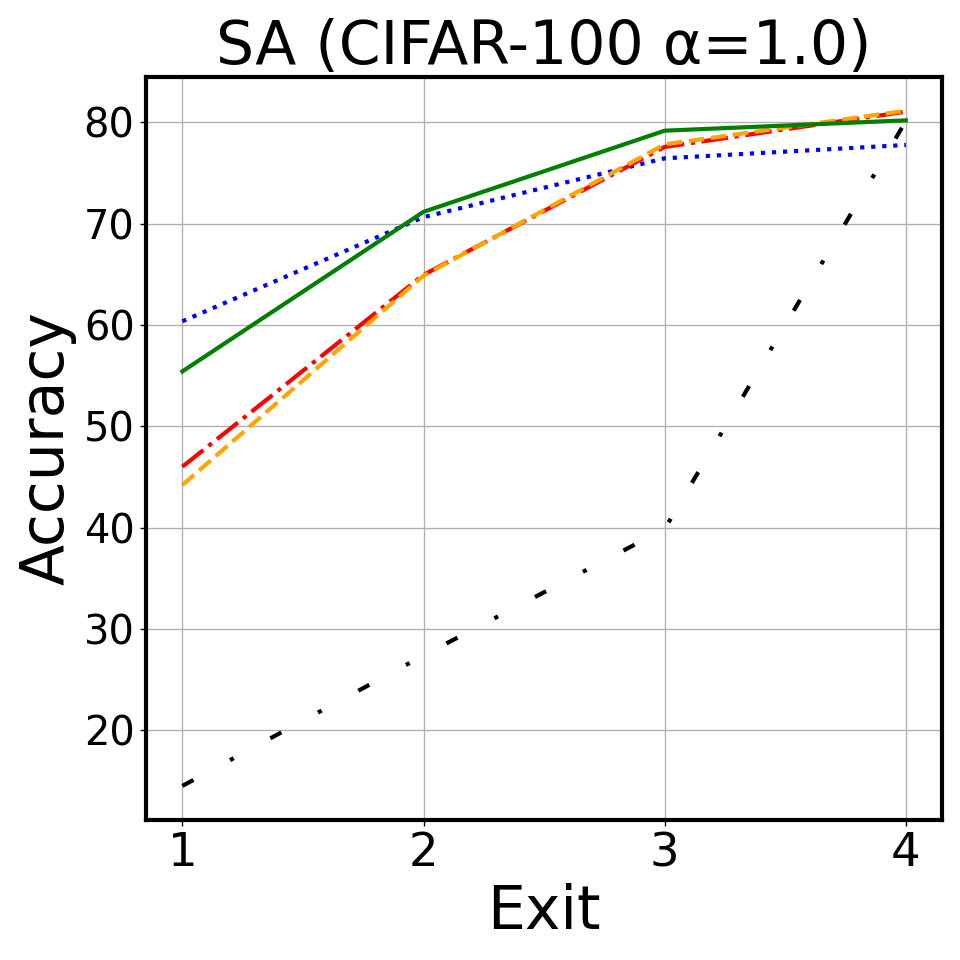}
\end{subfigure}
\begin{subfigure}{0.3\columnwidth}
    \includegraphics[trim=0 0 0 0, clip, width=0.97\columnwidth]{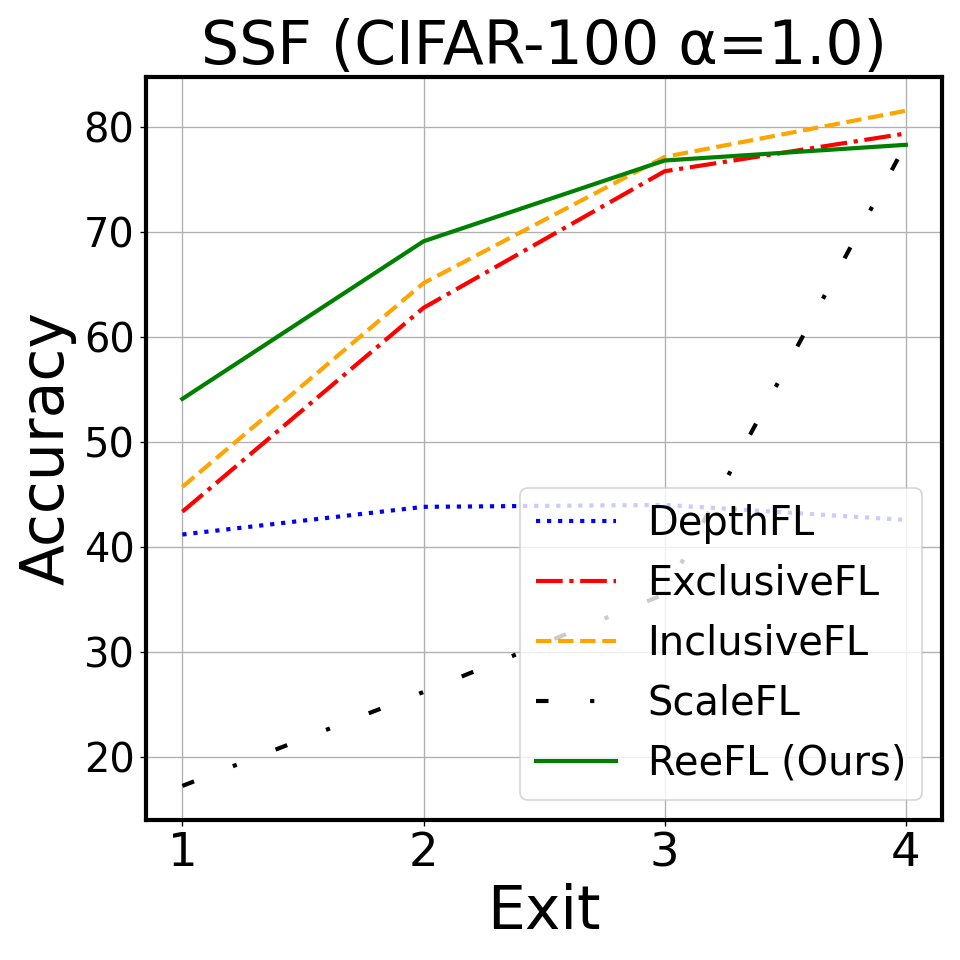}
\end{subfigure} \\

\begin{subfigure}{0.3\columnwidth}
    \includegraphics[trim=0 0 0 0, clip, width=0.97\columnwidth]{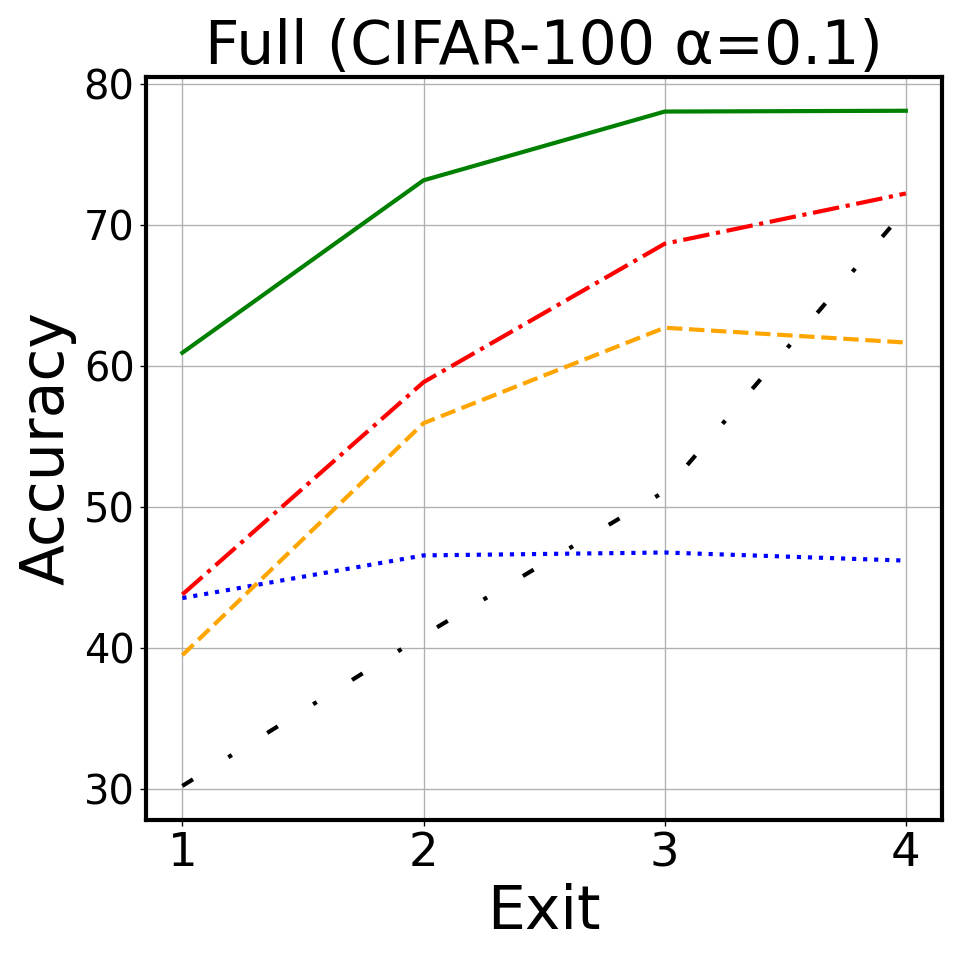}
\end{subfigure}
\begin{subfigure}{0.3\columnwidth}
    \includegraphics[trim=0 0 0 0, clip, width=0.97\columnwidth]{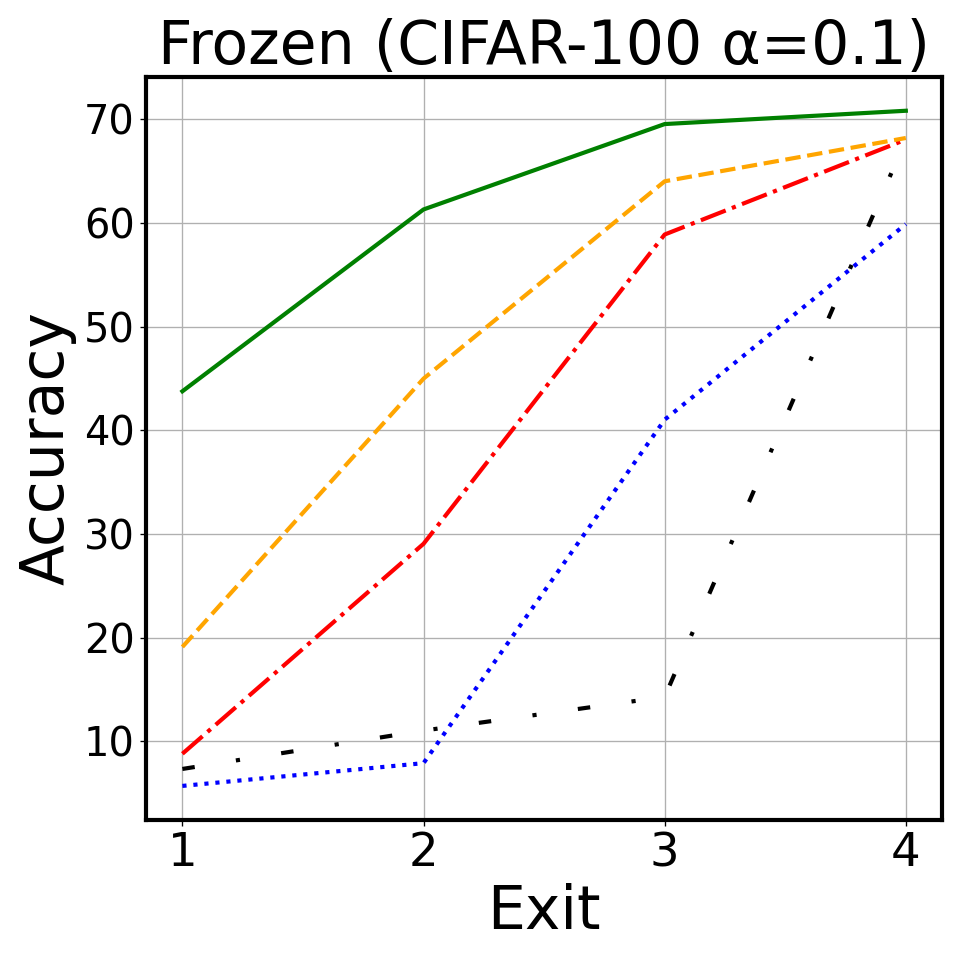}
\end{subfigure}
\begin{subfigure}{0.3\columnwidth}
    \includegraphics[trim=0 0 0 0, clip, width=0.97\columnwidth]{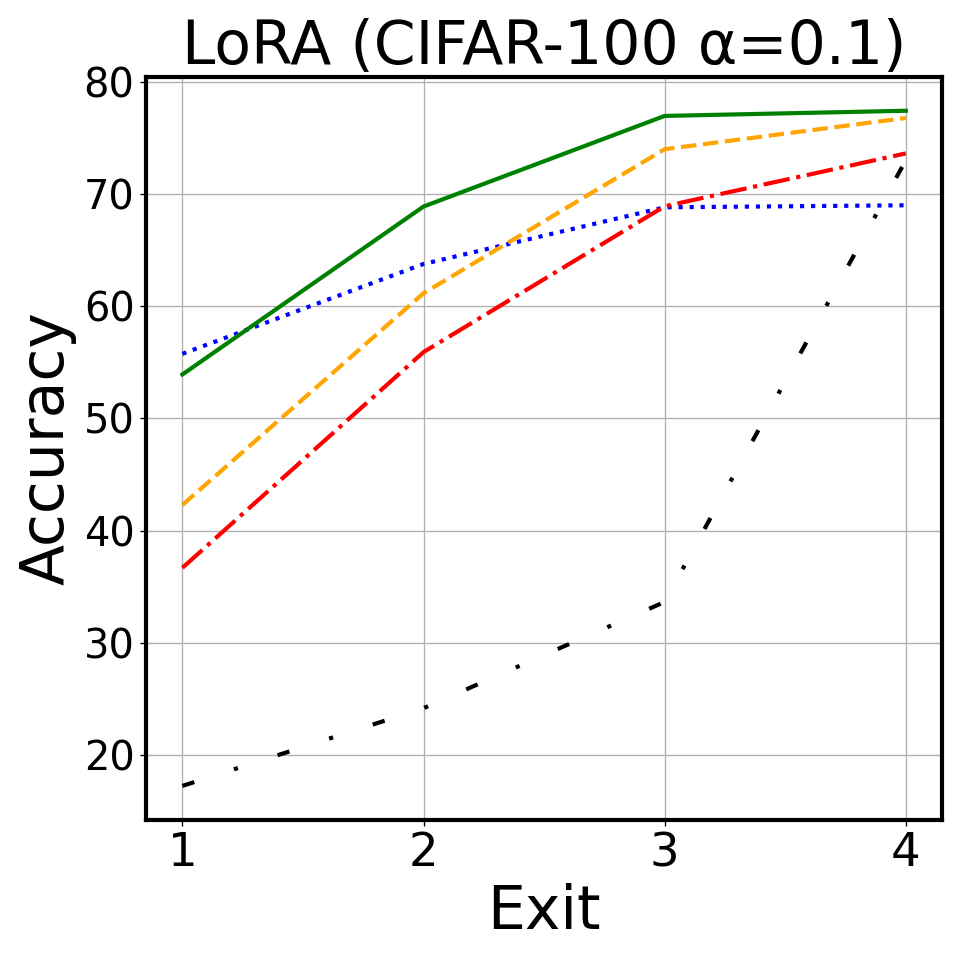}
\end{subfigure}
\begin{subfigure}{0.3\columnwidth}
    \includegraphics[trim=0 0 0 0, clip, width=0.97\columnwidth]{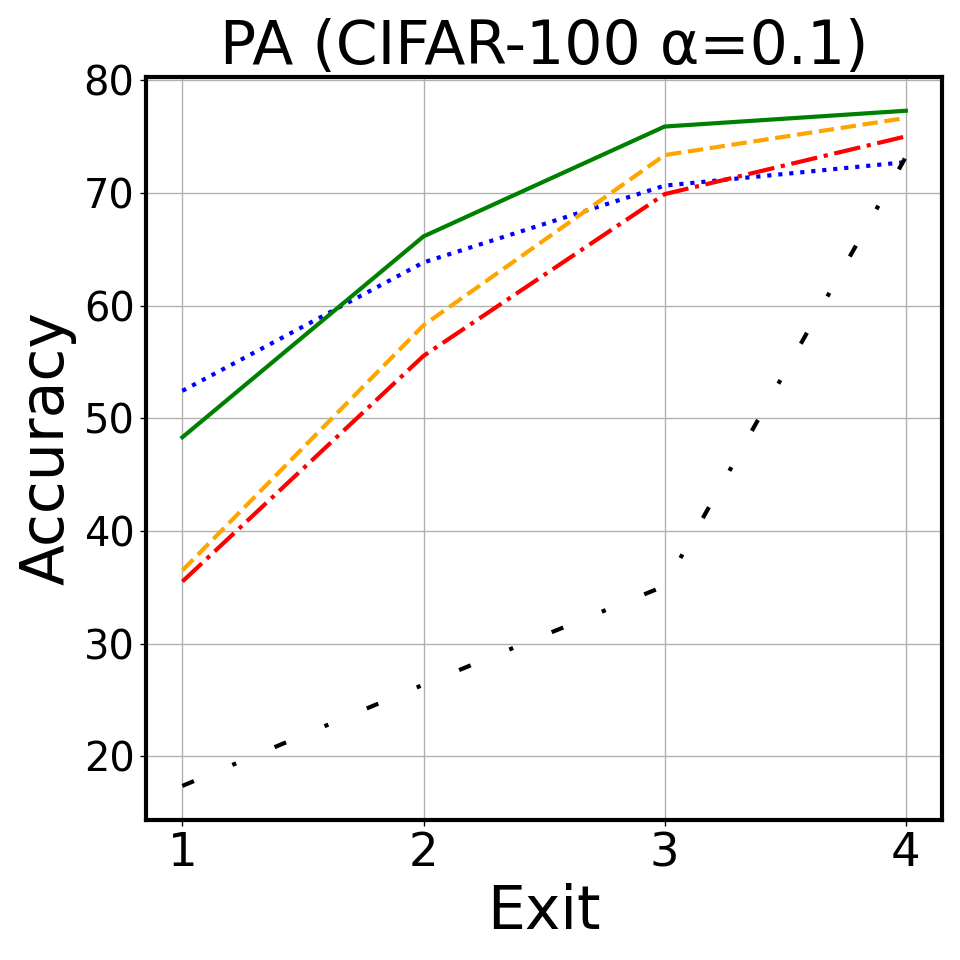}
\end{subfigure}
\begin{subfigure}{0.3\columnwidth}
    \includegraphics[trim=0 0 0 0, clip, width=0.97\columnwidth]{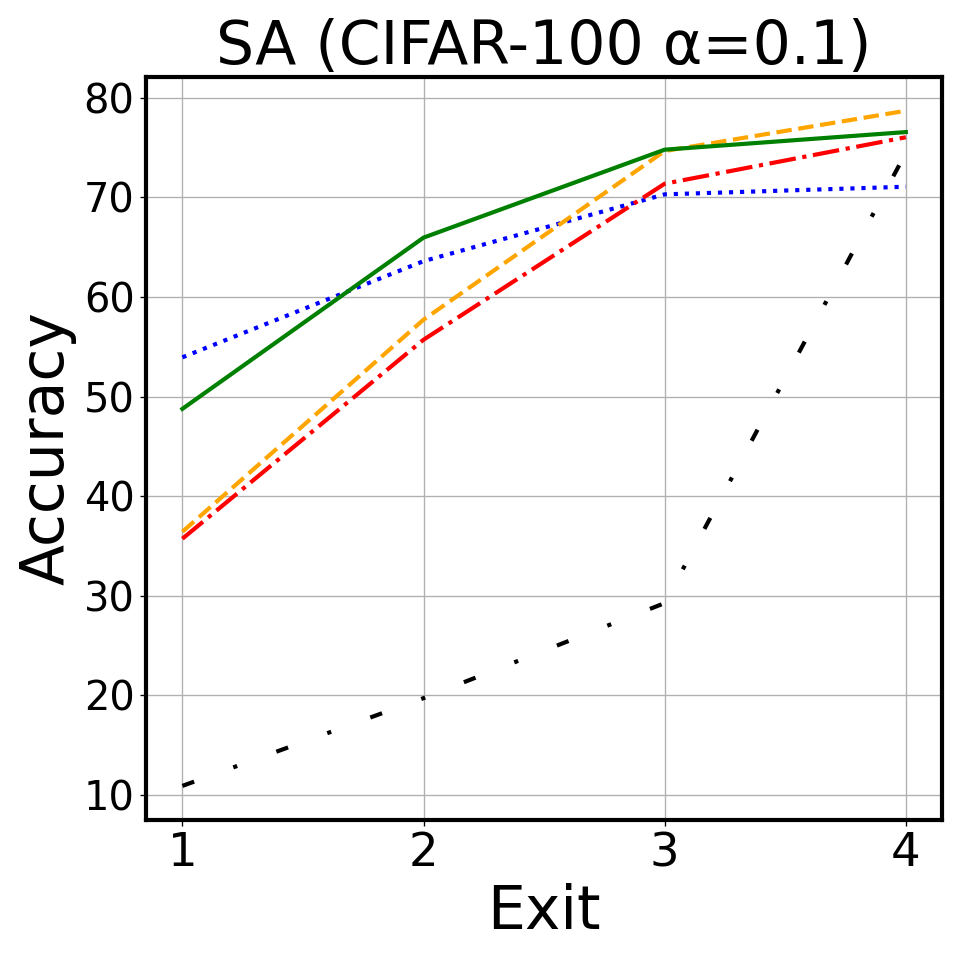}
\end{subfigure}
\begin{subfigure}{0.3\columnwidth}
    \includegraphics[trim=0 0 0 0, clip, width=0.97\columnwidth]{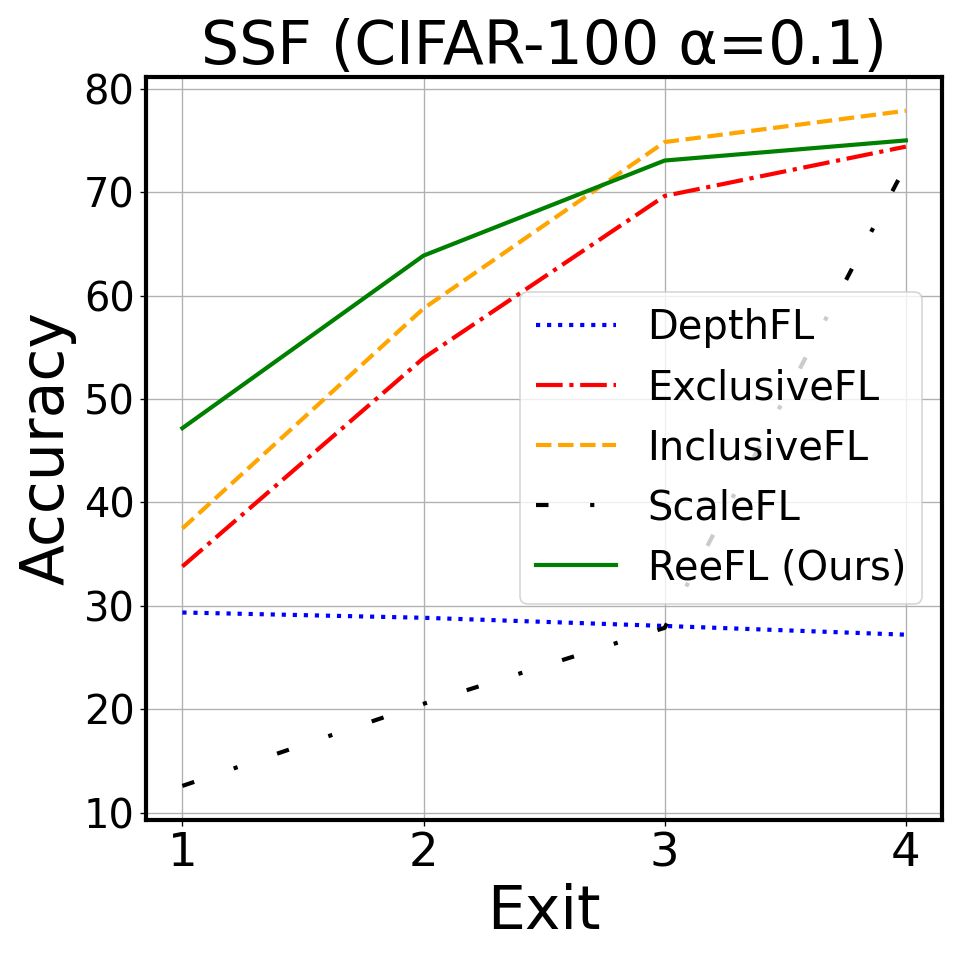}
\end{subfigure} \\

\begin{subfigure}{0.3\columnwidth}
    \includegraphics[trim=0 0 0 0, clip, width=0.97\columnwidth]{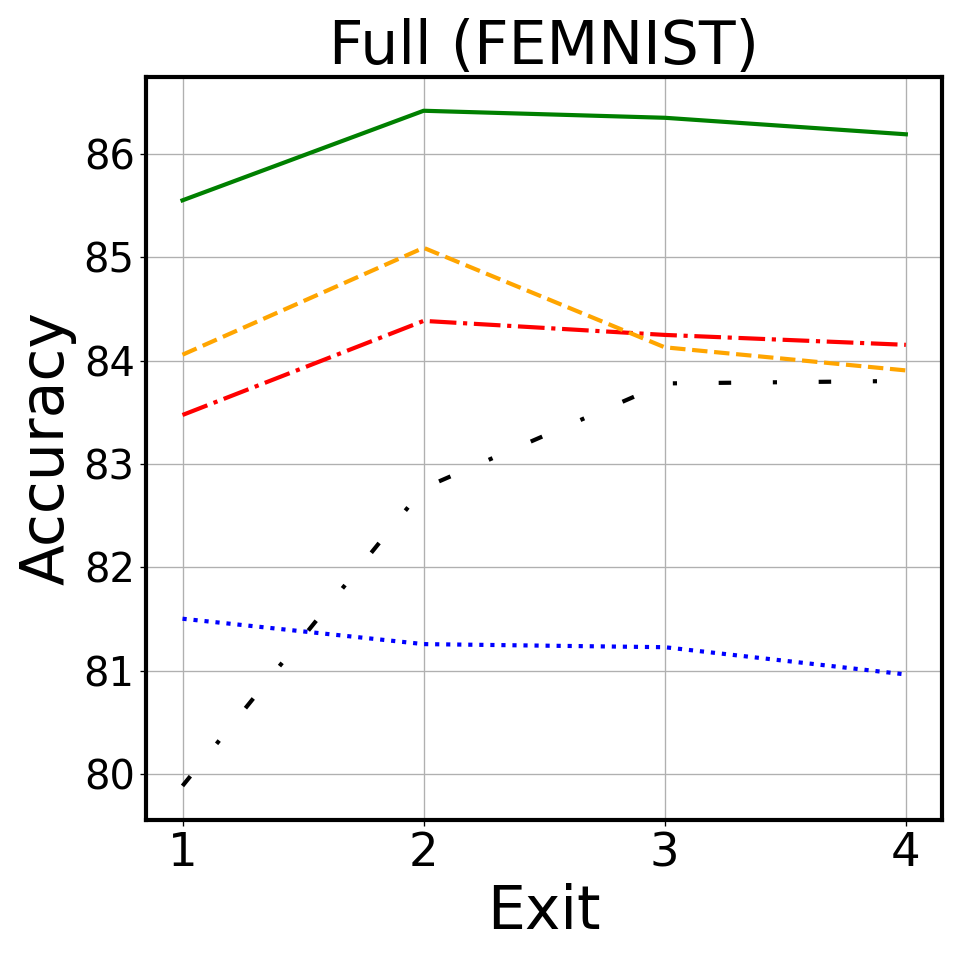}
\end{subfigure}
\begin{subfigure}{0.3\columnwidth}
    \includegraphics[trim=0 0 0 0, clip, width=0.97\columnwidth]{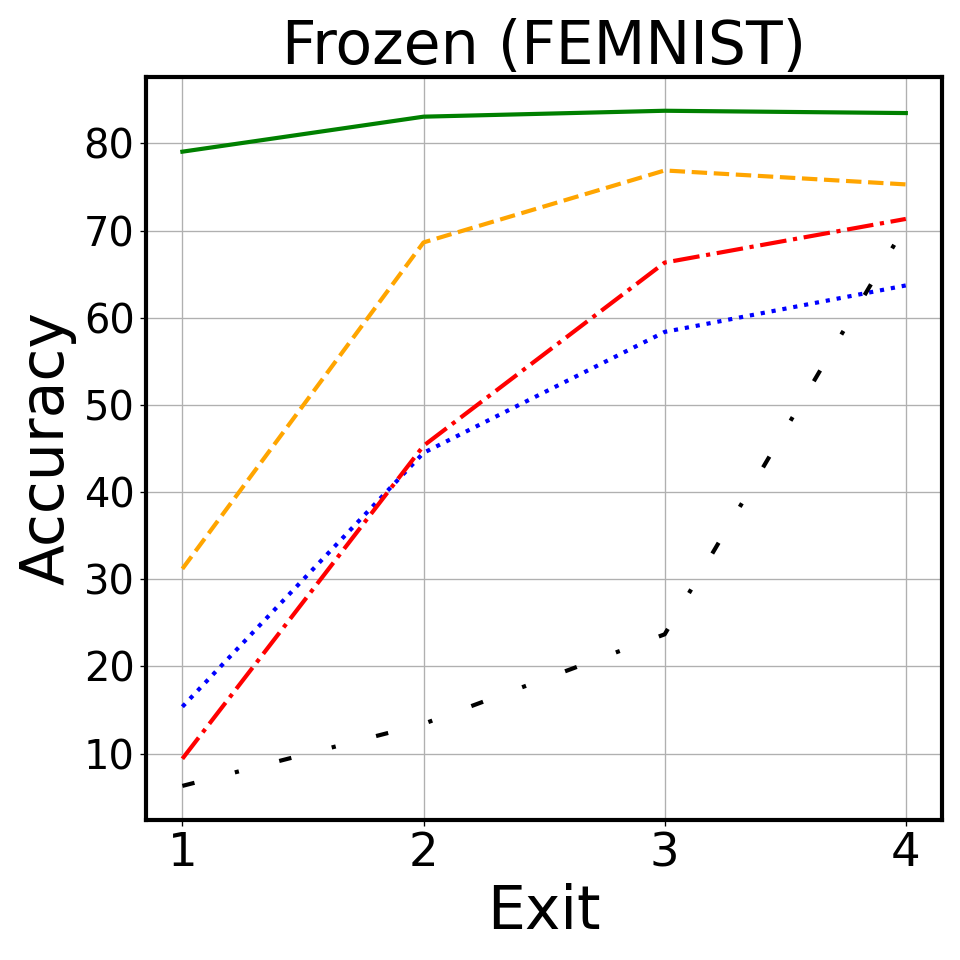}
\end{subfigure}
\begin{subfigure}{0.3\columnwidth}
    \includegraphics[trim=0 0 0 0, clip, width=0.97\columnwidth]{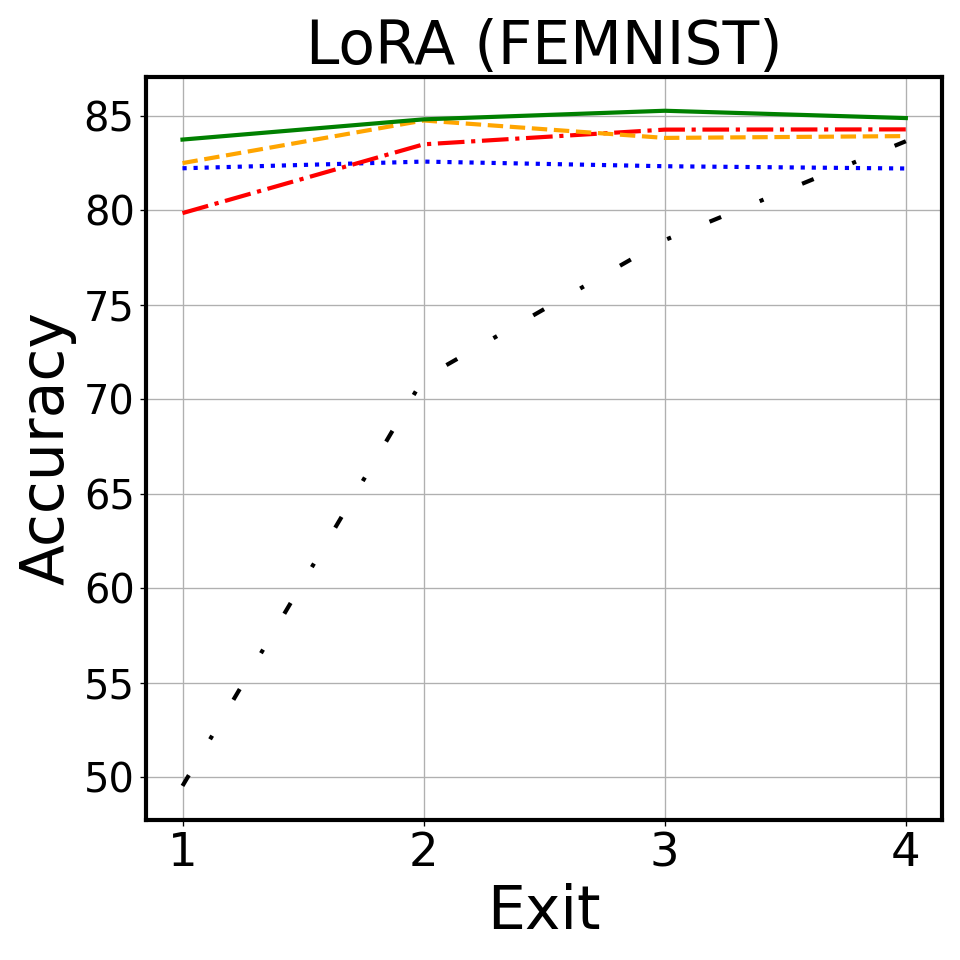}
\end{subfigure}
\begin{subfigure}{0.3\columnwidth}
    \includegraphics[trim=0 0 0 0, clip, width=0.97\columnwidth]{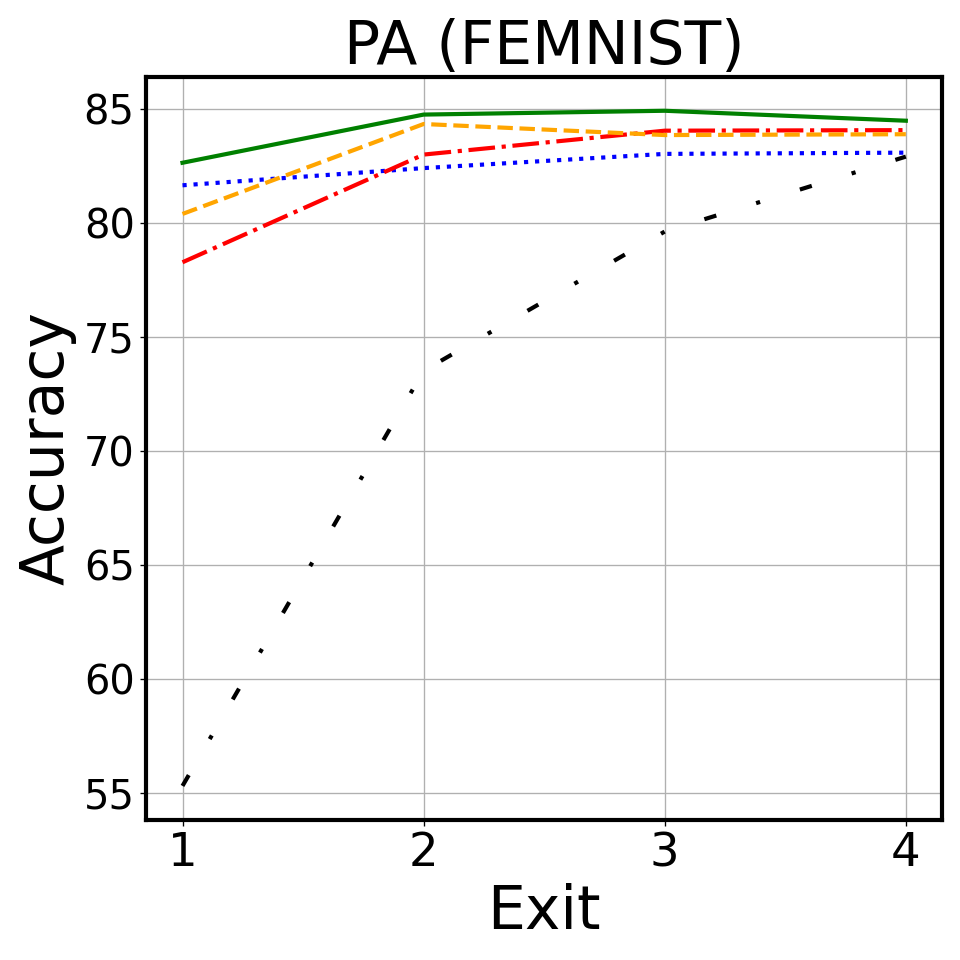}
\end{subfigure}
\begin{subfigure}{0.3\columnwidth}
    \includegraphics[trim=0 0 0 0, clip, width=0.97\columnwidth]{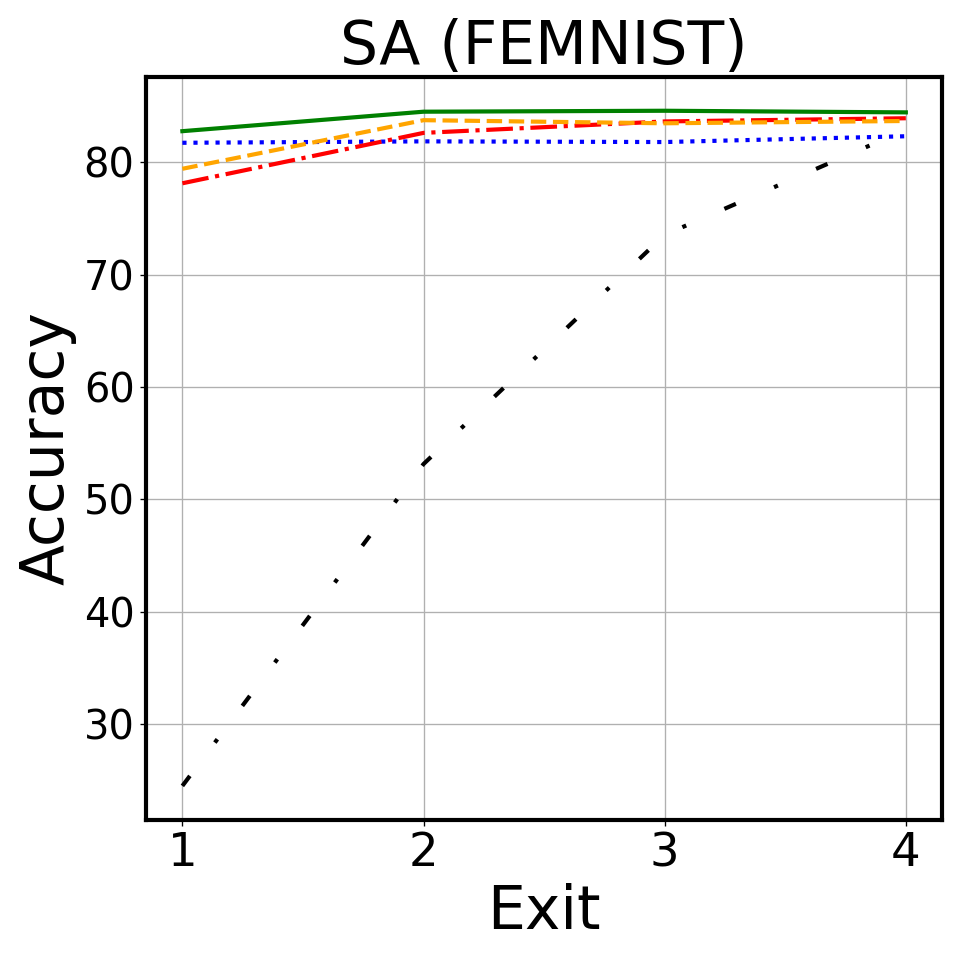}
\end{subfigure}
\begin{subfigure}{0.3\columnwidth}
    \includegraphics[trim=0 0 0 0, clip, width=0.97\columnwidth]{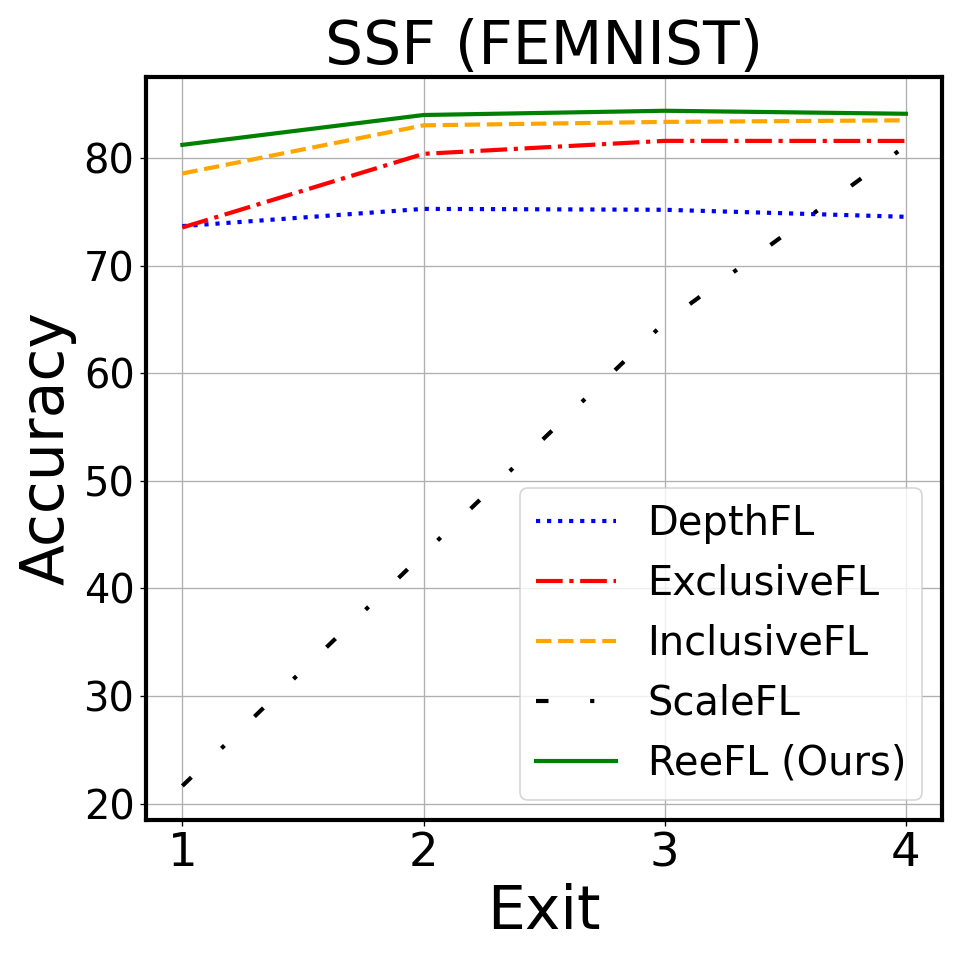}
\end{subfigure} \\

\begin{subfigure}{0.3\columnwidth}
    \includegraphics[trim=0 0 0 0, clip, width=0.97\columnwidth]{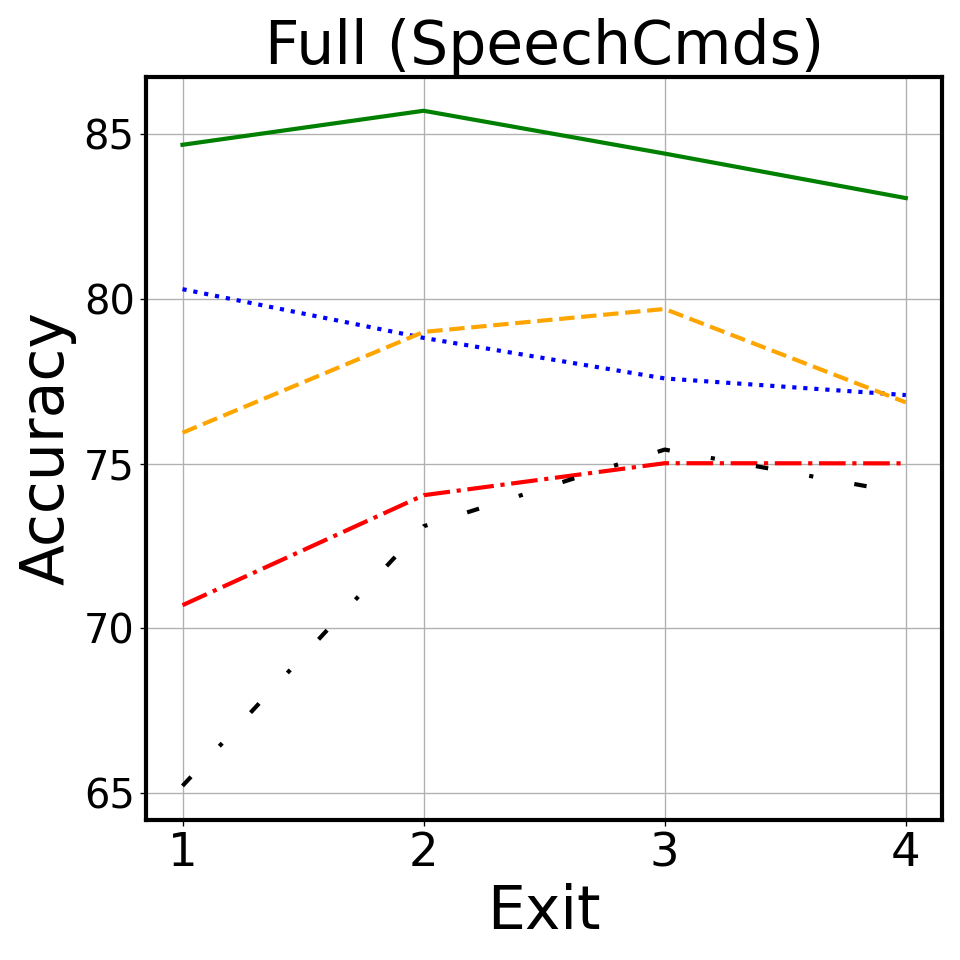}
\end{subfigure}
\begin{subfigure}{0.3\columnwidth}
    \includegraphics[trim=0 0 0 0, clip, width=0.97\columnwidth]{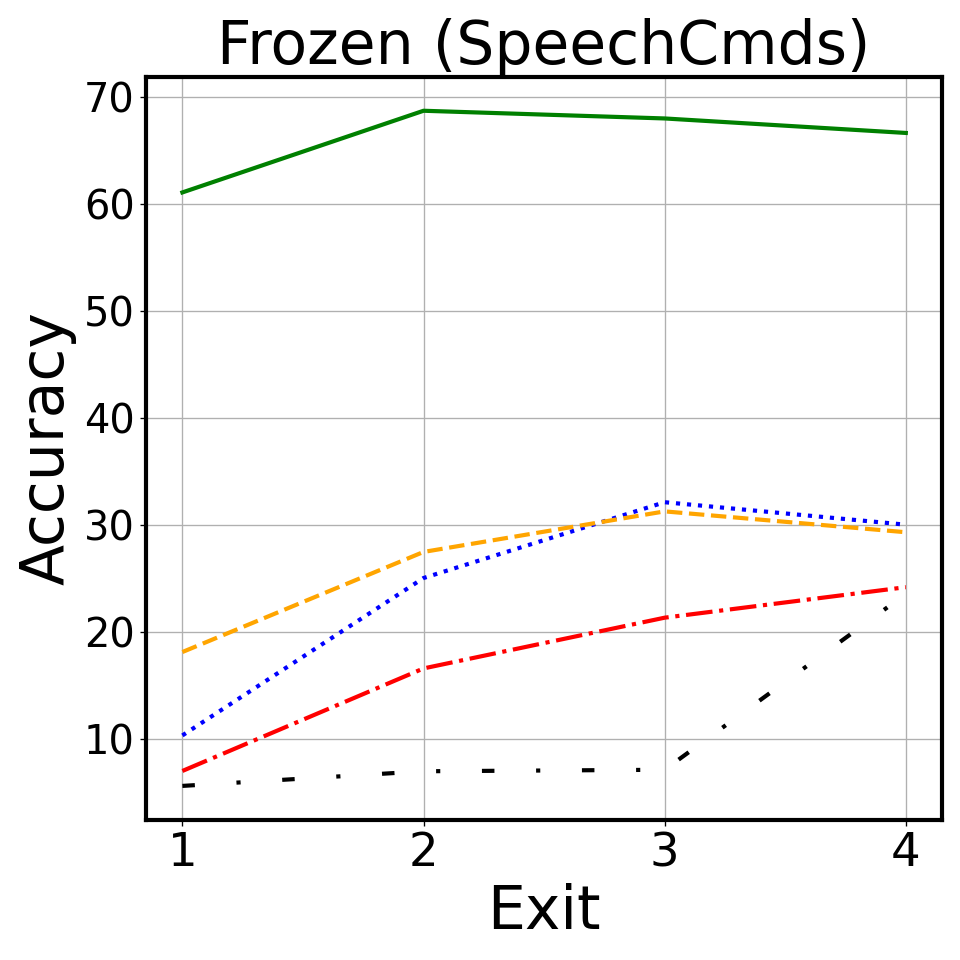}
\end{subfigure}
\begin{subfigure}{0.3\columnwidth}
    \includegraphics[trim=0 0 0 0, clip, width=0.97\columnwidth]{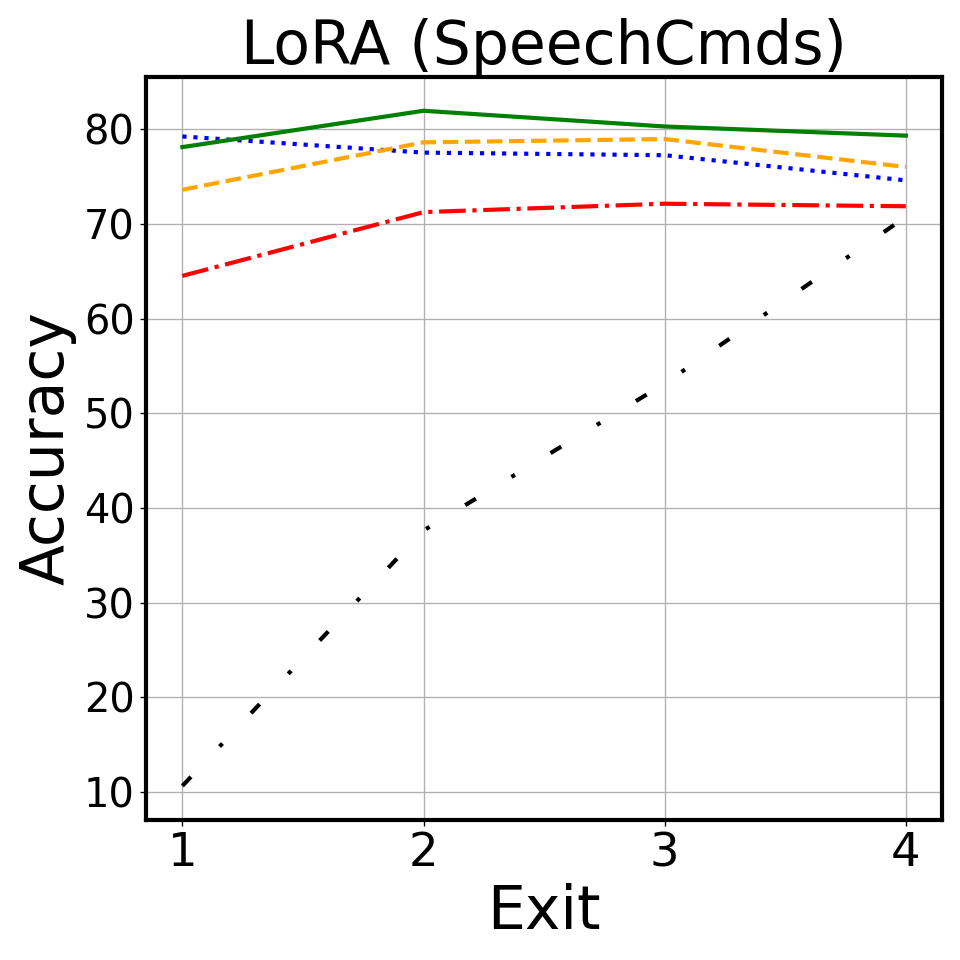}
\end{subfigure}
\begin{subfigure}{0.3\columnwidth}
    \includegraphics[trim=0 0 0 0, clip, width=0.97\columnwidth]{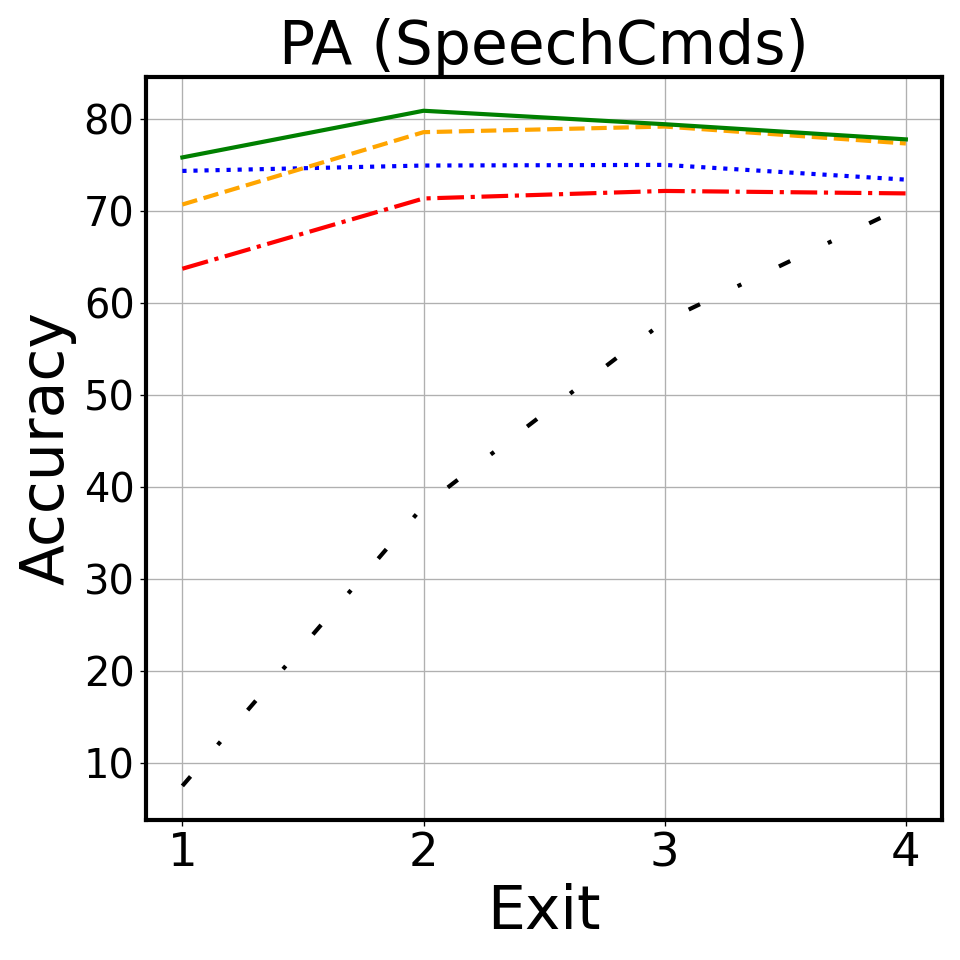}
\end{subfigure}
\begin{subfigure}{0.3\columnwidth}
    \includegraphics[trim=0 0 0 0, clip, width=0.97\columnwidth]{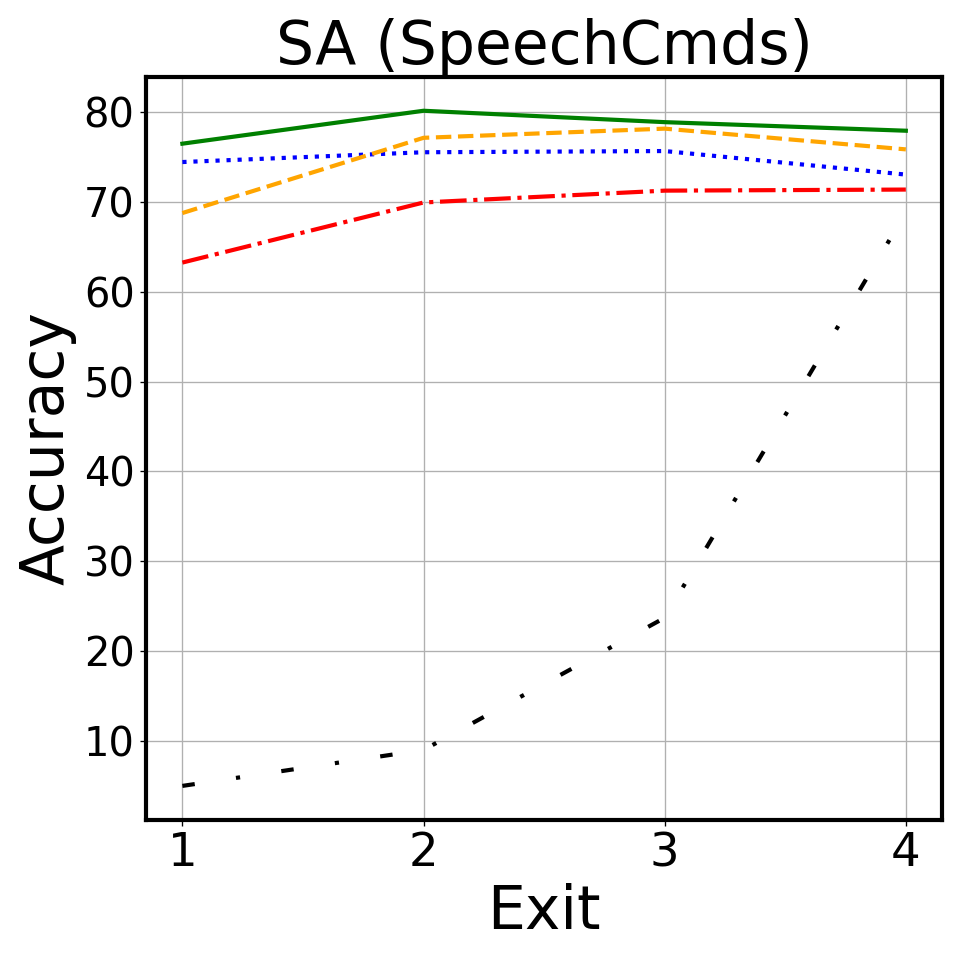}
\end{subfigure}
\begin{subfigure}{0.3\columnwidth}
    \includegraphics[trim=0 0 0 0, clip, width=0.97\columnwidth]{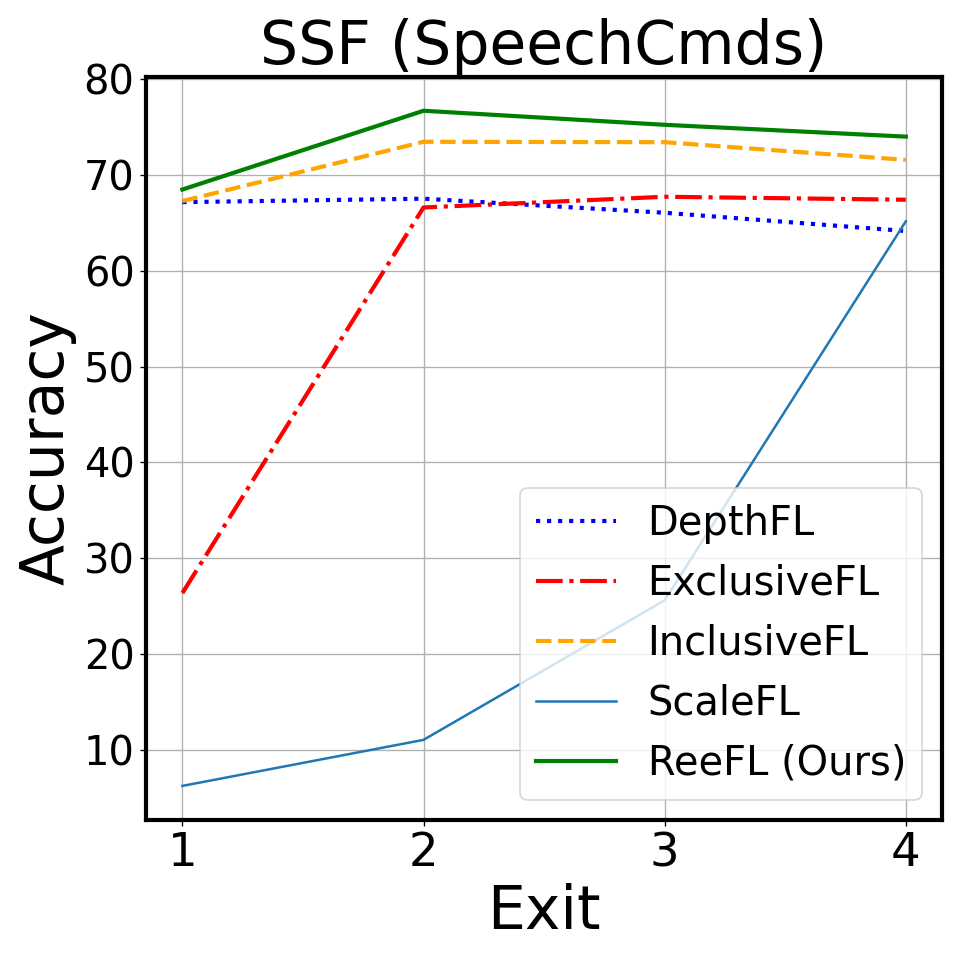}
\end{subfigure} \\

\
\vspace{-1.0em}
\caption{Fig~\ref{fig:ee_e12}'s extended results. Mean accuracy of each exit across 3 runs. }
\label{app:fig:ee_e4}
\end{figure*}

The following figures and tables extend the results in the main paper: Fig.~\ref{fig:profiling_cifar1000} \& \ref{fig:profiling_cifar0.1} show the accuracy and cost comparisons on CIFAR-100 $\alpha=1000$ and $\alpha=0.1$ respectively. Fig.~\ref{app:fig:ee_e4} \& \ref{app:fig:ee_e12} show the mean accuracy per exit for $4$ \& $12$ exits respectively. Fig.~\ref{fig:ablation1_ee_e12} shows the impact of our proposed knowledge distillation for $12$ exits. Lastly, Table.~\ref{app:tab:ablation1} shows an ablation on the aggregation strategies and knowledge distillation, comparing our \method{} with previous depth-based scaling methods for $12$ exits, and Table.~\ref{app:tab:ablation2} highlights the benefits of \method's feature modulation for $12$ exits.

\section{Attention Map Visualization}
\begin{figure*}[hbtp]
    \centering
    \includegraphics[width=.95\textwidth]{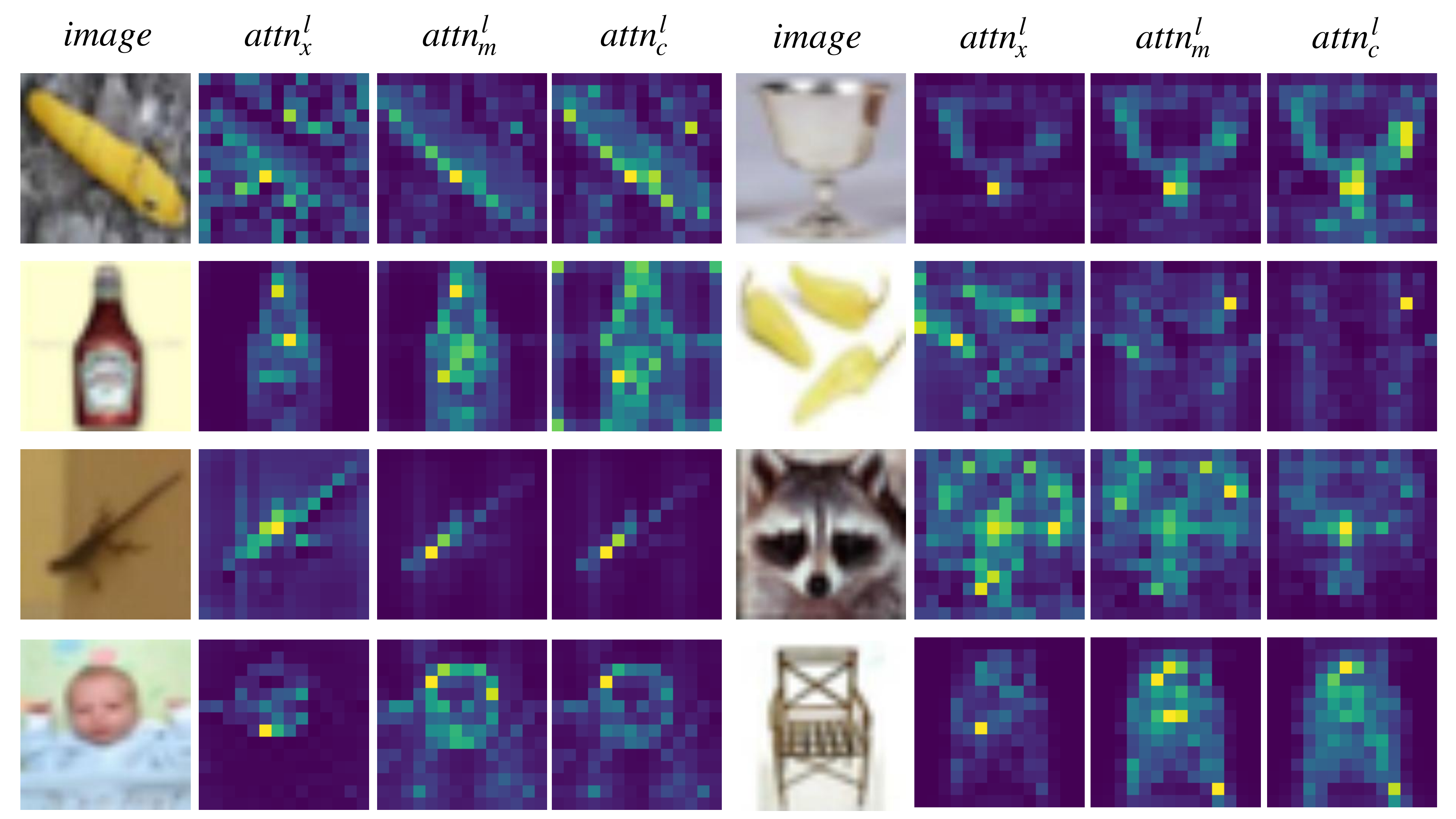}
    \caption{Attention map comparison between $\text{attn}_x^l$, $\text{attn}_m^l$ and $\text{attn}_c^l$. We select to visualize the earliest block whose classification is correct.}
    \label{fig:attn}
\end{figure*}

To understand the ``exploration'' and ``exploitation'' impact of \module{} on class tokens, $z_{\text{cls}}^l$, for $l= 1,\ldots, L$, we compare three attention maps:
\begin{align}
    &\text{attn}_x^l := \text{attentionMap}^l(z_{\text{cls}}^{l-1}, z_{1:n}^{l-1}) \\
    &\text{attn}_m^l := \text{attentionMap}^l(m_l^{l}, z_{1:n}^{l-1}) \\
    &\text{attn}_c^l := \text{attentionMap}^l(m_0^{l} + z_{\text{cls}}^{l}, z_{1:n}^{l-1})
\end{align}
where $\text{attentionMap}^l(a, b_{1:n})$ is the first row without the first element (i.e., attention between $a$ and $a$) of the attention map of $\text{MSA}^l(\text{LN}_1^l([a, b_1, \ldots, b_n]))$ of shape $(n+1, n+1, \#heads)$ after taking a mean operation over the multi-head dimension. We show a few examples for CIFAR-100 images in Fig~\ref{fig:attn}.

\section{Different Pretrained Models}\label{app:sec:more_models}

In this section, we further run experiments on the much larger DeiT-B model~\cite{Touvron21deit} and some additional preliminary experiments with a recent state space model (SSM), Vision Mamba (Vim)~\cite{zhu2024vision}. We then discuss the extendibility, limitation, and potential directions of applying \module{} to other architectures.

\textbf{DeiT-B}. We show results for 4 exits for Frozen and the best performing PEFT approach, LoRA, across 3 runs, in Table.~\ref{app:tab:deit-b}. Comparing Table.~\ref{app:tab:deit-b} and Table.~\ref{tab:mainres_e4}, utilizing the bigger DeiT-B results in a drop in performance compared with DeiT-S. This is primarily due to overfitting, fitting a much larger model on each client’s small dataset. Nonetheless, the relative performance ranking among these approaches stays the same across benchmarks for both DeiT-S and DeiT-B.

\textbf{Vim}. Besides transformers, we adopt pretrained Vim-T, the tiny variant of Vim which is much smaller than DeiT-S. As this is a different architecture, we run the same grid search detailed in Appendix Section.~\ref{app:sec:training_deets} and pick the best performing hyperparameters for each baseline. We show results for 4 exits comparing ReeFL with the second best-performing baseline, DepthFL, on both a frozen backbone and full fine-tuning in Table.~\ref{app:tab:vim}.

Comparing the Table.~\ref{app:tab:vim} with Table.~\ref{tab:mainres_e4}, utilizing features from a smaller frozen backbone model leads to a considerable drop in performance in all cases. Notably, for SpeechCommands, both ReeFL and DepthFL fail to utilize the pre-trained features. For full fine-tuning, we observe that using the smaller Vim-T mitigates the overfitting issue in DeiT-S+DepthFL, resulting in a gain in performance. Nonetheless, ReeFL still outperforms DepthFL and the performance gap between Vim-T and DeiT-S for ReeFL is consistent with the results shown in~\cite{zhu2024vision}. Note that as vision SSMs are a new addition to the literature, there is no consensus on how to effectively apply existing PEFT approaches to it. We leave the exploration of PEFT methods on SSM as a potential future work.

In general, \method{} can be applied, out-of-the-box, to any model that has a uniform state size at every layer. Extending to architectures with a non-uniform state size, for example ResNets, however, is non-trivial as \module{}’s feature modulation requires this state size to be uniform across layers. One possible future direction is to generate scaling and shifting parameters for feature modulation. 

\section{Additional Discussion}~\label{app:sec:detailed_discussion}

In this section, we supplement existing insights found in Section~\ref{sec:exp} by further discussing the performance gap found between \method{} and the second best-performing baseline DepthFL.
In Section.~\ref{sec:ablation}, we show that both knowledge distillation and feature modulation can help improve performance in \method{}. Nonetheless, in some cases, removing knowledge distillation and feature modulation in \method{} still leads to a better performance than the second best performing baseline, DepthFL. This performance gap comes with the use of the \module{} which fuses features from multiple layers for each exit and is trained on the full dataset in contrast with DepthFL which uses separate classifiers for each exit, where deeper classifiers are trained on partial data. 

To elucidate this, we focus on the same parameter aggregation (FedAvg) without knowledge distillation for both \method{} and DepthFL given a particular scenario (CIFAR-100 $\alpha=1000$, LoRA, 4 exits). In this scenario, DepthFL has a mean accuracy of 71.22 and ReeFL has a mean accuracy of 73.47 as shown in Table.~\ref{tab:ablation1}. We  remove 1) both the knowledge distillation and modulation of \method{} which results in a mean accuracy of 73.21 and 2) remove the \module{} module entirely and use a single shared classifier for different exits which results in a mean accuracy of 70.74. DepthFL outperforms the use of a single shared classifier (71.22 vs 70.74) as it uses multiple classifiers per exit, despite deeper classifiers being partially trained, to better handle the different feature representations of different layers. With the inclusion of the \module{} module, these multi-layer features are learnt to be aggregated to improve downstream classification, outperforming DepthFL (73.21 vs 71.22). We hypothesize that using FedDyn with \method{} may potentially lead to higher accuracy gains in some cases as it did with DepthFL and we leave that as a possible direction for future work.

We would also like to point out that in this scenario, simply using a shared classifier outperforms training with 25\% of the whole dataset: ExclusiveFL (70.74 vs 67.46). It also outperforms the case where each classifier is trained with separate subsets of the dataset: InclusiveFL (70.74 vs 69.37). This is not only because the shared classifier is trained on the full dataset but also because the features of the backbone model are fine-tuned jointly with the classifier via LoRA.  If we, instead, consider a frozen backbone (CIFAR-100 $\alpha=1000$, Frozen, 4 exits), using a shared classifier results in a mean accuracy of 41.74 which is significantly lower than ExclusiveFL (48.02) and InclusiveFL (53.99), as well as DepthFL, due to the differences in features among layers. Including the \module{} module can better handle these differences, achieving better performance over these baselines. 

\section{Ensemble Performance}\label{app:sec:ensemble}

We adopt the same ensemble method used in DepthFL: taking the average logits of all exits. Specifically, we pick the best performing run in Table.~\ref{tab:mainres_e4} and Table.~\ref{tab:mainres_e12}, for 4 and 12 exits respectively, and compute the ensemble performance for both DepthFL and \method{} (Table.~\ref{app:tab:ensemble_e4} and Table.~\ref{app:tab:ensemble_e12}).
Unsurprisingly, ensembles lead to an improvement in performance in most cases, with a drop in performance only in the case where the majority of the exits has poor performance.

\begin{table*}[t]
\caption{\textbf{4 exits} ensemble performance for both DepthFL and \method{}.}

\label{app:tab:ensemble_e4}
\begin{center}
\begin{tabular}{|l|l|ccc|c|c|}
\hline
\multicolumn{1}{|c|}{\multirow{2}{*}{Finetuning}} & \multicolumn{1}{c|}{\multirow{2}{*}{Approach}} & \multicolumn{3}{c|}{CIFAR-100} & \multirow{2}{*}{FEMNIST} & \multirow{2}{*}{SpeechCmds} \\ \cline{3-5}
\multicolumn{1}{|c|}{} & \multicolumn{1}{c|}{} & \multicolumn{1}{c|}{$\alpha$=1000} & \multicolumn{1}{c|}{$\alpha$=1.0} & $\alpha$=0.1 &  &  \\ \hline
Full & DepthFL & \multicolumn{1}{c|}{56.74} & \multicolumn{1}{c|}{54.66} & 46.13 & 81.77 & 78.33 \\
 & ReeFL (ours) & \multicolumn{1}{c|}{\textbf{81.16}} & \multicolumn{1}{c|}{\textbf{80.69}} & \textbf{79} & \textbf{86.71} & \textbf{85.06} \\ \hline
Frozen & DepthFL & \multicolumn{1}{c|}{66.67} & \multicolumn{1}{c|}{61.97} & 21.73 & 53.53 & 24.65 \\
 & ReeFL (ours) & \multicolumn{1}{c|}{\textbf{74.86}} & \multicolumn{1}{c|}{\textbf{73.58}} & \textbf{69.66} & \textbf{83.69} & \textbf{66.16} \\ \hline
LoRA & DepthFL & \multicolumn{1}{c|}{73.53} & \multicolumn{1}{c|}{72.08} & 66.37 & 81.61 & 76.66 \\
 & ReeFL (ours) & \multicolumn{1}{c|}{\textbf{80.32}} & \multicolumn{1}{c|}{\textbf{79.75}} & \textbf{76.05} & \textbf{85.38} & \textbf{79.97} \\ \hline
PA & DepthFL & \multicolumn{1}{c|}{75.21} & \multicolumn{1}{c|}{73.81} & 69.02 & 82.26 & 73.64 \\
 & ReeFL (ours) & \multicolumn{1}{c|}{\textbf{80.35}} & \multicolumn{1}{c|}{\textbf{79.34}} & \textbf{75.35} & \textbf{84.15} & \textbf{78.72} \\ \hline
SA & DepthFL & \multicolumn{1}{c|}{71.46} & \multicolumn{1}{c|}{73.61} & 67.16 & 82.52 & 74.66 \\
 & ReeFL (ours) & \multicolumn{1}{c|}{\textbf{72.25}} & \multicolumn{1}{c|}{\textbf{78.51}} & \textbf{73.86} & \textbf{84.6} & \textbf{78.12} \\ \hline
SSF & DepthFL & \multicolumn{1}{c|}{44.53} & \multicolumn{1}{c|}{41.94} & 31.36 & 74.88 & 65.04 \\
 & ReeFL (ours) & \multicolumn{1}{c|}{\textbf{70.17}} & \multicolumn{1}{c|}{\textbf{76.72}} & \textbf{72.4} & \textbf{84.37} & \textbf{74.36} \\ \hline
\end{tabular}
\end{center}
\end{table*}
\begin{table*}[t]
\caption{\textbf{12 exits} ensemble performance for both DepthFL and \method{}.}

\label{app:tab:ensemble_e12}
\begin{center}
\begin{tabular}{|l|l|ccc|c|c|}
\hline
\multicolumn{1}{|c|}{\multirow{2}{*}{Finetuning}} & \multicolumn{1}{c|}{\multirow{2}{*}{Approach}} & \multicolumn{3}{c|}{CIFAR-100} & \multirow{2}{*}{FEMNIST} & \multirow{2}{*}{SpeechCmds} \\ \cline{3-5}
\multicolumn{1}{|c|}{} & \multicolumn{1}{c|}{} & \multicolumn{1}{c|}{$\alpha$=1000} & \multicolumn{1}{c|}{$\alpha$=1.0} & $\alpha$=0.1 &  &  \\ \hline
Full & DepthFL & \multicolumn{1}{c|}{47.26} & \multicolumn{1}{c|}{45.96} & 35.3 & 80.53 & 70.18 \\
 & ReeFL (ours) & \multicolumn{1}{c|}{\textbf{74.9}} & \multicolumn{1}{c|}{\textbf{73.8}} & \textbf{66.08} & \textbf{85.75} & \textbf{81.95} \\ \hline
Frozen & DepthFL & \multicolumn{1}{c|}{65.74} & \multicolumn{1}{c|}{61.75} & 23.15 & 53.35 & 21.32 \\
 & ReeFL (ours) & \multicolumn{1}{c|}{\textbf{71.72}} & \multicolumn{1}{c|}{\textbf{71.5}} & \textbf{63.58} & \textbf{84.35} & \textbf{60.01} \\ \hline
LoRA & DepthFL & \multicolumn{1}{c|}{69.26} & \multicolumn{1}{c|}{67.82} & 56 & 79.53 & 69.83 \\
 & ReeFL (ours) & \multicolumn{1}{c|}{\textbf{76.03}} & \multicolumn{1}{c|}{\textbf{75.63}} & \textbf{69.01} & \textbf{85.12} & \textbf{75.48} \\ \hline
PA & DepthFL & \multicolumn{1}{c|}{71.26} & \multicolumn{1}{c|}{70.49} & 1.02 & 5.01 & 65.21 \\
 & ReeFL (ours) & \multicolumn{1}{c|}{\textbf{75.91}} & \multicolumn{1}{c|}{\textbf{75.44}} & \textbf{68.56} & \textbf{77.66} & \textbf{72.32} \\ \hline
SA & DepthFL & \multicolumn{1}{c|}{73.88} & \multicolumn{1}{c|}{71.92} & 1 & 4.11 & 65.44 \\
 & ReeFL (ours) & \multicolumn{1}{c|}{\textbf{76.52}} & \multicolumn{1}{c|}{\textbf{74.86}} & \textbf{68.04} & \textbf{85.25} & \textbf{71.86} \\ \hline
SSF & DepthFL & \multicolumn{1}{c|}{52.5} & \multicolumn{1}{c|}{48.62} & 1.02 & 5.21 & 3.11 \\
 & ReeFL (ours) & \multicolumn{1}{c|}{\textbf{72.65}} & \multicolumn{1}{c|}{\textbf{72.49}} & \textbf{65.81} & \textbf{84.54} & \textbf{67.24} \\ \hline
\end{tabular}
\end{center}
\end{table*}

\begin{table*}[t]
\caption{Experiments on DeiT-B. Mean and standard deviation (SD) of the mean performance of all \textbf{4 exits} across 3 runs.}

\label{app:tab:deit-b}
\begin{center}
\begin{tabular}{|l|l|ccc|c|c|}
\hline
\multicolumn{1}{|c|}{\multirow{2}{*}{Finetuning}} & \multicolumn{1}{c|}{\multirow{2}{*}{Approach}} & \multicolumn{3}{c|}{CIFAR-100} & \multirow{2}{*}{FEMNIST} & \multirow{2}{*}{SpeechCmds} \\ \cline{3-5}
\multicolumn{1}{|c|}{} & \multicolumn{1}{c|}{} & \multicolumn{1}{c|}{$\alpha$=1000} & \multicolumn{1}{c|}{$\alpha$=1.0} & $\alpha$=0.1 &  &  \\ \hline
\multirow{5}{*}{Frozen} & ExclusiveFL & \multicolumn{1}{c|}{46.36±0.03} & \multicolumn{1}{c|}{45.56±0.05} & 40.92±0.03 & 49.03±0.01 & 19.62±0.05 \\
 & InclusiveFL & \multicolumn{1}{c|}{52.97±0.11} & \multicolumn{1}{c|}{51.97±0.01} & 48.24±0.04 & 64.37±0.03 & 30.14±0.01 \\
 & ScaleFL & \multicolumn{1}{c|}{29.96±0.0} & \multicolumn{1}{c|}{28.92±0.0} & 26.43±0.0 & 29.5±0.01 & 11.95±0.01 \\
 & DepthFL & \multicolumn{1}{c|}{46.67±0.21} & \multicolumn{1}{c|}{42.75±0.37} & 28.41±0.1 & 35.2±1.0 & 24.68±0.33 \\
 & ReeFL (ours) & \multicolumn{1}{c|}{\textbf{67.6±0.03}} & \multicolumn{1}{c|}{\textbf{66.47±0.08}} & \textbf{62.36±0.14} & \textbf{81.06±0.03} & \textbf{65.37±0.24} \\ \hline
\multirow{5}{*}{LoRA} & ExclusiveFL & \multicolumn{1}{c|}{65.24±0.06} & \multicolumn{1}{c|}{64.22±0.04} & 57.23±0.08 & 81.68±0.06 & 67.04±0.06 \\
 & InclusiveFL & \multicolumn{1}{c|}{66.77±0.01} & \multicolumn{1}{c|}{66.32±0.01} & 61.23±0.04 & 83.3±0.01 & 76.04±0.13 \\
 & ScaleFL & \multicolumn{1}{c|}{51.11±0.03} & \multicolumn{1}{c|}{50.27±0.06} & 43.49±0.12 & 66.74±0.04 & 39.25±0.11 \\
 & DepthFL & \multicolumn{1}{c|}{70.4±0.11} & \multicolumn{1}{c|}{69.04±0.14} & 61.85±0.05 & 80.78±0.71 & 76.38±0.19 \\
 & ReeFL (ours) & \multicolumn{1}{c|}{\textbf{71.92±0.01}} & \multicolumn{1}{c|}{\textbf{71.29±0.1}} & \textbf{66.94±0.01} & \textbf{83.85±0.16} & \textbf{77.07±0.05} \\ \hline
\end{tabular}
\end{center}
\end{table*}

\begin{table*}[t]
\caption{Experiments on Vim comparing ReeFL with DepthFL. Mean and standard deviation (SD) of the mean performance of all \textbf{4 exits} across 3 runs.}

\label{app:tab:vim}
\begin{center}

\begin{tabular}{|l|l|ccc|c|c|}
\hline
\multicolumn{1}{|c|}{\multirow{2}{*}{Finetuning}} & \multicolumn{1}{c|}{\multirow{2}{*}{Approach}} & \multicolumn{3}{c|}{CIFAR-100} & \multirow{2}{*}{FEMNIST} & \multirow{2}{*}{SpeechCmds} \\ \cline{3-5}
\multicolumn{1}{|c|}{} & \multicolumn{1}{c|}{} & \multicolumn{1}{c|}{$\alpha$=1000} & \multicolumn{1}{c|}{$\alpha$=1.0} & $\alpha$=0.1 &  &  \\ \hline
Frozen & DepthFL & \multicolumn{1}{c|}{37.23±0.01} & \multicolumn{1}{c|}{36.53±0.08} & 27.82±0.42 & 45.41±0.65 & \textbf{18.92±0.02} \\
 & ReeFL (ours) & \multicolumn{1}{c|}{\textbf{46.63±0.48}} & \multicolumn{1}{c|}{\textbf{45.43±1.28}} & \textbf{38.53±0.3} & \textbf{74.17±0.37} & 11.95±1.45 \\ \cline{2-7} \hline
Full & DepthFL & \multicolumn{1}{c|}{64.36±0.11} & \multicolumn{1}{c|}{63.87±0.1} & 56.93±0.18 & 80.0±0.25 & 70.07±0.68 \\
 & ReeFL (ours) & \multicolumn{1}{c|}{\textbf{74.08±0.1}} & \multicolumn{1}{c|}{\textbf{73.79±0.1}} & \textbf{70.68±0.23} & \textbf{85.42±0.0} & \textbf{81.87±0.28} \\ \hline
\end{tabular}

\end{center}
\end{table*}


\begin{table*}[t]
\caption{Ablation study on different aggregation strategies and knowledge distillation of depth-based scaling methods on $12$ exits.}
\label{app:tab:ablation1}
%
\begin{center}
\resizebox{1.0\textwidth}{!}{
\begin{tabular}{|l|lcl|ccc|c|c|}
\hline
\multicolumn{1}{|c|}{\multirow{2}{*}{Finetuning}} & \multicolumn{1}{c|}{\multirow{2}{*}{Approach}} & \multicolumn{1}{c|}{\multirow{2}{*}{Distillation}} & \multicolumn{1}{c|}{\multirow{2}{*}{Aggregation}} & \multicolumn{3}{c|}{CIFAR-100} & \multirow{2}{*}{FEMNIST} & \multirow{2}{*}{SpeechCmds} \\ \cline{5-7}
\multicolumn{1}{|c|}{} & \multicolumn{1}{c|}{} & \multicolumn{1}{c|}{} & \multicolumn{1}{c|}{} & \multicolumn{1}{c|}{$\alpha$=1000} & \multicolumn{1}{c|}{$\alpha$=1.0} & $\alpha$=0.1 &  &  \\ \hline
\multirow{8}{*}{Frozen} & InclusiveFL & - & FedAvg & \multicolumn{1}{c|}{36.98±0.01} & \multicolumn{1}{c|}{34.59±0.02} & 26.44±0.02 & 40.88±0.03 & 13.93±0.05 \\
 & InclusiveFL & - & FedAdam & \multicolumn{1}{c|}{43.53±0.03} & \multicolumn{1}{c|}{41.6±0.04} & 32.28±0.05 & 53.35±0.03 & 20.22±0.05 \\
 & DepthFL & \xmark & FedAvg & \multicolumn{1}{c|}{42.6±0.02} & \multicolumn{1}{c|}{41.87±0.02} & 36.62±0.01 & 43.11±0.17 & 15.47±0.03 \\
 & DepthFL & \xmark & FedDyn & \multicolumn{1}{c|}{45.12±0.02} & \multicolumn{1}{c|}{43.4±0.16} & 36.09±0.39 & 48.52±0.23 & 21.19±0.17 \\
 & DepthFL & \cmark & FedAvg & \multicolumn{1}{c|}{43.31±0.0} & \multicolumn{1}{c|}{42.55±0.0} & 36.75±0.13 & 43.25±0.01 & 16.01±0.2 \\
 & DepthFL & \cmark & FedDyn & \multicolumn{1}{c|}{45.22±0.06} & \multicolumn{1}{c|}{42.93±0.18} & 25.2±1.89 & 41.29±0.1 & 21.3±0.02 \\
 & ReeFL (ours) & \xmark & FedAvg & \multicolumn{1}{c|}{58.28±0.03} & \multicolumn{1}{c|}{57.7±0.03} & 51.6±0.32 & 74.87±0.15 & \textbf{59.69±0.57} \\
 & ReeFL (ours) & \cmark & FedAvg & \multicolumn{1}{c|}{\textbf{58.74±0.05}} & \multicolumn{1}{c|}{\textbf{58.1±0.09}} & \textbf{51.33±0.36} & \textbf{75.36±0.02} & 59.5±0.5 \\ \hline
\multirow{10}{*}{LoRA} & InclusiveFL & \xmark & FedAvg & \multicolumn{1}{c|}{55.79±0.03} & \multicolumn{1}{c|}{54.17±0.01} & 40.35±0.06 & 73.24±0.01 & 55.01±0.14 \\
 & InclusiveFL & \xmark & FedAdam & \multicolumn{1}{c|}{55.9±0.08} & \multicolumn{1}{c|}{54.54±0.15} & 41.53±0.15 & 78.21±0.15 & 66.18±0.58 \\
 & InclusiveFL & \cmark & FedAvg & \multicolumn{1}{c|}{52.17±0.01} & \multicolumn{1}{c|}{50.24±0.07} & 37.61±0.05 & 69.22±0.07 & 53.53±0.01 \\
 & InclusiveFL & \cmark & FedAdam & \multicolumn{1}{c|}{54.13±0.19} & \multicolumn{1}{c|}{52.69±0.1} & 39.32±0.09 & 78.32±0.14 & 67.49±0.03 \\
 & DepthFL & \xmark & FedAvg & \multicolumn{1}{c|}{63.13±0.08} & \multicolumn{1}{c|}{62.79±0.07} & 56.69±0.23 & 77.55±0.1 & 67.47±0.23 \\
 & DepthFL & \xmark & FedDyn & \multicolumn{1}{c|}{62.79±0.11} & \multicolumn{1}{c|}{61.55±0.17} & 53.5±0.15 & 78.87±0.27 & 70.91±0.2 \\
 & DepthFL & \cmark & FedAvg & \multicolumn{1}{c|}{62.73±0.0} & \multicolumn{1}{c|}{62.17±0.01} & 54.9±0.01 & 77.54±0.0 & 66.62±0.08 \\
 & DepthFL & \cmark & FedDyn & \multicolumn{1}{c|}{62.74±0.09} & \multicolumn{1}{c|}{61.3±0.11} & 51.9±0.15 & 78.97±0.15 & 70.66±0.07 \\
 & ReeFL (ours) & \xmark & FedAvg & \multicolumn{1}{c|}{65.16±0.04} & \multicolumn{1}{c|}{64.2±0.09} & 58.69±0.09 & 79.68±0.19 & 74.18±0.25 \\
 & ReeFL (ours) & \cmark & FedAvg & \multicolumn{1}{c|}{\textbf{65.7±0.09}} & \multicolumn{1}{c|}{\textbf{65.01±0.03}} & \textbf{58.85±0.05} & \textbf{81.06±0.06} & \textbf{75.96±0.18} \\ \hline
\end{tabular}
}
\end{center}
\end{table*}
\begin{table*}[t]
\caption{Mean Accuracy with and without ReeFL's feature modulation on $12$ exits.}
\label{app:tab:ablation2}
\begin{scriptsize}
\begin{center}
\resizebox{1.0\textwidth}{!}{
\begin{tabular}{|l|c|ccc|c|c|}
\hline
\multicolumn{1}{|c|}{\multirow{2}{*}{Finetuning}} & \multirow{2}{*}{Modulation} & \multicolumn{3}{c|}{CIFAR-100} & \multirow{2}{*}{FEMNIST} & \multirow{2}{*}{SpeechCmds} \\ \cline{3-5}
\multicolumn{1}{|c|}{} &  & \multicolumn{1}{c|}{$\alpha$=1000} & \multicolumn{1}{c|}{$\alpha$=1.0} & $\alpha$=0.1 &  &  \\ \hline
\multirow{2}{*}{Frozen} & \xmark & \multicolumn{1}{c|}{43.4±0.1} & \multicolumn{1}{c|}{41.88±0.29} & 25.6±17.4 & 49.59±0.37 & 17.52±0.07 \\
 & \cmark & \multicolumn{1}{c|}{\textbf{58.74±0.05}} & \multicolumn{1}{c|}{\textbf{58.1±0.09}} & \textbf{51.33±0.36} & \textbf{75.36±0.02} & \textbf{59.5±0.5} \\ \hline
\multirow{2}{*}{LoRA} & \xmark & \multicolumn{1}{c|}{64.99±0.12} & \multicolumn{1}{c|}{64.68±0.1} & 58.69±0.15 & 80.75±0.07 & \textbf{76.5±0.27} \\
 & \cmark & \multicolumn{1}{c|}{\textbf{65.7±0.09}} & \multicolumn{1}{c|}{\textbf{65.01±0.03}} & \textbf{58.85±0.05} & \textbf{81.06±0.06} & 75.96±0.18 \\ \hline
\multirow{2}{*}{PA} & \xmark & \multicolumn{1}{c|}{60.5±0.14} & \multicolumn{1}{c|}{60.28±0.07} & 53.97±0.1 & 77.01±0.05 & 71.13±0.08 \\
 & \cmark & \multicolumn{1}{c|}{\textbf{62.35±0.07}} & \multicolumn{1}{c|}{\textbf{61.47±0.12}} & \textbf{55.67±0.14} & \textbf{77.62±0.04} & \textbf{72.37±0.02} \\ \hline
\multirow{2}{*}{SA} & \xmark & \multicolumn{1}{c|}{61.83±0.05} & \multicolumn{1}{c|}{60.95±0.12} & \textbf{55.19±0.12} & 76.75±0.06 & 70.69±0.23 \\
 & \cmark & \multicolumn{1}{c|}{\textbf{63.28±0.13}} & \multicolumn{1}{c|}{\textbf{61.94±0.2}} & 54.99±0.1 & \textbf{77.43±0.04} & \textbf{71.52±0.13} \\ \hline
\multirow{2}{*}{SSF} & \xmark & \multicolumn{1}{c|}{58.18±0.09} & \multicolumn{1}{c|}{57.46±0.13} & 51.33±0.64 & 75.79±0.08 & 65.02±0.23 \\
 & \cmark & \multicolumn{1}{c|}{\textbf{61.85±0.02}} & \multicolumn{1}{c|}{\textbf{61.25±0.17}} & \textbf{54.42±0.13} & \textbf{78.05±0.03} & \textbf{67.44±0.24} \\ \hline
\end{tabular}
}
\end{center}
\end{scriptsize}

\end{table*}

\end{document}